\documentclass[runningheads]{llncs}

\usepackage{eccv}

\usepackage{eccvabbrv}
\usepackage{bm}
\usepackage{pifont}
\usepackage{graphicx}
\usepackage{booktabs}
\usepackage{overpic}
\usepackage{algorithm}
\usepackage{algpseudocode}
\usepackage{tabularx} 
 
\newcolumntype{Y}{>{\centering\arraybackslash}X}
\usepackage{makecell}  
\usepackage[table]{xcolor}
\usepackage{multirow}  
\usepackage{xcolor}
\usepackage{rotating}
\usepackage{marvosym}
\usepackage[accsupp]{axessibility}  % Improves PDF readability for those with disabilities.
\renewcommand{\algorithmicrequire}{\textbf{Input:}}
\renewcommand{\algorithmicensure}{\textbf{Output:}}

\usepackage[pagebackref,breaklinks,colorlinks,citecolor=eccvblue]{hyperref}
\usepackage{orcidlink}
\definecolor{ColorFMSoftBlue}{HTML}{EAF3FF}
\definecolor{ColorFMSoftGold}{HTML}{FFF4DB}
\begin{document}

\title{ColorFM: An Optimization-to-Learning Framework for Color Transfer via Flow Matching}

\titlerunning{ColorFM for Color Transfer}

\author{Yuhang He\inst{1} \and
Kai Zhang\inst{1}\textsuperscript{\Letter}\orcidlink{0000-0002-6319-3722} \and Xiaoming Li\inst{1} \and Du Chen\inst{2} \and Jian Yang\inst{1}}

\authorrunning{Y. He et al.}

\institute{School of Intelligence Science and Technology, Nanjing University, China \and
VIVO BlueImage Lab, China\\
\url{https://github.com/cszn/ColorFM}
}

\maketitle
\begingroup
\renewcommand{\thefootnote}{\Letter}
\footnotetext{Corresponding author (kaizhang@nju.edu.cn)}
\endgroup
\begin{abstract}
Color transfer aims to align the color distribution of a source image with that of a reference image while preserving structural and semantic consistency. However, existing methods often suffer from inaccurate global mapping, semantic misalignment, and visual artifacts. To address these issues, we propose ColorFM, an optimization-to-learning framework. ColorFM connects online optimization to offline inference by reformulating color transfer as the transport of pixel distributions along velocity fields via Flow Matching. Specifically, we introduce ColorFM-O, an instance-specific optimization scheme that fits the velocity field through hierarchical color coupling guided by semantic priors. By numerically integrating the induced flow trajectories, ColorFM-O produces precise and semantically consistent color transfer results, while generating high-quality paired data as pseudo-supervision. Building upon this, we design ColorFM-L, an efficient feed-forward model trained on the generated pairs. Through implicit state modeling, ColorFM-L extracts deep semantic features to predict flow parameters for bidirectional linearized transport, ensuring accurate color transfer. Extensive experiments demonstrate that ColorFM-L outperforms state-of-the-art methods in visual quality, structural fidelity, and semantic consistency, successfully combining the accuracy of optimization with the speed of feed-forward inference.
\end{abstract}

\section{Introduction}
\label{sec:intro}

 Color retouching plays a crucial role in visual communication and digital photography, as it shapes the perceived style and atmosphere of an image. However, the strong interdependence among color channels makes precise adjustment difficult for non-experts. While traditional approaches like image filters and Look-Up Tables (LUTs) offer simplified solutions, they typically apply rigid transformations that disregard the intrinsic color distribution of the source image, often leading to suboptimal results. 
To overcome these limitations, many automated color transfer techniques have been proposed. Given a reference style image, these methods aim to transfer its color style to the content image while maintaining photorealism.

Broadly, existing methods can be categorized into two paradigms: online optimization and offline inference. Optimization-based approaches \cite{pitie2007automated,pitie2007linear,luan2017deep,chen2023nlut,li2025d, pitie2005n} are often instance-specific and computationally intensive. 
Moreover, their reliance on hand-crafted or imperfect optimization objectives frequently results in inaccurate color transfer.
Conversely, offline techniques are predominantly learning-based \cite{li2018closed,yoo2019photorealistic,chiu2022photowct2,ke2023neural,gong2025sa,larchenko2025color,wen2023cap,xia2020joint,lin2023adacm,ho2021deep,an2020ultrafast,hong2021domain  }. Feature-transformation-based methods \cite{li2018closed,yoo2019photorealistic,chiu2022photowct2, wen2023cap} match deep statistics between images but often introduce visual artifacts or severe color banding. Mapping-based approaches \cite{ke2023neural,gong2025sa,lin2023adacm} (\eg, LUT generation~\cite{zeng2020learning, liu20234d}) learn direct color transformations, yet their performance is limited by data scarcity and synthetic dataset biases, leading to poor generalization in complex real-world scenes. Furthermore, a recent flow-based method \cite{larchenko2025color} focuses on learning global distribution alignment, often overlooking image content, which degrades visual quality.

\begin{figure*}[t]
  \centering
  \setlength{\tabcolsep}{0.5pt} 
  \renewcommand{\arraystretch}{0.5} 
  \begin{tabularx}{\textwidth}{Y Y Y Y Y} 
    \makebox[\linewidth][r]{\hspace{-6pt}\rotatebox{90}{\scriptsize Style}}&
    \includegraphics[width=\linewidth]{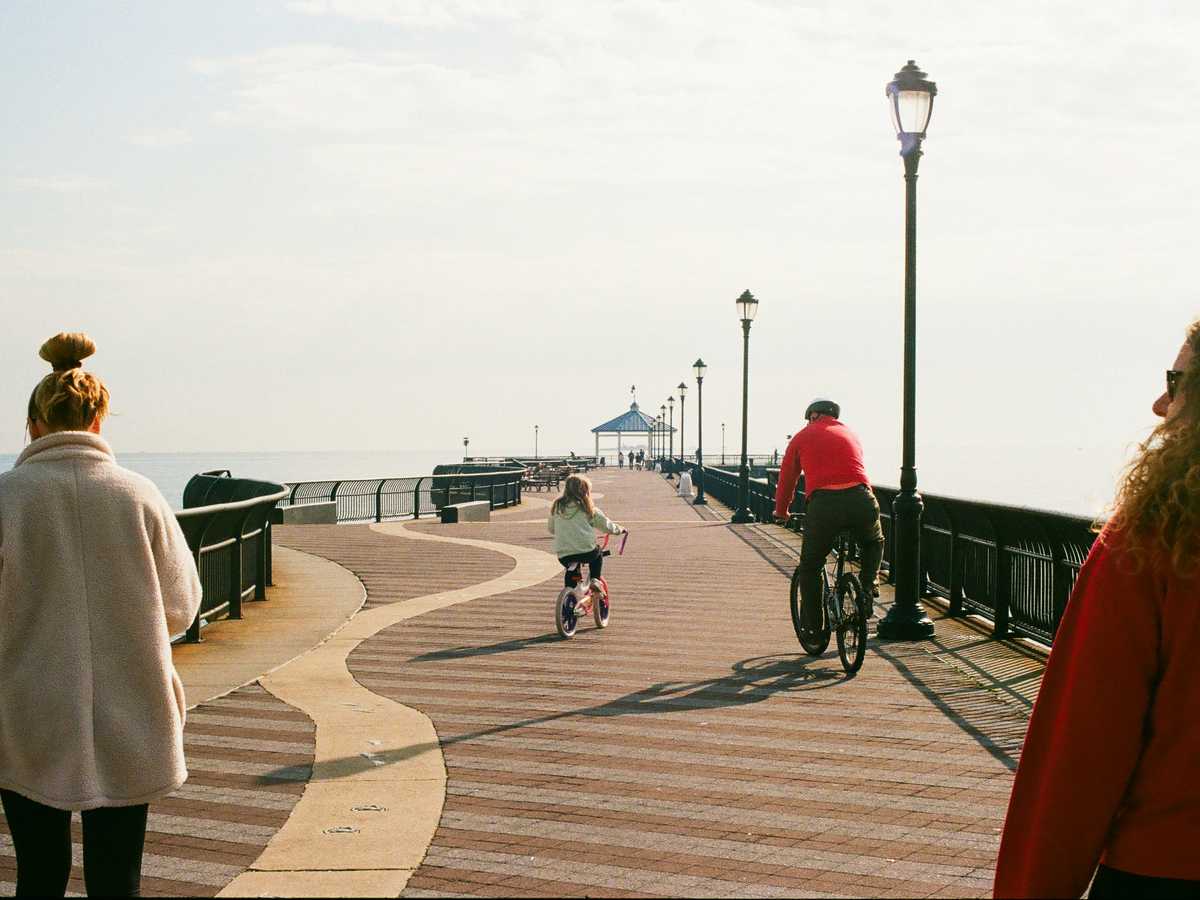} &
    \includegraphics[width=\linewidth]{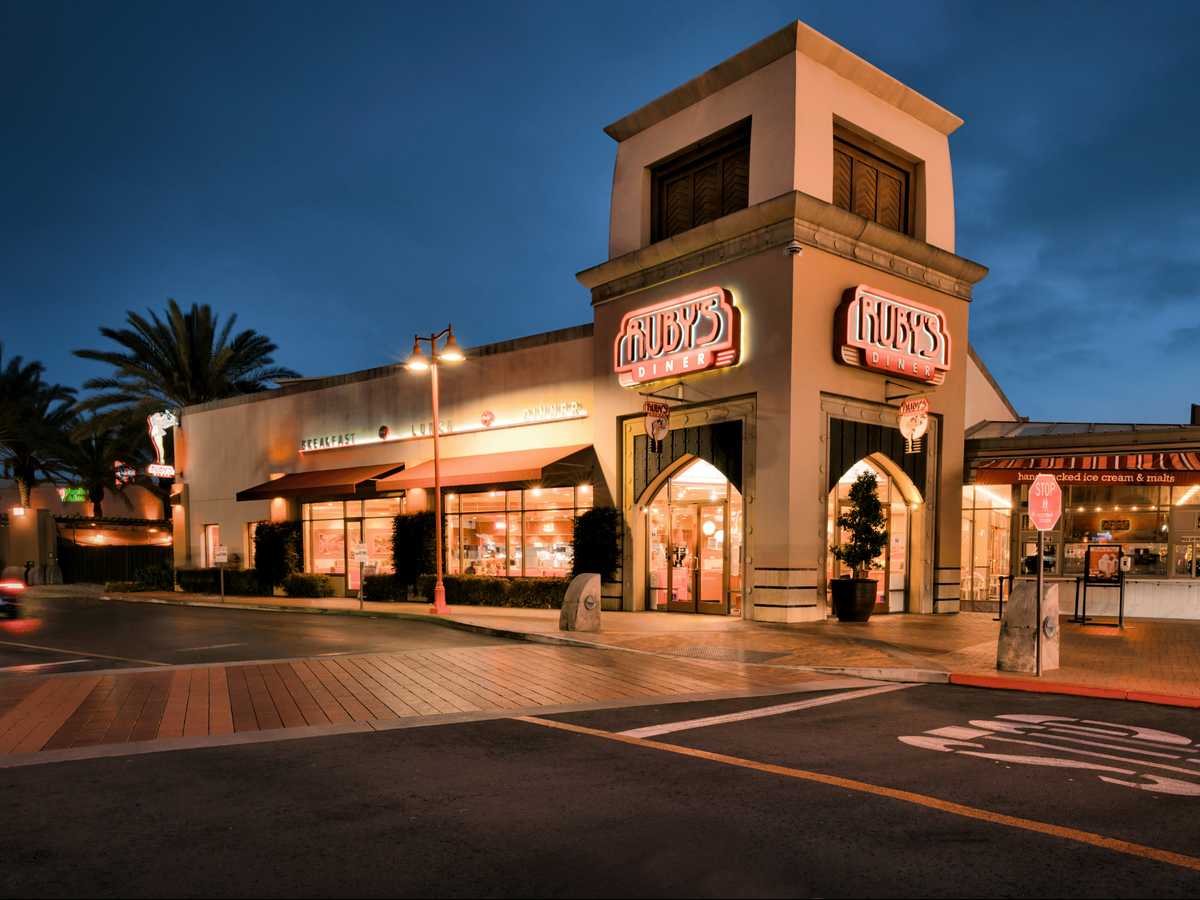} &
    \includegraphics[width=\linewidth]{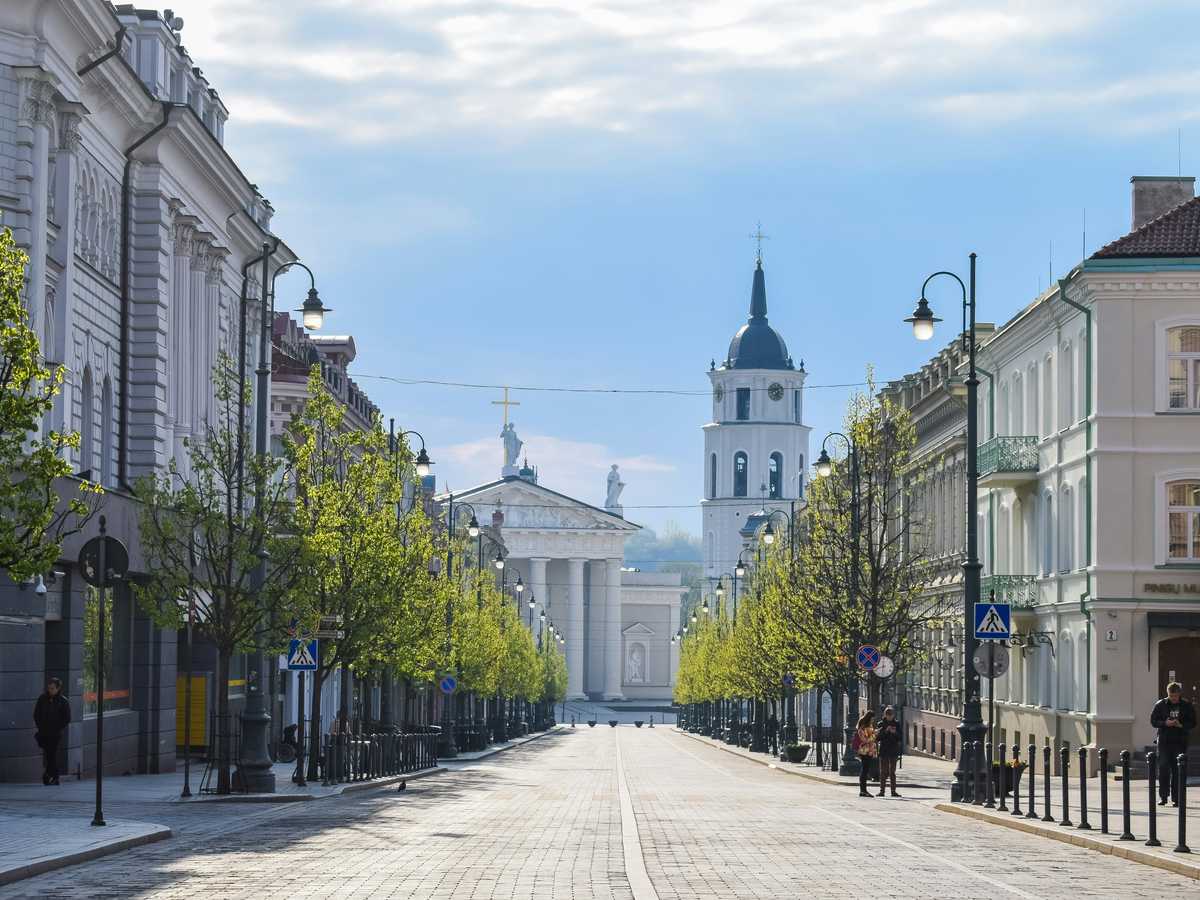} &
    \includegraphics[width=\linewidth]{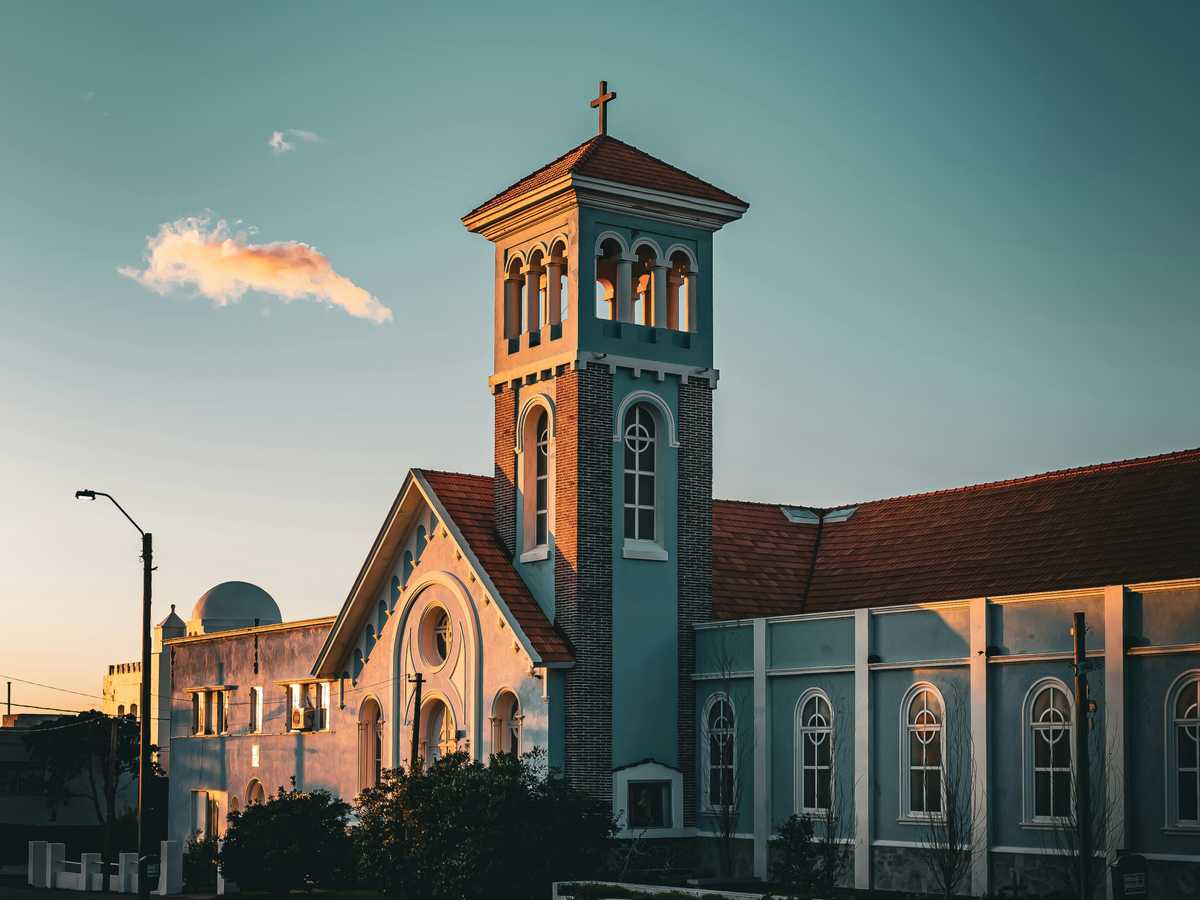} \\

    \includegraphics[width=\linewidth]{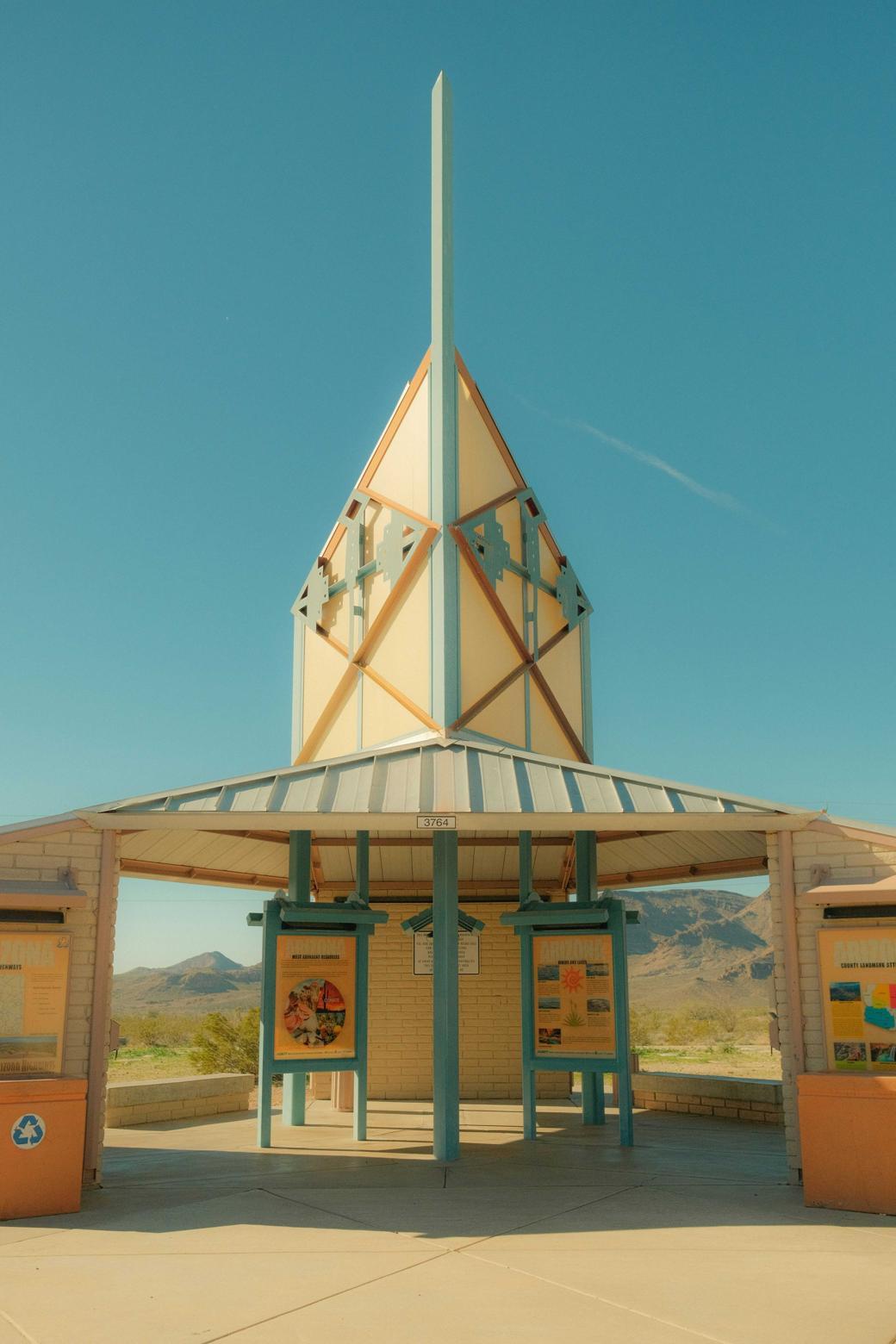} &
    \includegraphics[width=\linewidth]{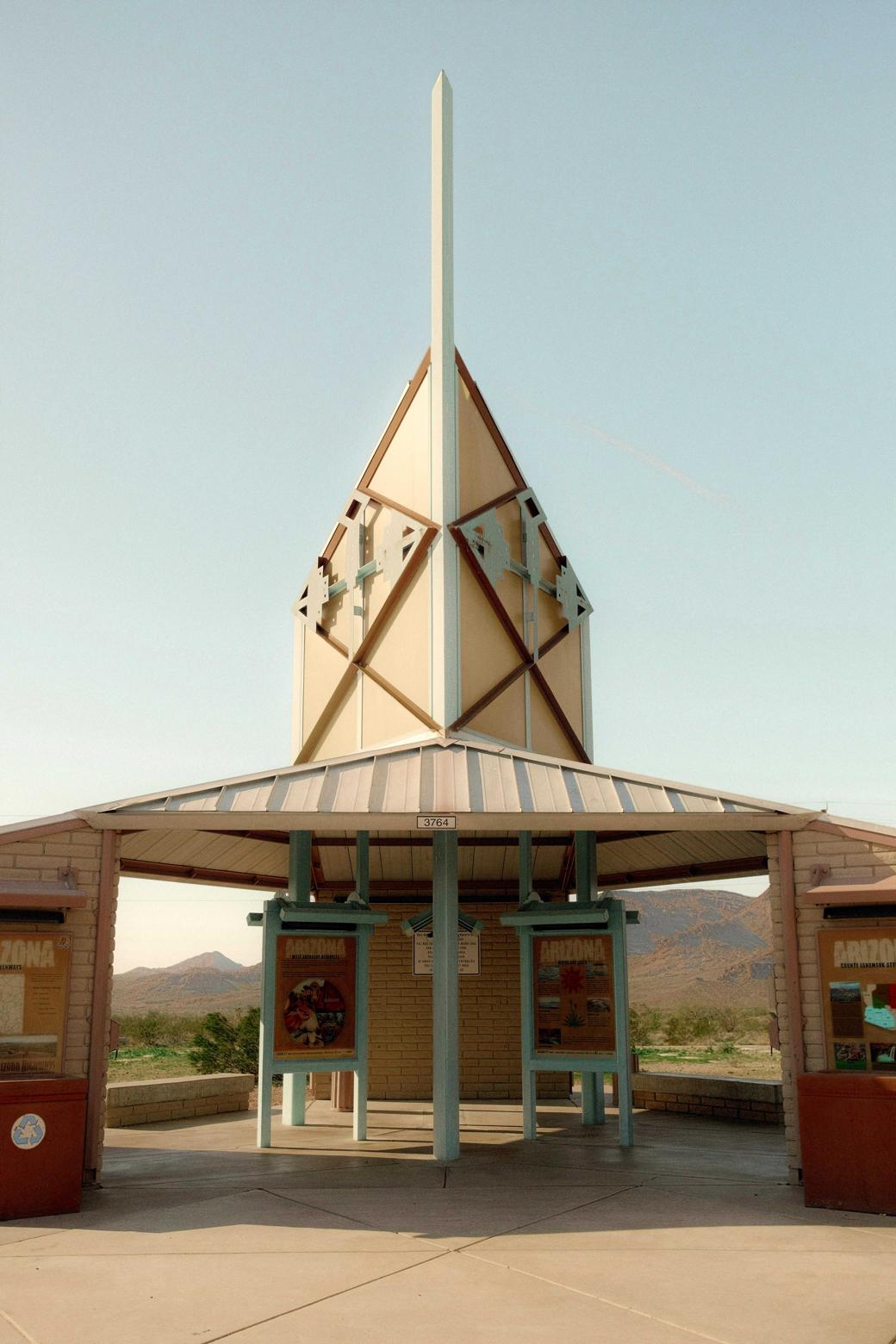} &
    \includegraphics[width=\linewidth]{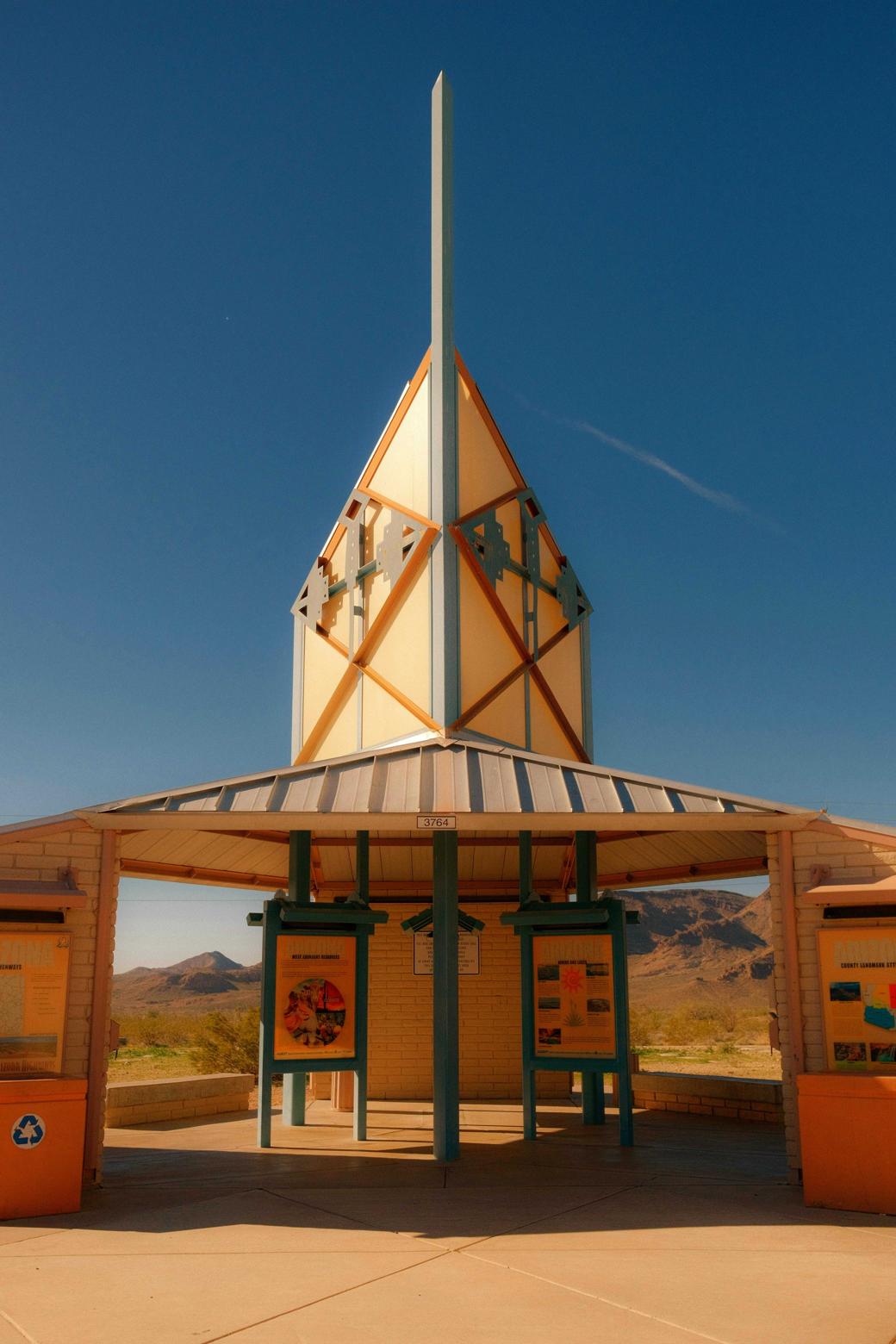} &
    \includegraphics[width=\linewidth]{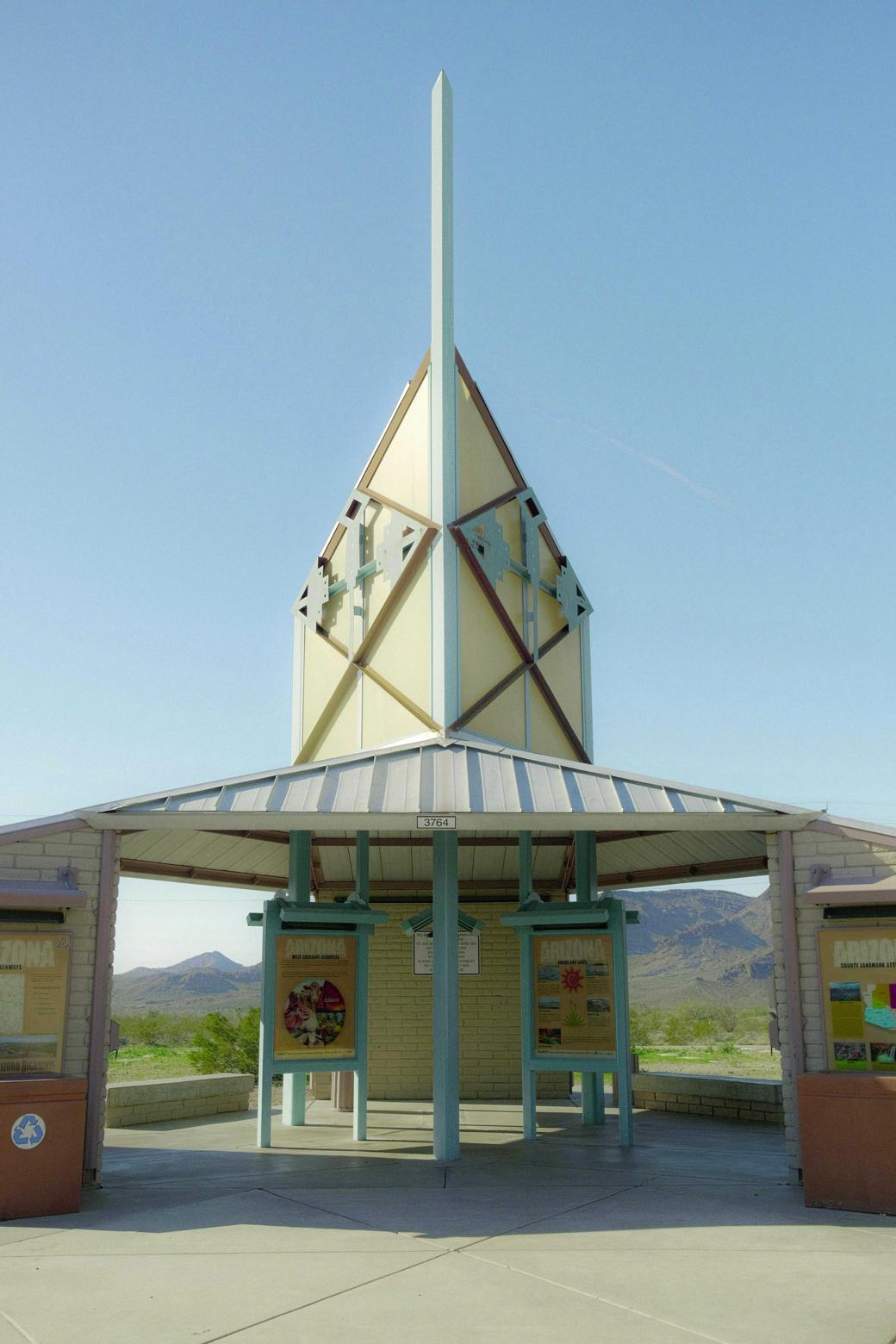} &
    \includegraphics[width=\linewidth]{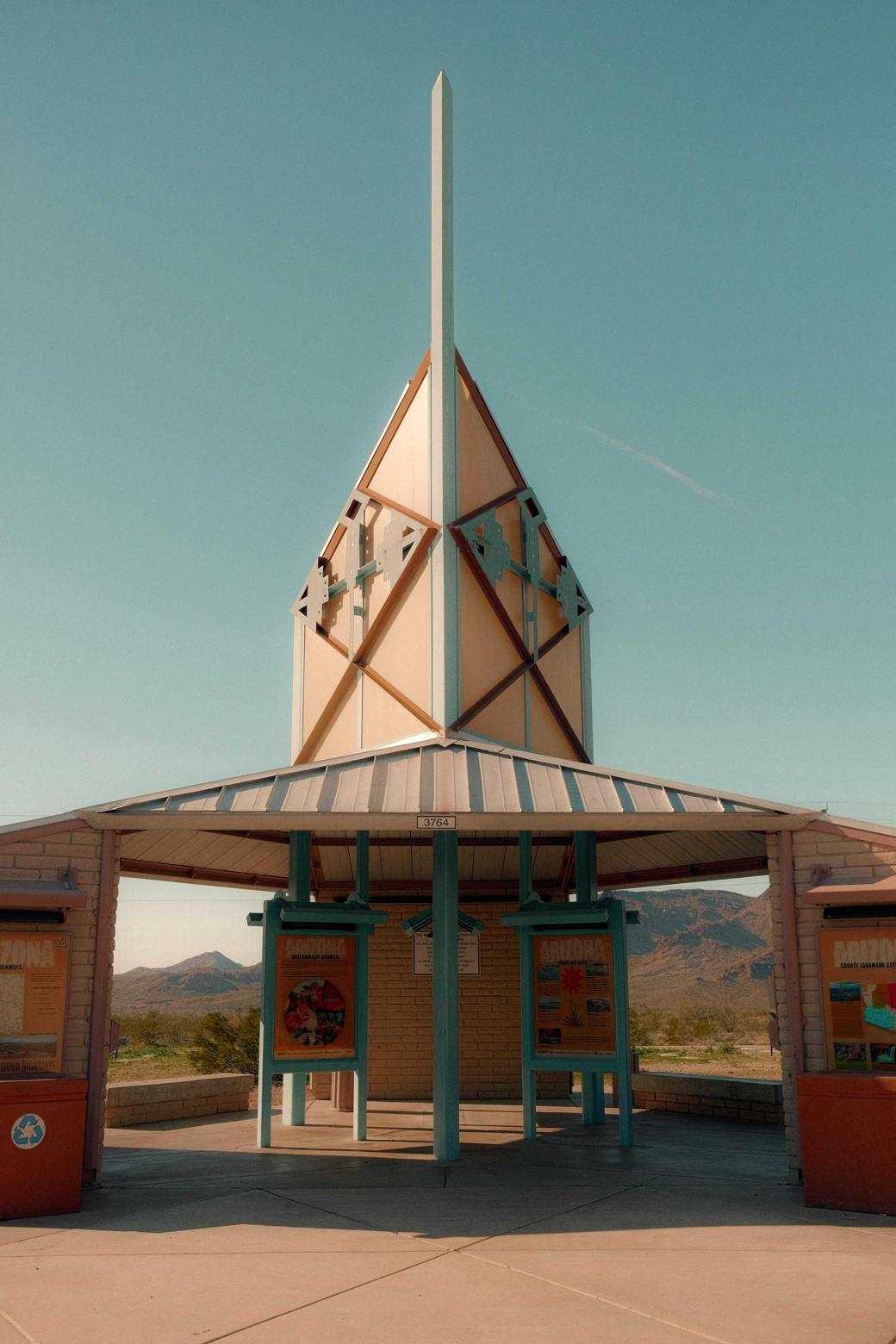}
    \\
    \scriptsize  Content & 
    \multicolumn{4}{c}{\scriptsize Results of ColorFM-L on different styles }
  \end{tabularx}
  \caption{\textbf{Color Transfer Results of ColorFM-L.} ColorFM-L enables precise color transfer across diverse styles while preserving structural fidelity and semantic consistency.}
  \label{fig:show}
\end{figure*}

Beyond these specific limitations, a more fundamental issue lies in the isolated perspective of existing approaches, in which online and offline paradigms are treated as disjoint. In this paper, we bridge this divide by proposing ColorFM, an optimization-to-learning framework that reformulates color transfer as transporting probability distributions via Flow Matching (FM)~\cite{lipman2022flow,liu2022flow,albergo2022building}. 
This formulation provides a unified view of instance-specific optimization (ColorFM-O) and feed-forward inference (ColorFM-L), while enabling consistency to be transferred from optimization to learning. As shown in \cref{fig:show}, ColorFM-L achieves precise color transfer across diverse styles while retaining structural details and semantic alignment.

In summary, our contributions include:

(1) We propose ColorFM, an optimization-to-learning framework that formulates color transfer as the transport of pixel distributions in color space via Flow Matching. This framework seamlessly integrates precise instance-specific optimization with efficient feed-forward inference.

(2) We introduce ColorFM-O, an instance-specific optimization scheme. By optimizing the velocity fields with explicit semantic alignment and hierarchical color coupling, it enables precise color transfer. The resulting outputs serve as large-scale pseudo-supervised pairs for offline learning.

(3) We design ColorFM-L, a feed-forward model trained on the generated pseudo pairs. By predicting flow parameters through implicit state modeling, it executes efficient color transfer via bidirectional linearized transport. Experiments demonstrate that ColorFM-L achieves a superior balance between style and content, outperforming state-of-the-art methods in overall visual quality.

\section{Related Work}

\subsection{Color Transfer}
Color transfer, or photorealistic style transfer, is a technique designed to transfer the color style of a reference image to a content image. Distinct from style transfer tasks \cite{huang2017arbitrary,gatys2016image,li2017universal,park2019arbitrary ,wang2024instantstyle,deng2022stytr2,ye2025stylemaster}, color transfer requires preserving the original structure while maintaining photorealistic quality. Generally, existing approaches can be categorized into two main streams: online optimization and offline inference.

\noindent\textbf{Online Optimization.}
Optimization-based methods typically require iterative tuning for each input pair. Early works~\cite{pitie2007automated,pitie2007linear,pitie2005n} rely on linear mappings under rigid Gaussian assumptions, which fail to capture complex, multi-modal distributions. In the deep learning era, DPST~\cite{luan2017deep} formulates the transfer objective by integrating a Matting Laplacian regularization to optimize the content image. However, its iterative optimization is computationally intensive, which can take several minutes. Similarly, NLUT \cite{chen2023nlut} fine-tunes a network to predict image-adaptive LUTs~\cite{zeng2020learning} at test time, often introducing noticeable artifacts. More recently, D-LUT~\cite{li2025d} utilizes score matching~\cite{song2019generative} to generate LUTs derived solely from the style image, thereby overlooking the content information. Regardless of their individual constraints, a common shortcoming across these methods is their inability to achieve precise, high-fidelity color transfer in complex scenarios. To address these limitations, we present ColorFM-O, which constructs velocity fields to enable precise and semantically aligned transfer. 
 
 \noindent\textbf{Offline Inference.}
In contrast to online optimization approaches, offline inference methods \cite{li2018closed,yoo2019photorealistic,chiu2022photowct2,ke2023neural,gong2025sa,larchenko2025color,wen2023cap,xia2020joint,lin2023adacm,ho2021deep,an2020ultrafast,hong2021domain } enable efficient color transfer. Early representative works \cite{li2018closed,yoo2019photorealistic,chiu2022photowct2,hong2021domain,an2020ultrafast, wen2023cap} employ Whitening and Coloring Transforms (WCT) for feature alignment. However, they are computationally prohibitive for high-resolution inputs and prone to visual artifacts. 
Unlike methods that rely on deep feature statistics, Neural-Preset \cite{ke2023neural} leverages a large collection of LUTs for self-supervised learning. However, even an extensive set of LUTs struggles to encompass the vast diversity of real-world color distributions, leading to limited stylistic expressiveness. 
Similarly, SA-LUT \cite{gong2025sa} generates synthetic training pairs by applying LUTs in the LOG color space. 
However, this strategy relies on an ideal alignment assumption that often fails with real-world chromatic discrepancies, leading to imperfect supervision and inconsistent stylization. Most recently, ModFlows~\cite{larchenko2025color} applies FM via a shared uniform intermediate color distribution. This indirect mapping severs semantic correspondence and restricts transfer to global means, often yielding visual artifacts and color banding. Furthermore, its reliance on numerical ODE integration precludes efficient one-step inference. To overcome these limitations, we propose ColorFM-L, which predicts flow parameters to enable rapid and precise color transfer while preserving structural integrity.

\subsection{Flow Matching}
Flow Matching (FM)~\cite{lipman2022flow,liu2022flow,albergo2022building} is a framework that models the continuous transport map between two distributions. Formally, it transports source samples to a target distribution by solving an ODE defined by a learned velocity field. Since sampling requires numerical integration, inference efficiency is strongly influenced by the curvature of the flow trajectories.

Ideally, straight trajectories enable efficient one-step or few-step inference. To achieve this,  Rectified Flow~\cite{liu2022flow, liu2023instaflow}  straightens trajectories through iterative self-distillation. While effective, this approach is computationally expensive due to the necessity of multiple retraining rounds.
Alternatively, Optimal Transport (OT) theoretically provides an ideal coupling for minimizing transport cost and straightening trajectories. However, computing exact OT plans scales cubically, making it intractable for dense distribution matching tasks. Mini-batch approximations like OT-CFM~\cite{tong2023improving, pooladian2023multisample} have been proposed to mitigate this cost, but they struggle to capture global distribution statistics, leading to suboptimal mappings~\cite{tong2023improving,fatras2019learning} (\eg, color desaturation).
In contrast, our hierarchical color coupling strategy effectively constructs a globally consistent coupling that yields quasi-linear flow trajectories for efficient inference.

\begin{figure}[tb]
  \centering
  \begin{overpic}[width=\linewidth]{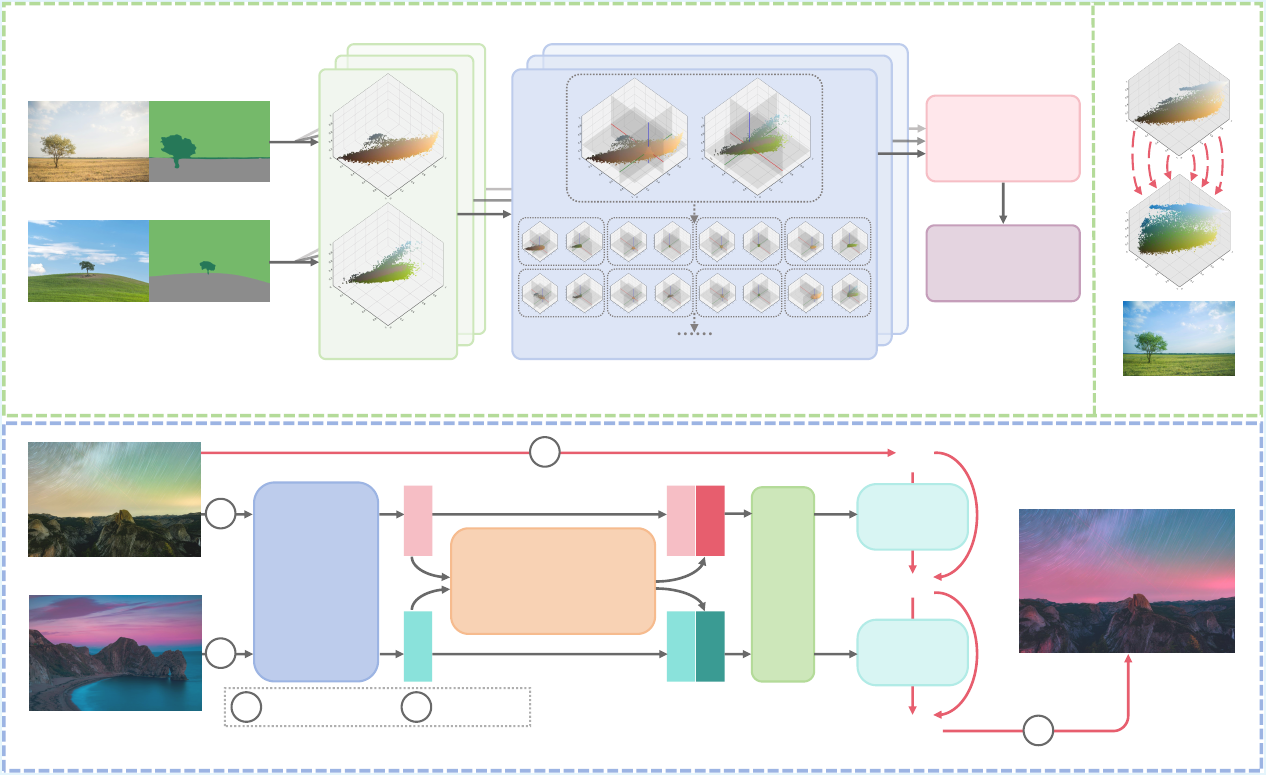}
\put(50,28.7){\makebox(0,0)[b]{\scriptsize (a) ColorFM-O\textcolor{gray}{ptimization}}}
\put(50,0.5){\makebox(0,0)[b]{\scriptsize (b) ColorFM-L\textcolor{gray}{earning}}}
 \put(7,44.5){\makebox(0,0)[b]{\scriptsize $I_c$}}
 \put(16.3,44.5){\makebox(0,0)[b]{\scriptsize $M_c$}}
  \put(7,35){\makebox(0,0)[b]{\scriptsize $I_s$}}
  \put(16.3,35){\makebox(0,0)[b]{\scriptsize $M_s$}}
\put(30.6,33.2){\makebox(0,0)[b]{\scalebox{0.5}{$(\pi_{content}^{grass}, \pi_{style}^{grass})$}}}

\put(55,33.2){\makebox(0,0)[b]{\scalebox{0.5}{leaf-node random pairing}}}

\put(79.2,51.4){\makebox(0,0)[b]{\scalebox{0.5}{Construct Dataset}}}
\put(79.2,48.4){\makebox(0,0)[b]{\scalebox{0.55}{$D = \{(x_0^j, x_1^j)\}_{j=1}^M$}}}

\put(79.2,41){\makebox(0,0)[b]{\scalebox{0.5}{Optimize $v_\theta$}}}

\put(79.2,38.5){\makebox(0,0)[b]{\scalebox{0.7}{$\mathcal{L}_{FM}$}}}

\put(31.6,58.3){\makebox(0,0)[b]{\scalebox{0.5}{Explicit Semantic Alignment}}}
\put(55.1,58.3){\makebox(0,0)[b]{\scalebox{0.5}{Hierarchical Color Coupling}}}
\put(93.1,58.3){\makebox(0,0)[b]{\scalebox{0.5}{ODE inference via $v_\theta$}}}
\put(98,50){\makebox(0,0)[b]{\scalebox{0.4}{$t = 0$}}}
\put(98,39.7){\makebox(0,0)[b]{\scalebox{0.4}{$t = 1$}}}
 \put(93.3,29.7){\makebox(0,0)[b]{\scalebox{0.6}{$I_{out}$}}}

   \put(9,14.9){\makebox(0,0)[b]{\scriptsize $I_c$}}
   \put(9,2.7){\makebox(0,0)[b]{\scriptsize $I_s$}}
   \put(25,13.2){\makebox(0,0)[b]{\scriptsize \shortstack[c]{Feature \\ Encoder}}}
   \put(43.6,13.2){\makebox(0,0)[b]{\scriptsize \shortstack[c]{Cross-Attention \\ Block}}}
   \put(33,19.4){\makebox(0,0)[b]{\scriptsize $f_c$}}
    \put(53.8,19.4){\makebox(0,0)[b]{\scriptsize $f_c$}}
   \put(56.3,19.4){\makebox(0,0)[b]{\scriptsize $\hat{f}_c$}}
   \put(33,8.7){\makebox(0,0)[b]{\scriptsize $f_s$}}
   \put(53.8,8.7){\makebox(0,0)[b]{\scriptsize $f_s$}}
    \put(56.3,8.7){\makebox(0,0)[b]{\scriptsize $\hat{f}_s$}}

    \put(72,19.6){\makebox(0,0)[b]{\scalebox{0.6}{ $v(m_c, 0; \Theta_c)$}}}
     \put(72.4,24.6){\makebox(0,0)[b]{\scalebox{0.8}{$m_c$}}}
     \put(72.1,14.6){\makebox(0,0)[b]{\scriptsize $z$}}
     \put(72,3.3){\makebox(0,0)[b]{\scalebox{0.8}{$m_{out}$}}}

     \put(91.2, 7.1){\makebox(0,0)[b]{\scalebox{0.8}{$I_{out}$}}}
   % \put(17.4,19.75){\makebox(0,0)[b]{\scalebox{0.6}{$\bm{\downarrow}$}}}
   \put(17.4,19.9){\makebox(0,0)[b]{\scalebox{0.6}{\textcolor[HTML]{595959}{$\bm{\downarrow}$}}}}
   \put(17.4,8.8){\makebox(0,0)[b]{\scalebox{0.6}{\textcolor[HTML]{595959}{$\bm{\downarrow}$}}}}
   \put(25.4,4.5){\makebox(0,0)[b]{\scalebox{0.6}{:Downsample}}}
   \put(37.4,4.5){\makebox(0,0)[b]{\scalebox{0.6}
   {:Reshape}}}
   \put(19.45,4.6){\makebox(0,0)[b]{\scalebox{0.6}{\textcolor[HTML]{595959}{$\bm{\downarrow}$}}}}
   \put(32.9,4.8){\makebox(0,0)[b]{\scalebox{0.6}
   {\textcolor[HTML]{595959}{$\mathrm{R}$}}}}
   \put(43.1,25){\makebox(0,0)[b]{\scalebox{0.6}{\textcolor[HTML]{595959}{$\mathrm{R}$}}}}
   \put(82,3){\makebox(0,0)[b]{\scalebox{0.6}{\textcolor[HTML]{595959}{$\mathrm{R}$}}}}
    \put(72,8.8){\makebox(0,0)[b]{\scalebox{0.6}{ $v(z, 1; \Theta_s)$}}}
    \put(78.4,16.5){\makebox(0,0)[b]{\scalebox{0.4}{forward}}}
    \put(78.3,12.7){\makebox(0,0)[b]{\scalebox{0.4}{reverse}}}
   \begin{turn}{90}
    \put(14.7,-63.5){\makebox(0,0)[b]{\scriptsize \shortstack[c]{Param \\ Generator}}}
   \end{turn}
   
  \end{overpic}
  \caption{
  \textbf{Overview of the ColorFM Framework.}
  Our framework consists of two components: (a) \textbf{ColorFM-O} first matches pixel distributions across semantic regions using $M_c$ and $M_s$, and then applies a hierarchical color coupling strategy to each pair. This process constructs the training dataset $D$ to learn the velocity field $v_\theta$, enabling precise color transfer via ODE integration.
(b) \textbf{ColorFM-L} extracts features from resized $I_c$ and $I_s$ to predict parameters $\Theta_c$ and $\Theta_s$. It then treats the flattened content $I_c$ as a sequence and performs bidirectional one-step flow matching operations with the predicted parameters to generate the stylized output $I_{out}$.}
  \label{fig:ColorFM}
\end{figure}
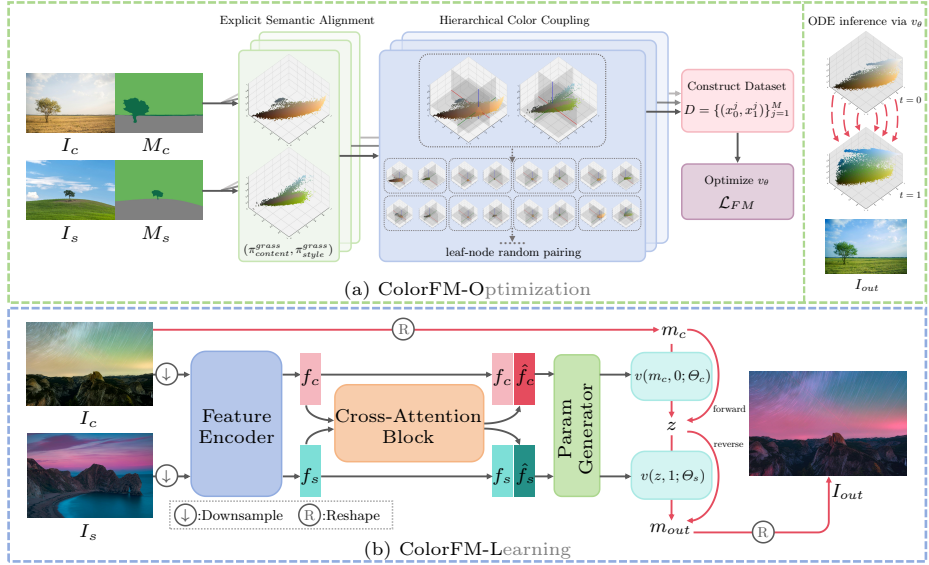

\section{ColorFM}
\subsection{Preliminaries: Flow Matching}
FM~\cite{lipman2022flow,liu2022flow,albergo2022building } learns a continuous flow that transports samples from a source distribution $\pi_{0}$ to a target $\pi_{1}$. Given a source-target pair $(x_0, x_1)$ sampled from a joint coupling and a time step $t \sim \mathcal{U}[0, 1]$, the intermediate state $x_t$ follows a linear interpolation trajectory:
\begin{align}
    x_t = (1-t) \cdot x_0 + t \cdot x_1,
\end{align}
the conditional velocity field $v_{\theta}(t, x)$, parameterized by a neural network, is then optimized to match the conditional vector field $u_t(x|x_0, x_1) = x_1 - x_0$. The FM objective is therefore:
\begin{align}
\label{eq:fm_loss}
    \mathcal{L}_{\textit{FM}} = \mathbb{E}_{t, x_0, x_1} \left[ \| v_\theta(t, x_t) - (x_1 - x_0) \|^2 \right].
\end{align}
At inference, transport is performed by solving the ODE $dx/dt = v_\theta(t, x)$ starting from $x_0$, yielding the transported sample at $t=1$.

\renewcommand{\algorithmicrequire}{\textbf{Input:}}
\renewcommand{\algorithmicensure}{\textbf{Output:}}
\begin{algorithm}[t] 
\footnotesize
\caption{Hierarchical Color Coupling}
\label{alg:recursive_partition}
\begin{algorithmic}[1]
    \Require Source $\mathcal{X}_0$, Target $\mathcal{X}_1$, depth $d$, max depth $D_{max}$
    \Ensure Coupled pairs $\mathcal{P} = \{(x_0^{(i)}, x_1^{(i)})\}$
    \Function{RecursivePartition}{$\mathcal{X}_0, \mathcal{X}_1, d, D_{max}$}
        \If{$d = D_{max}$ \textbf{or} $\min(|\mathcal{X}_0|, |\mathcal{X}_1|) = 0$} \Comment{\textbf{Base Case}}
            \State \Return $\min(|\mathcal{X}_0|, |\mathcal{X}_1|)$ random pairs from $\mathcal{X}_0$ and $\mathcal{X}_1$
        \EndIf
        
        \State $\mu_i \gets \text{Mean}(\mathcal{X}_i)$ for $i \in \{0, 1\}$; $\mathcal{P} \gets \emptyset$ \Comment{\textbf{Partition}}
        \For{$k \in \{0, \dots, 7\}$} 
            \State $\mathcal{S}_i \gets \{ x \in \mathcal{X}_i \mid \mathcal{O}(x - \mu_i) = k \}$ for $i \in \{0, 1\}$ \Comment{$\mathcal{O}(v)$: octant index of $v$ }
            \State $\mathcal{P} \gets \mathcal{P} \cup \Call{RecursivePartition}{\mathcal{S}_0, \mathcal{S}_1, d+1, D_{max}}$
        \EndFor
        \If{$\mathcal{P} = \emptyset$} \Comment{\textbf{Fallback Case}}
            \State \Return $\min(|\mathcal{X}_0|, |\mathcal{X}_1|)$ random pairs from $\mathcal{X}_0$ and $\mathcal{X}_1$
        \EndIf
        \State \Return $\mathcal{P}$
    \EndFunction
\end{algorithmic}
\end{algorithm}

\subsection{Problem Formulation}
We reformulate color transfer within the FM framework by learning an RGB-space velocity field from source-target color pairs. Given the content distribution $\pi_{content}$ and the style distribution $\pi_{style}$, the velocity field $v_t$ evolves a sampled content pixel $x_0=(r_0,g_0,b_0) \sim \pi_{content}$ toward a target color $x_1 \sim \pi_{style}$.

Guided by this formulation, the core objective of ColorFM is to estimate the velocity field $v_t$. ColorFM-O learns the field via iterative optimization for given image pairs, while ColorFM-L functions as an offline model that learns to predict the flow parameters under supervision.

\subsection{ColorFM-O}
As illustrated in \cref{fig:ColorFM}(a),
ColorFM-O aims to learn high-fidelity color transport trajectories by directly optimizing the velocity field via FM. To facilitate this, we construct the source-target pairs through explicit semantic alignment and hierarchical color coupling.

\noindent\textbf{Explicit Semantic Alignment.}
Prior approaches~\cite{luan2017deep, yoo2019photorealistic,chiu2022photowct2, wen2023cap} utilizing semantic masks typically process each region independently to ensure semantic alignment. However, this independent treatment often disrupts spatial continuity, resulting in visible seam artifacts or halo effects along object boundaries. 

To address this issue, we integrate semantic correspondence into a single unified velocity field, thereby reconciling region-aware stylization with global consistency. Specifically, given semantic masks $M_c$ and $M_s$ generated by a pre-trained segmentation model, we extract pixel sets corresponding to each shared label $l$ to form source-target distribution pairs $(\pi_{content}^{l}, \pi_{style}^{l})$. Meanwhile, pixels belonging to unmatched or background regions are aggregated into a residual global distribution pair. Notably, if no residual style region is available , the remaining content pixels are instead paired with the global style distribution. This process yields a collection of $N$ distribution pairs $\{(\pi_{content}^i, \pi_{style}^i)\}_{i=1}^N$.  Instead of training separate models for each region, we optimize a unified velocity field $v_\theta$ to learn the transport dynamics for all pairs simultaneously.  This design implicitly imposes spatial regularization, enabling the network to achieve precise semantic alignment while maintaining seamless global coherence.

\noindent\textbf{Hierarchical Color Coupling.}
After obtaining the semantic distribution pairs $\{(\pi_{content}^i, \pi_{style}^i)\}_{i=1}^N$, the subsequent challenge lies in establishing a high-quality coupling between the source samples $\mathcal{X}_0 \sim \pi_{content}^i$ and target samples $\mathcal{X}_1 \sim \pi_{style}^i$ to construct the training set for the velocity field. Standard random matching~\cite{lipman2022flow,liu2022flow} often yields geometrically incoherent trajectories, causing severe color banding. 
However, alternative approximations like mini-batch OT~\cite{tong2023improving, pooladian2023multisample} fail to capture global distribution statistics, leading to color desaturation.
To balance quality and efficiency, we propose the hierarchical color coupling (HCC) strategy.

Our key insight is that preserving relative locality in color space suffices to eliminate banding and ensure flow coherence. 
Guided by this, HCC recursively aligns distributions in a coarse-to-fine manner (see \cref{alg:recursive_partition}). Specifically, at each hierarchy level, we center the subsets by subtracting their respective means to align their relative positions, and then partition the space into octants based on the coordinate signs. This recursive process continues until a maximum depth is reached, after which points within corresponding leaf nodes are coupled via random pairing. Finally, by independently applying this strategy to each of the $N$ semantic distribution pairs defined previously, we aggregate the coupled samples into a unified training dataset $D = \{(x_0^j, x_1^j)\}_{j=1}^M$. This dataset $D$ serves as a static, geometrically optimized coupling plan, ready for supervision.

\noindent\textbf{Velocity Field Optimization.}
 Given the constructed dataset $D$, we optimize the network $v_\theta$ using the objective in \cref{eq:fm_loss}. At inference time, stylization is performed by numerically integrating the learned ODE in a pixel-wise manner to transport the content image $I_c$ to the stylized output $I_{out}$.

 We then synthesize a diverse collection of triplets $(I_c, I_s, I_{out})$ by executing ColorFM-O on various content-style pairs. These triplets provide the essential, semantically aligned supervision for ColorFM-L.

\subsection{ColorFM-L}
Despite the high-fidelity performance of ColorFM-O, its reliance on costly online optimization and per-instance hyperparameter tuning (\eg, iteration steps) limits its scalability. 
To achieve real-time inference, we propose ColorFM-L (see \cref{fig:ColorFM}(b)).
Instead of iterative solving, ColorFM-L is trained using the pseudo-ground truth images generated by ColorFM-O.
Conceptually, ColorFM-L decouples the color transfer task into two collaborative stages: (1) predicting flow parameters via implicit state modeling, and (2) executing the color transfer via bidirectional linearized transport. We detail these components below.

\noindent\textbf{Implicit State Modeling.}
Directly predicting the transport between arbitrary content-style pairs is non-trivial due to significant domain gaps. Existing methods~\cite{ke2023neural, larchenko2025color} typically rely on a fixed, pre-defined intermediate distribution to bridge this gap. However, restricting diverse images into a rigid, hand-crafted prior often limits the model's expressivity.

To address this, we introduce an implicit state modeling scheme. Rather than adhering to a fixed prior, we hypothesize the existence of an optimal intermediate state within the content-style space that acts as a bridge. Accordingly, we employ a weight-sharing encoder $E_\phi$ to predict the specific transport parameters that drive the flow toward this state, thereby implicitly modeling it. Specifically, $E_\phi$ comprises a feature backbone $E_{\text{feat}}$, a cross-attention block $E_{\text{attn}}$, and a parameter generator $E_{\text{param}}$. The parameter prediction process is formulated as follows:
\begin{equation} 
\begin{gathered} 
f_c = E_{\text{feat}}(\tilde{I}_c), \quad f_s = E_{\text{feat}}(\tilde{I}_s), \\ 
\hat{f}_c = E_{\text{attn}}(f_c, f_s), \quad \hat{f}_s = E_{\text{attn}}(f_s, f_c), \\
\Theta_c = E_{\text{param}}([f_c, \hat{f}_c]), \quad \Theta_s = E_{\text{param}}([f_s, \hat{f}_s]). 
\end{gathered} 
\end{equation}
Here, $\tilde{I}_c$ and $\tilde{I}_s$ denote the content and style images resized to a fixed resolution, and $[\cdot, \cdot]$ represents the concatenation operation. $\hat{f}_c$ and $\hat{f}_s$ are the features generated via cross-attention using deep semantic features $f_c$ and $f_s$ as queries, respectively. 
Importantly, $\Theta_c$ and $\Theta_s$ denote the predicted weights of a globally shared, bias-free pixel-wise MLP. This MLP explicitly defines the velocity field $v(\cdot; \Theta)$ driving the flow toward the implicit state. 

\noindent\textbf{Bidirectional Linearized Transport.}
Given the parameters $\Theta$, ColorFM-L executes the color transfer process. To accelerate inference, we leverage the quasi-linearity of ColorFM-O trajectories, which is inherently induced by the HCC strategy. 
This geometric regularity allows us to approximate the flow path as a straight line, enabling efficient one-step transport without iterative integration.

Formally, we first flatten the spatial dimensions of the content image $I_c \in \mathbb{R}^{3 \times H \times W}$ to obtain a pixel matrix $m_c \in \mathbb{R}^{HW \times 3}$. The transfer process is formulated as a composition of a forward flow to the intermediate state and a backward flow to the target style domain:
\begin{equation}
\begin{aligned}
z = m_c + v(m_c, 0; \Theta_c), \
m_{out} = z - v(z, 1; \Theta_s),
\end{aligned}
\end{equation}
where $z$ denotes the intermediate implicit state. Effectively, this formulation corresponds to a bidirectional one-step Euler integration (\ie, $\Delta t = 1$). Finally, the stylized feature $m_{out}$ is reshaped back to $\mathbb{R}^{3 \times H \times W}$ to yield the final output $I_{out}$. Furthermore, we can perform the inverse color transfer by directly reusing the predicted parameters and simply reversing the flow (\ie, mapping $I_s$ forward via $\Theta_s$ and then backward via $\Theta_c$).

\noindent\textbf{Optimization Loss.}
The entire pipeline is differentiable and trained end-to-end using the following objective:
\begin{equation}
    \mathcal{L}_{total} = \|I_{out} - I_{gt}\|^2 + \lambda \cdot \mathcal{L}_{\textit{LPIPS}}(I_{out}, I_{gt}),
\end{equation}
where  $I_{gt}$ denotes the pseudo-ground truth synthesized by ColorFM-O, and $\mathcal{L}_{\textit{LPIPS}}$~\cite{zhang2018unreasonable} represents the perceptual loss. We set $\lambda = 0.1$ to balance pixel-wise accuracy with perceptual fidelity.

\section{Experiments}

\subsection{Implementation Details} 
\noindent\textbf{Datasets.} We use the optimization-based ColorFM-O to construct a large-scale dataset of 237,408 triplets $(I_c, I_s, I_{out})$. 
Source images are collected from Unsplash and DIV2K~\cite{agustsson2017ntire}. 
To ensure semantic consistency, Unsplash images are categorized into seven classes (Urban, Portrait, Nature, Indoor, Animal, Still Life, and Random) with intra-category pairing. 
In contrast, DIV2K images are randomly paired to enhance diversity.
This data composition enables ColorFM-L to effectively learn both semantic alignment and precise color transfer.

For the test dataset, we collected 40 high-quality images from Unsplash and performed exhaustive pairwise permutations to create 1,560 content-style pairs, ensuring strictly no overlap with the training set. This dataset covers diverse semantic scenes, enabling a comprehensive evaluation of model performance.

 \noindent\textbf{Network Configuration.}
For ColorFM-O, the velocity field is parameterized by a lightweight two-layer bias-free MLP with 512 hidden units and the Swish activation~\cite{ramachandran2017searching}. 
We use a pre-trained SegFormer-B5~\cite{xie2021segformer}, fine-tuned on ADE20K~\cite{zhou2017scene}, for semantic mask extraction. Hierarchical color coupling is configured with a maximum tree depth of $D_{max}=3$. 

For ColorFM-L, we adopt ViT-S/16~\cite{dosovitskiy2020image, oquab2023dinov2} as $E_{\text{feat}}$ with input images resized to $256 \times 256$. $E_{\text{attn}}$ is implemented as a cross-attention layer following the design of a standard ViT attention block. The parameter generator $E_{\text{param}}$ then applies global average pooling and multiple MLP heads to project features into $\Theta$. These parameters instantiate a globally shared pixel-wise MLP with channel dimensions $4 \!\to\! 16 \!\to\! 16 \!\to\! 16 \!\to\! 3$ and Swish activation, where the input includes the time coordinate.

\noindent\textbf{Training and Inference.}
For ColorFM-O, we use the Adam~\cite{kingma2014adam} optimizer with a learning rate of $5 \times 10^{-4}$. The optimization is run for 700 steps with a batch size of 4,096 sampled pixels. During inference, the ODE is solved using the midpoint method with 5 integration steps to ensure trajectory precision.

For ColorFM-L, we use the Adam optimizer with a batch size of 48. The model is trained for 50 epochs. The learning rate is initialized to $1 \times 10^{-5}$ and maintained for the first 20 epochs, after which it is gradually decayed to $1 \times 10^{-6}$ using a cosine annealing schedule over the remaining 30 epochs. All experiments are conducted on a single NVIDIA RTX 4090 GPU.
\begin{figure*}[t]
  \centering
  \setlength{\tabcolsep}{0.5pt} 
  \renewcommand{\arraystretch}{0.5} 
  
  \begin{tabularx}{\textwidth}{Y Y Y Y Y Y Y Y} 

    \includegraphics[width=\linewidth]{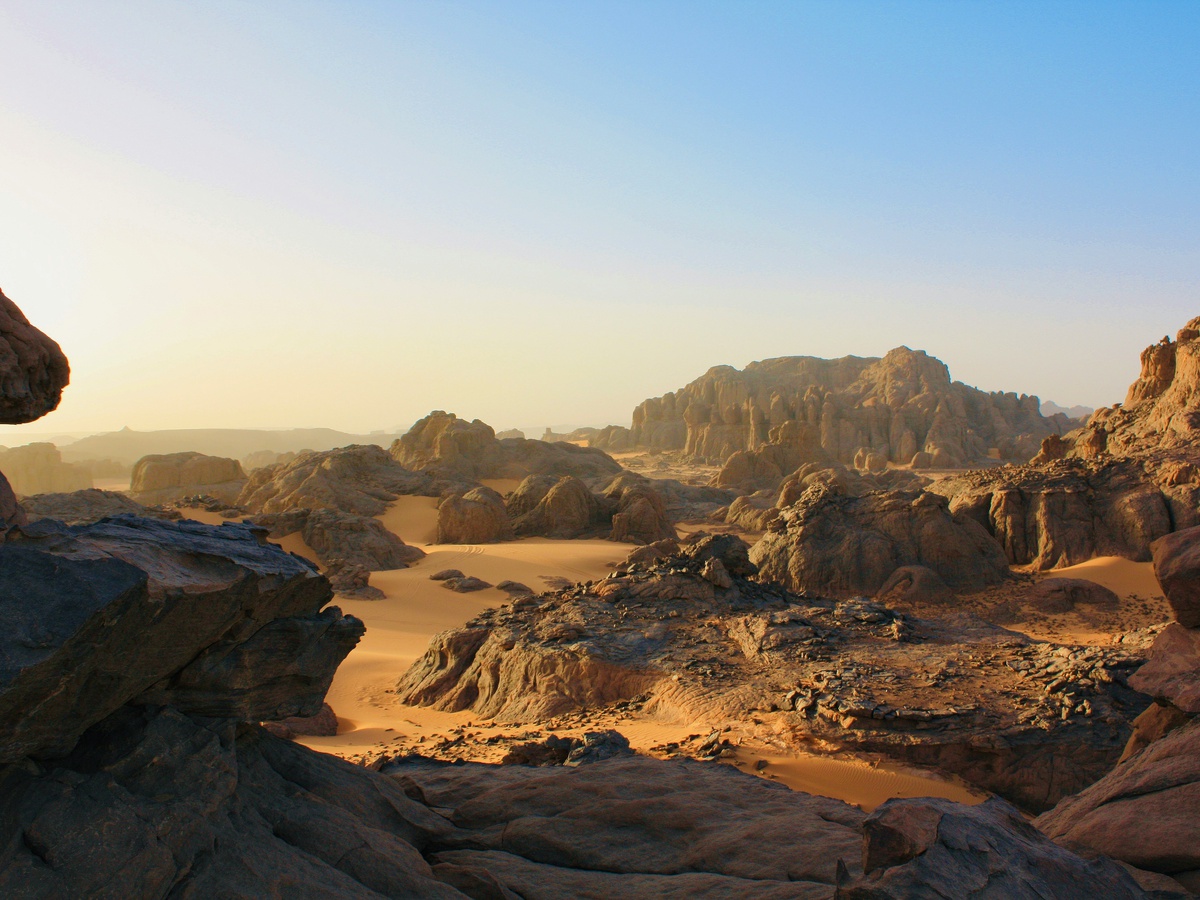} &
    \includegraphics[width=\linewidth]{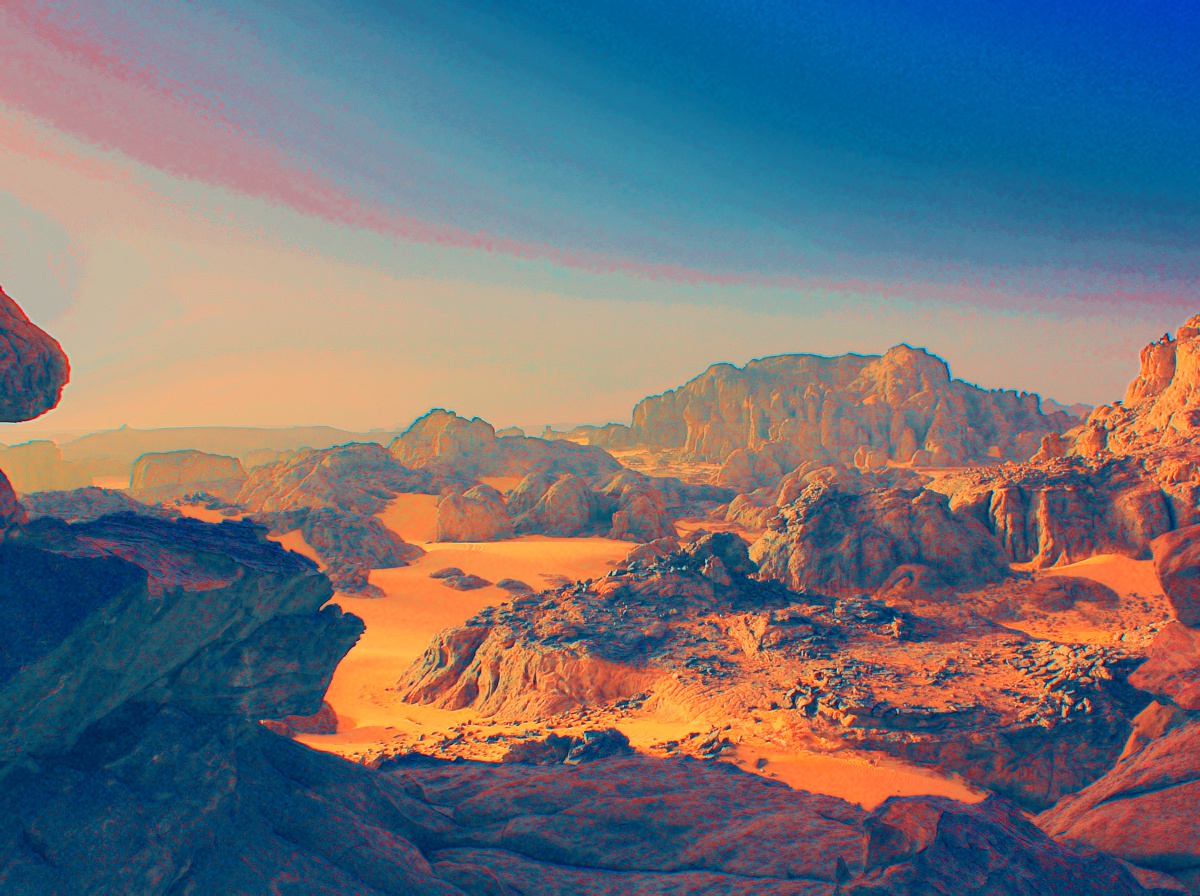} &
    \includegraphics[width=\linewidth]{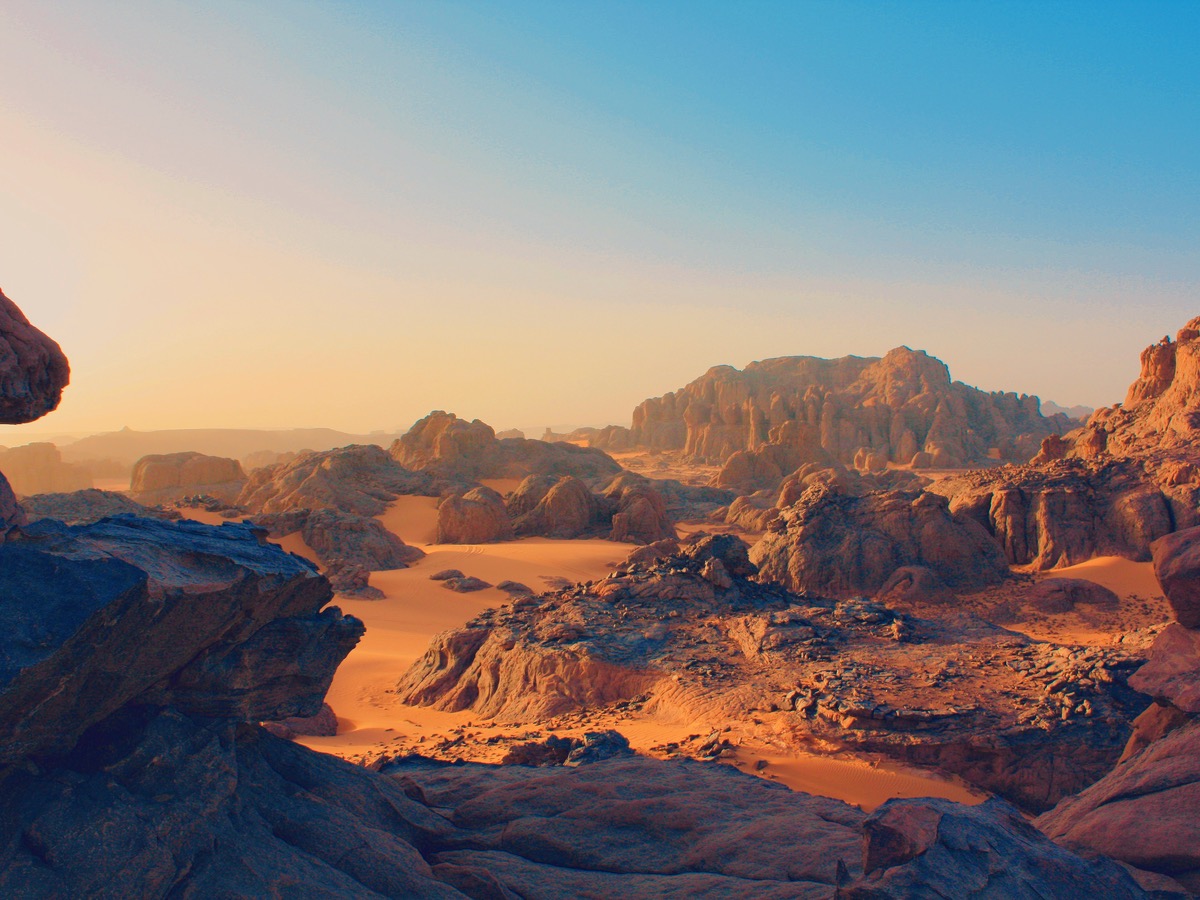} &
    \includegraphics[width=\linewidth]{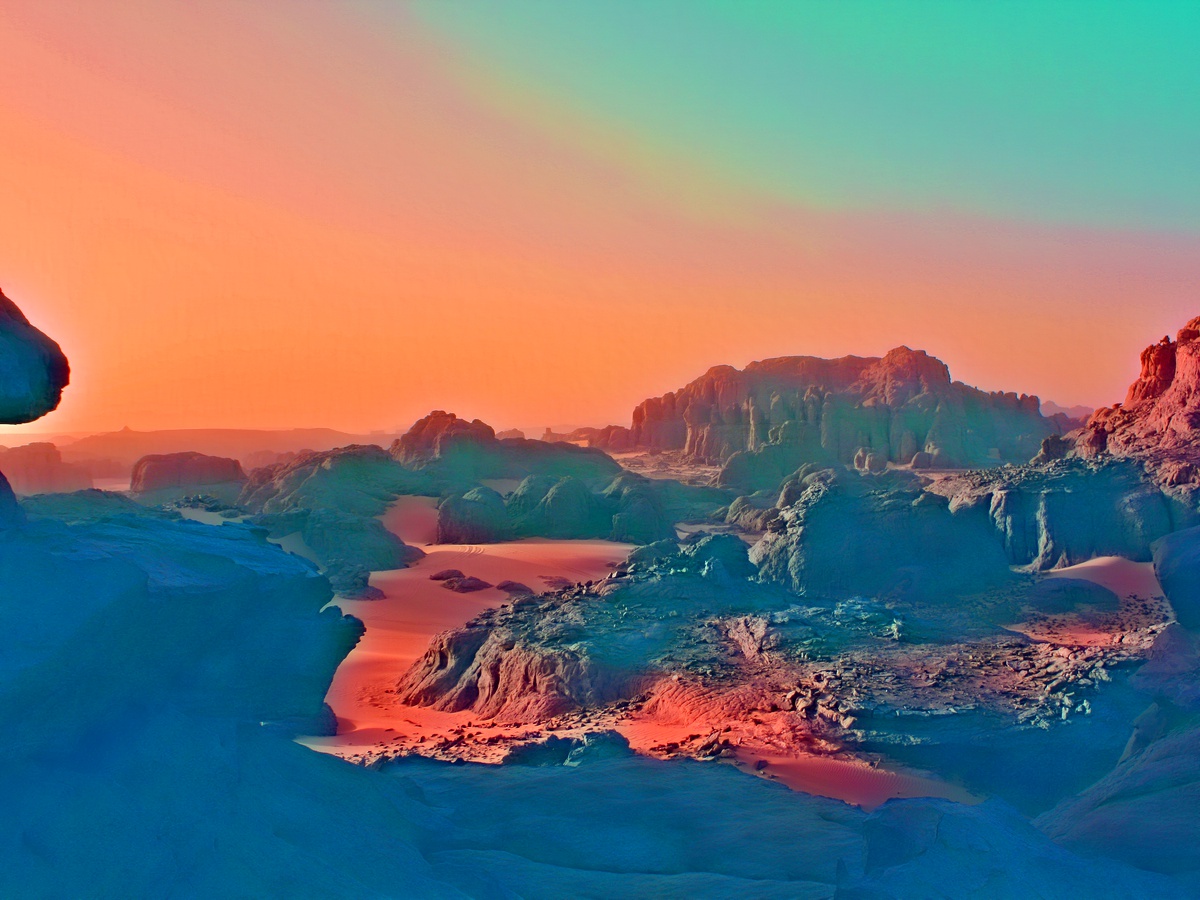} &
    \includegraphics[width=\linewidth]{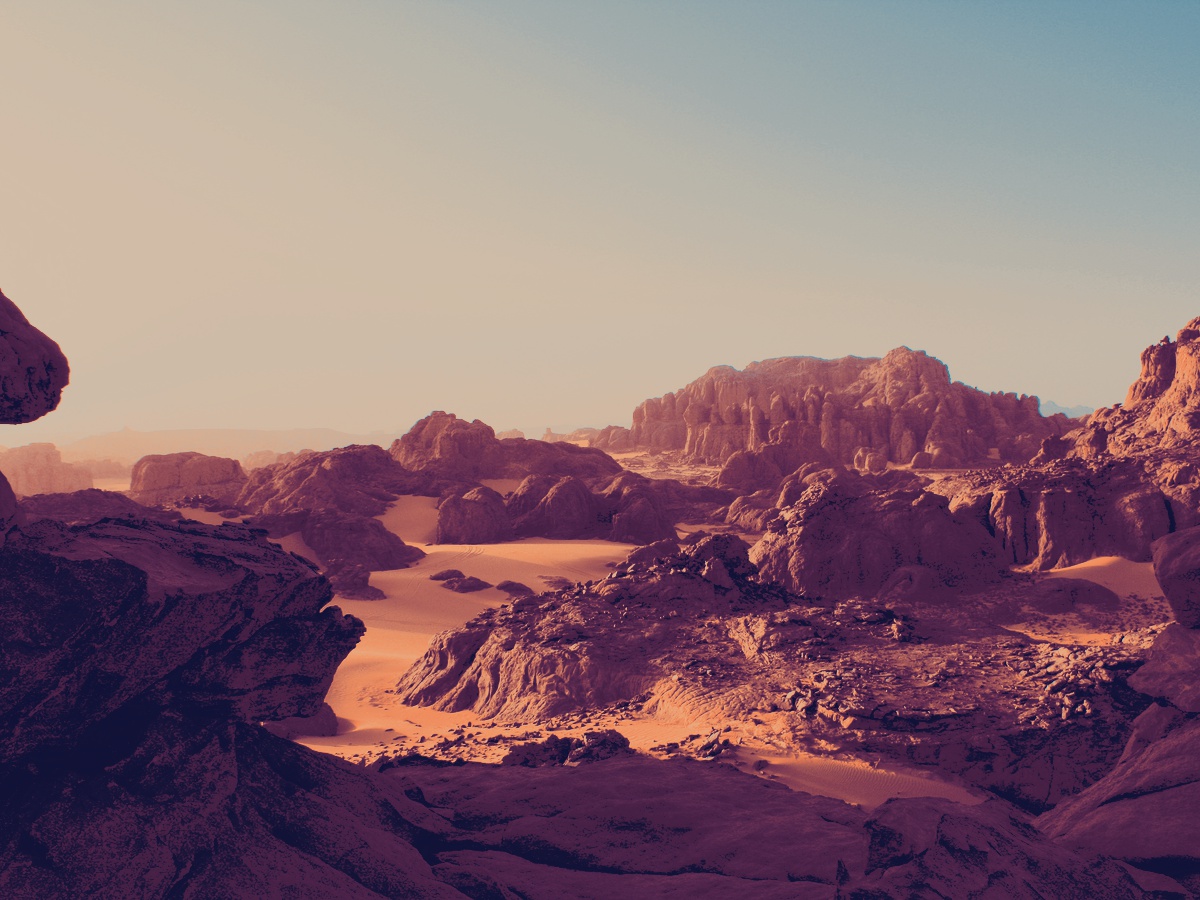} &
    \includegraphics[width=\linewidth]{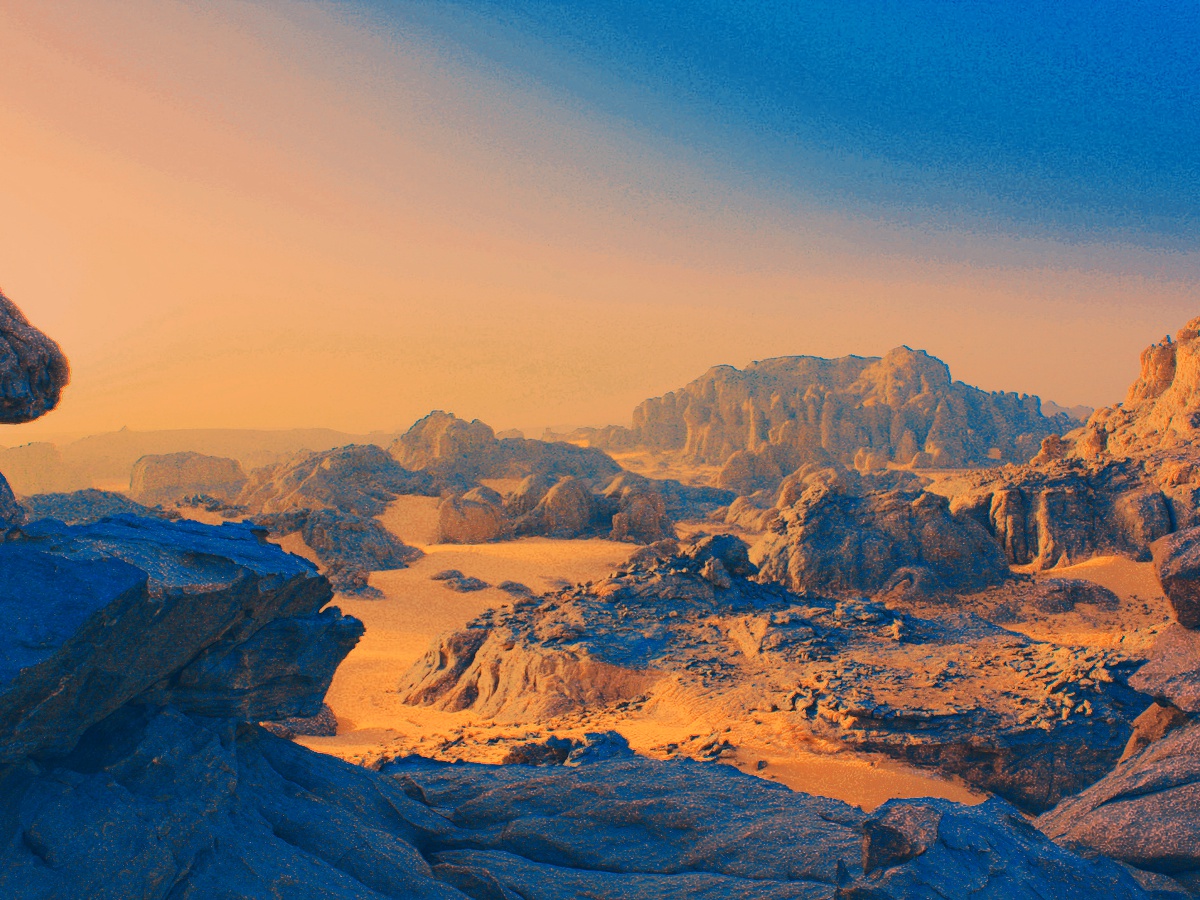} &
    \includegraphics[width=\linewidth]{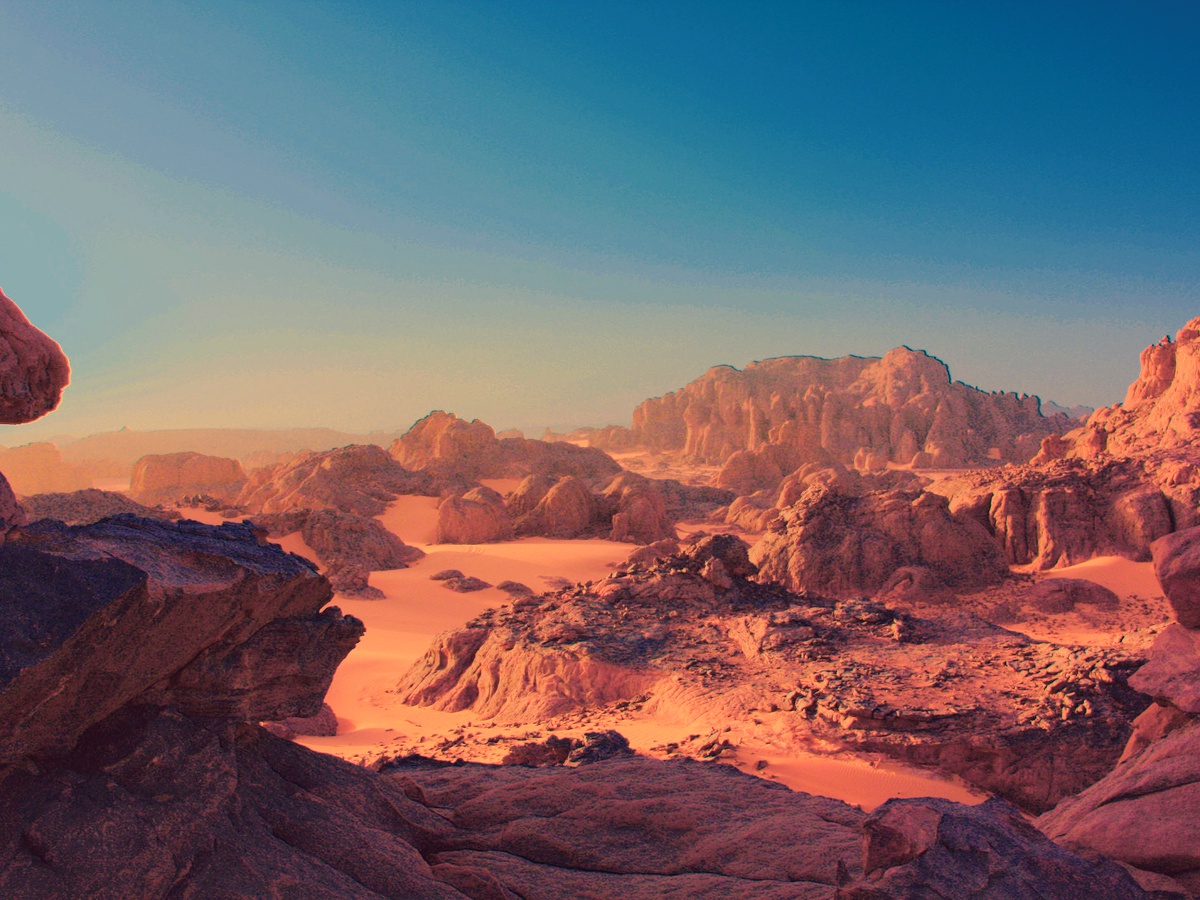}&
    \includegraphics[width=\linewidth]{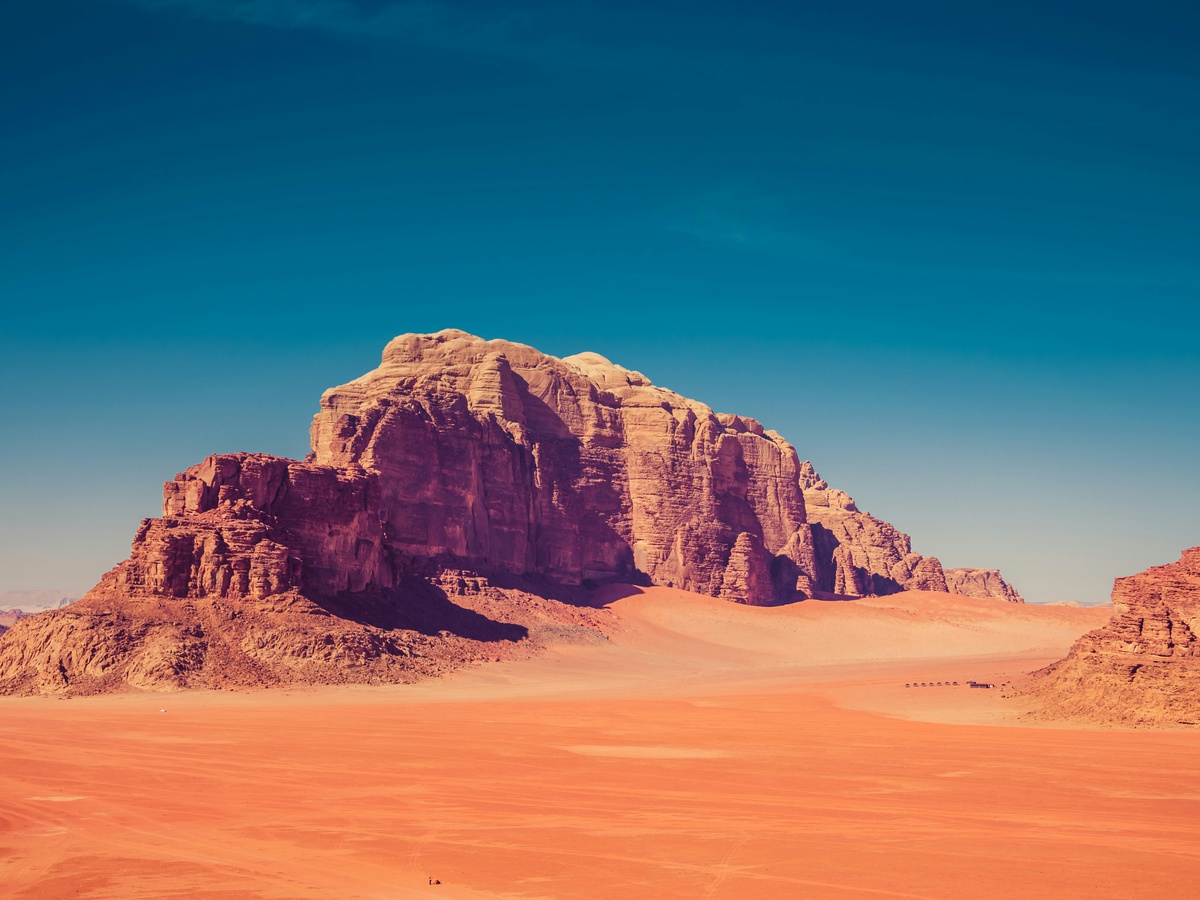}\\

            \includegraphics[width=\linewidth]{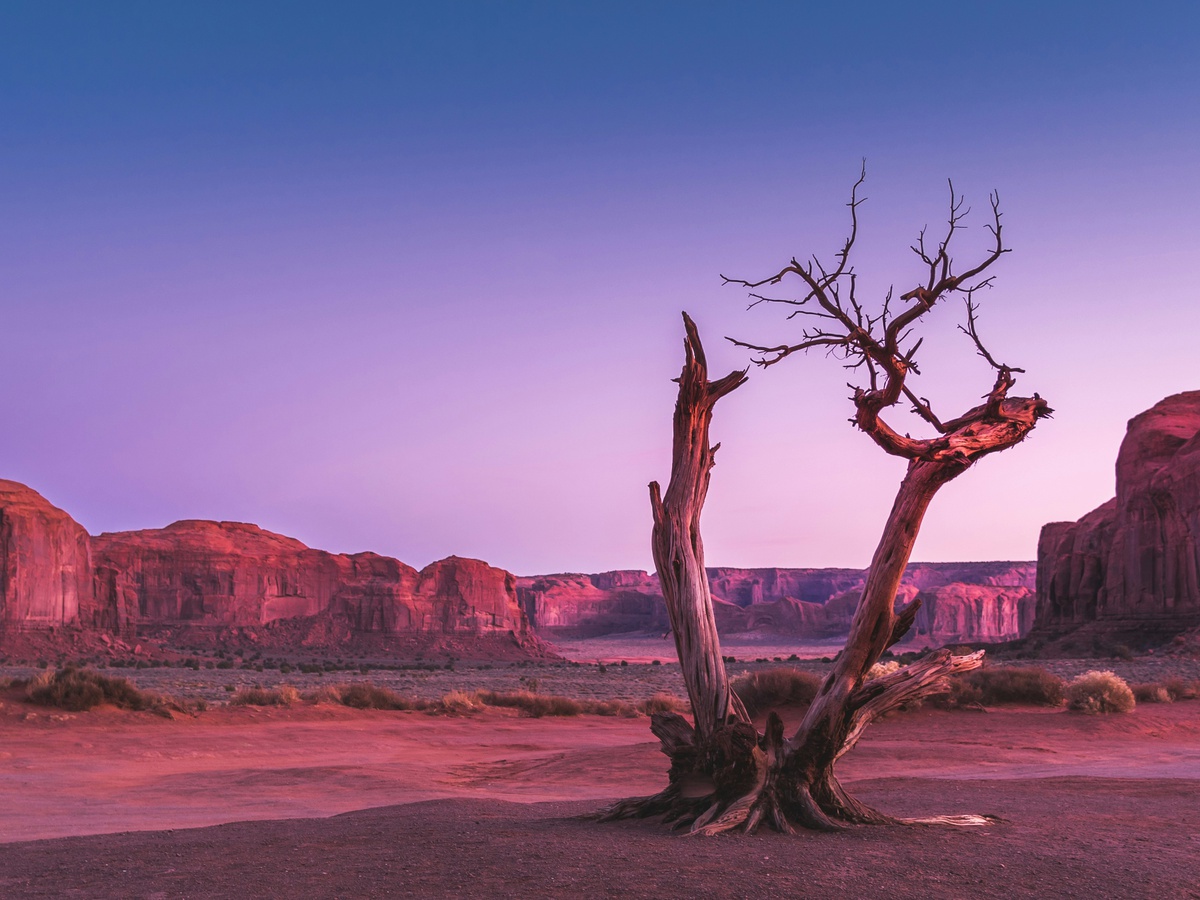} &
    \includegraphics[width=\linewidth]{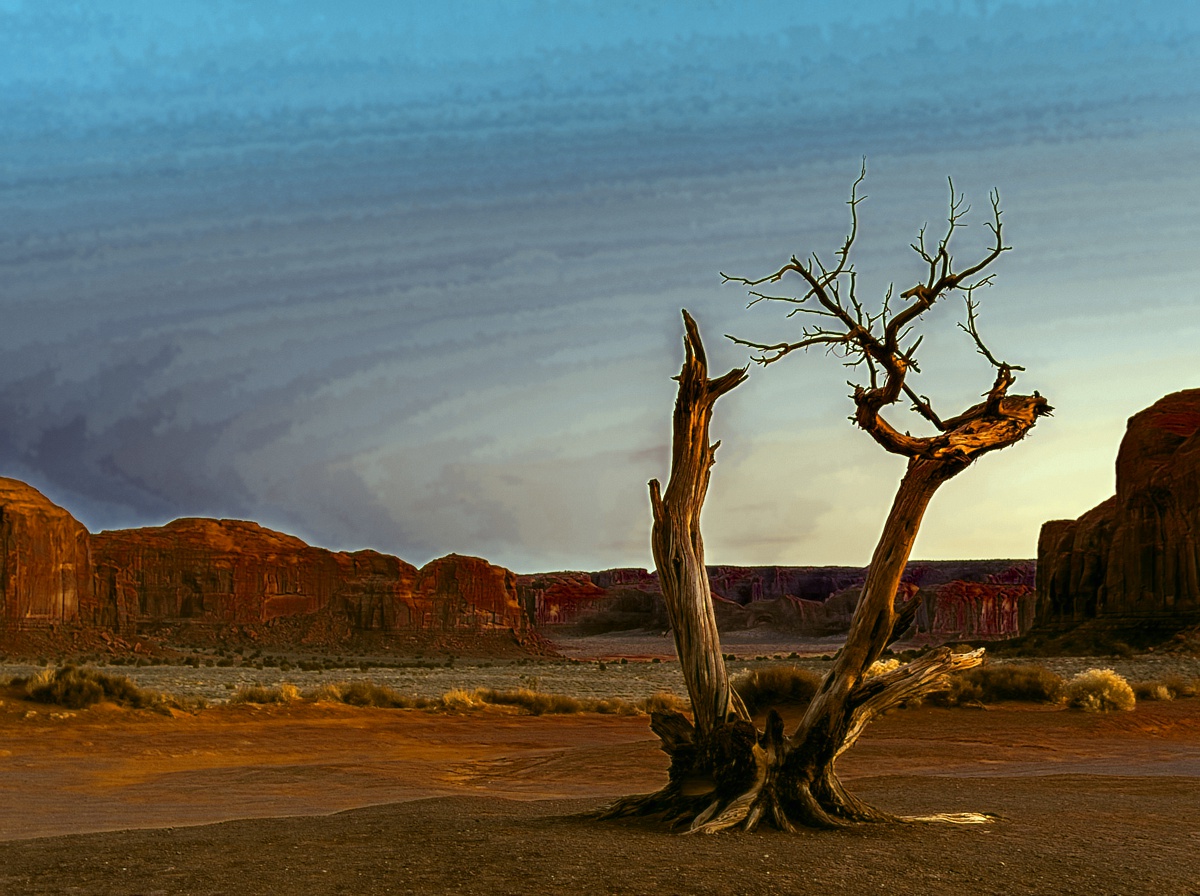} &
    \includegraphics[width=\linewidth]{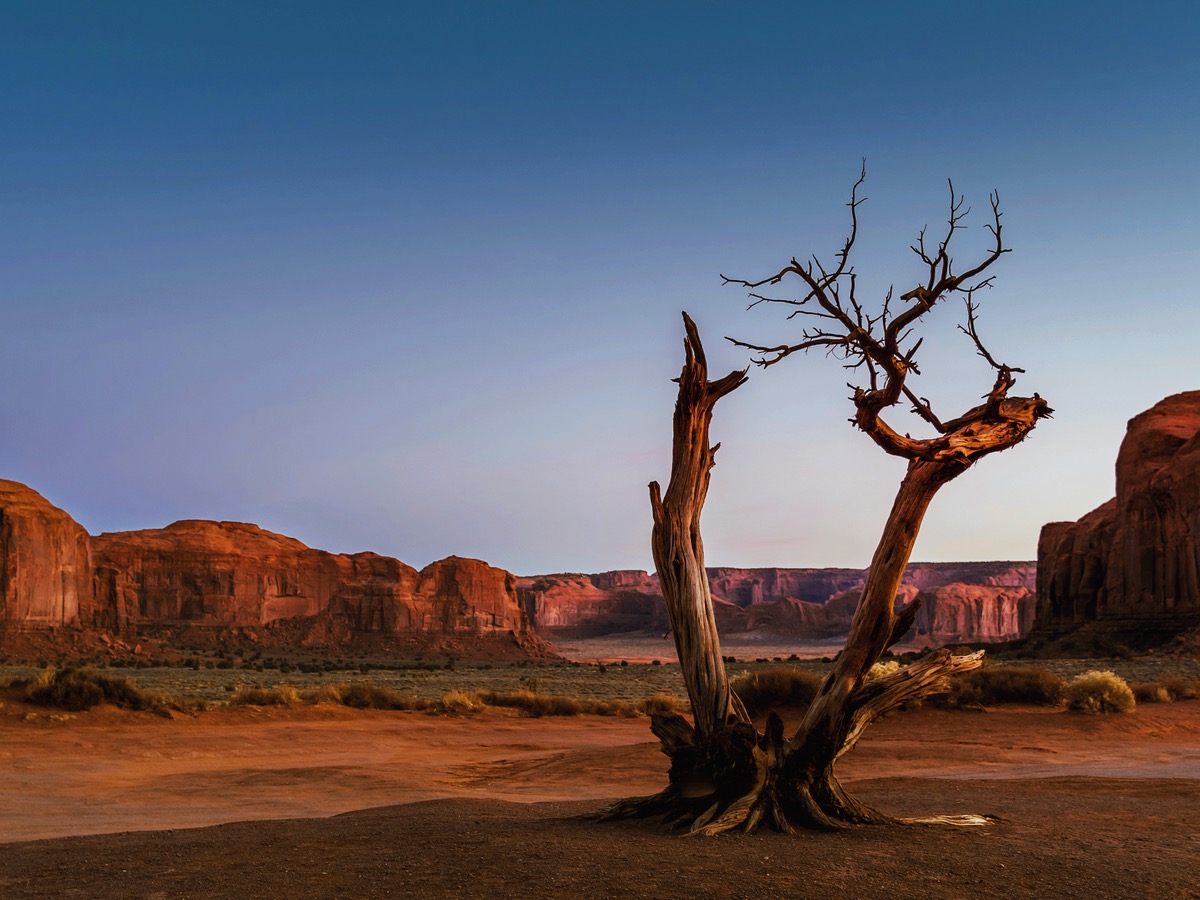} &
    \includegraphics[width=\linewidth]{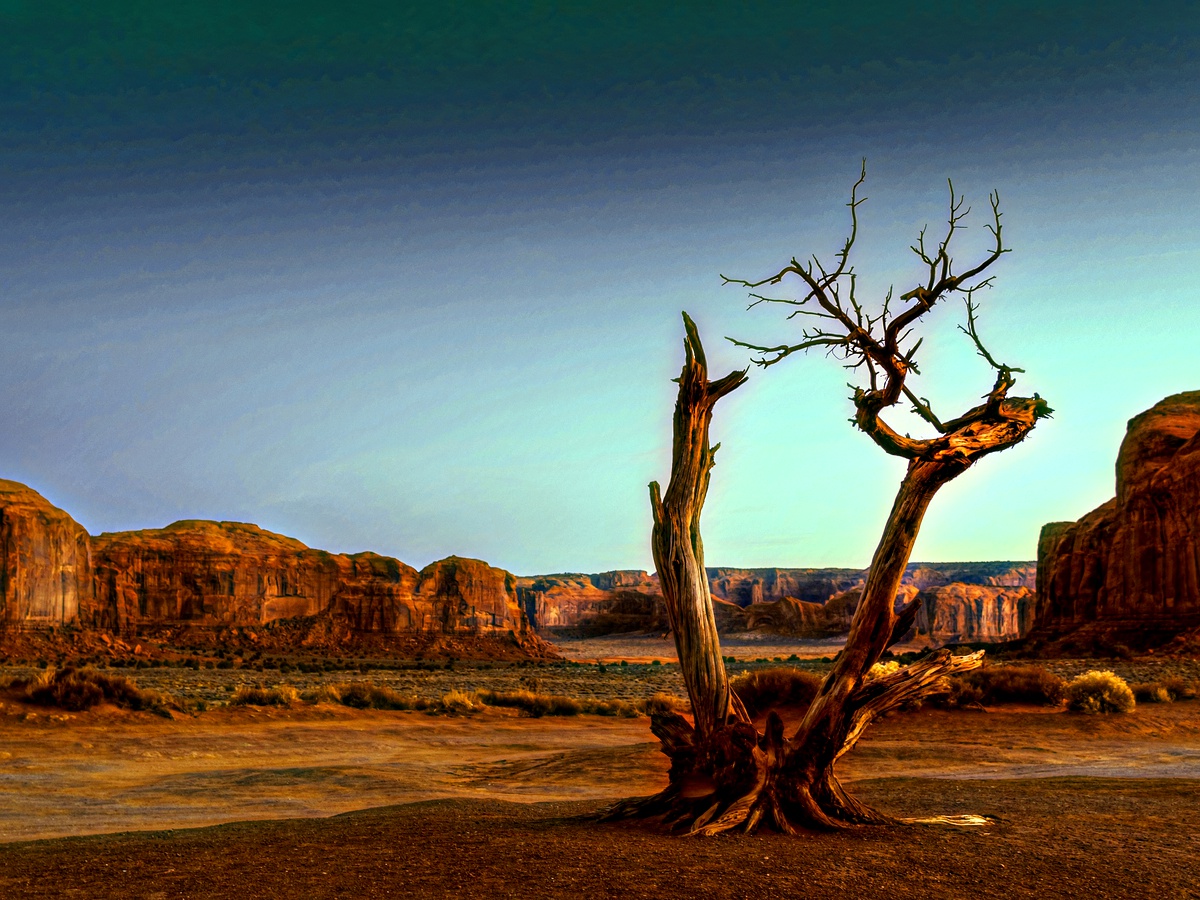} &
    \includegraphics[width=\linewidth]{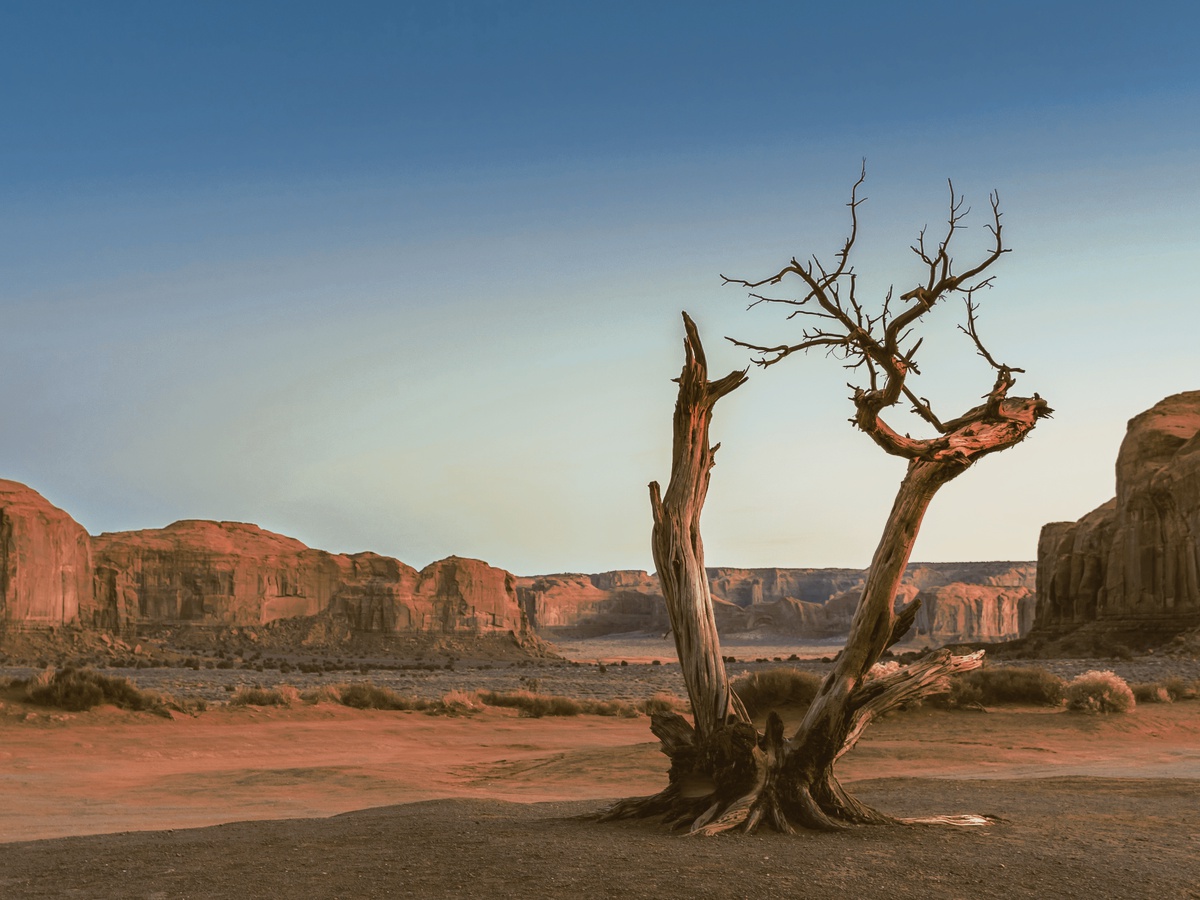} &
    \includegraphics[width=\linewidth]{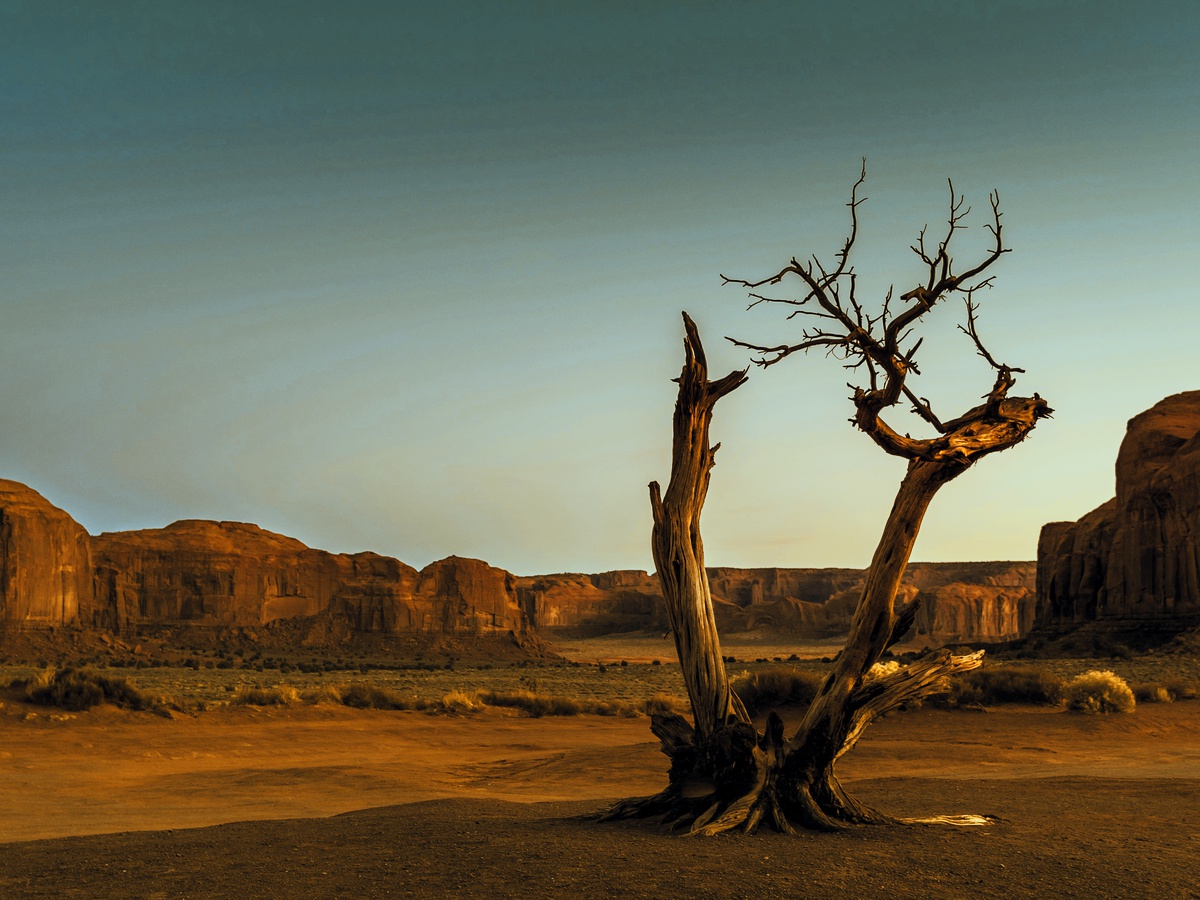} &
    \includegraphics[width=\linewidth]{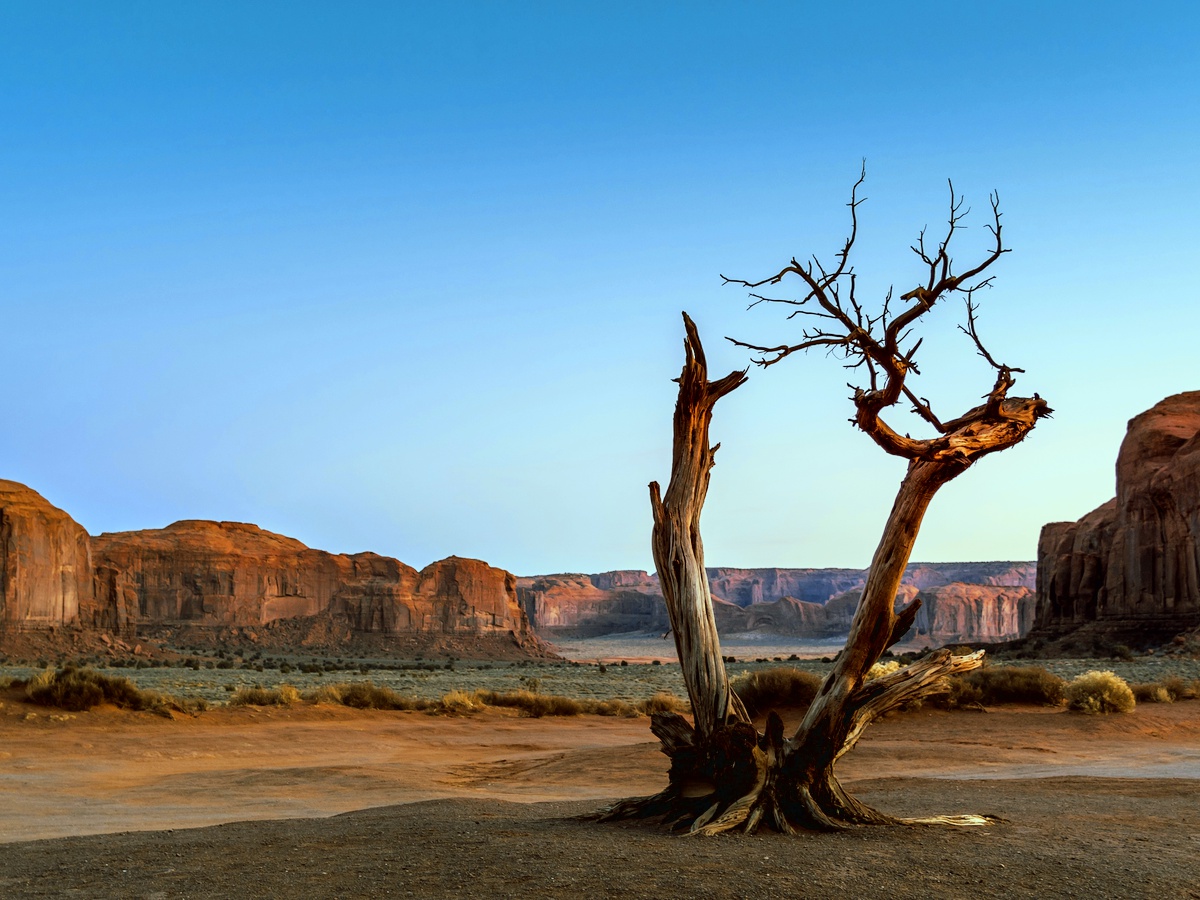}&
    \includegraphics[width=\linewidth]{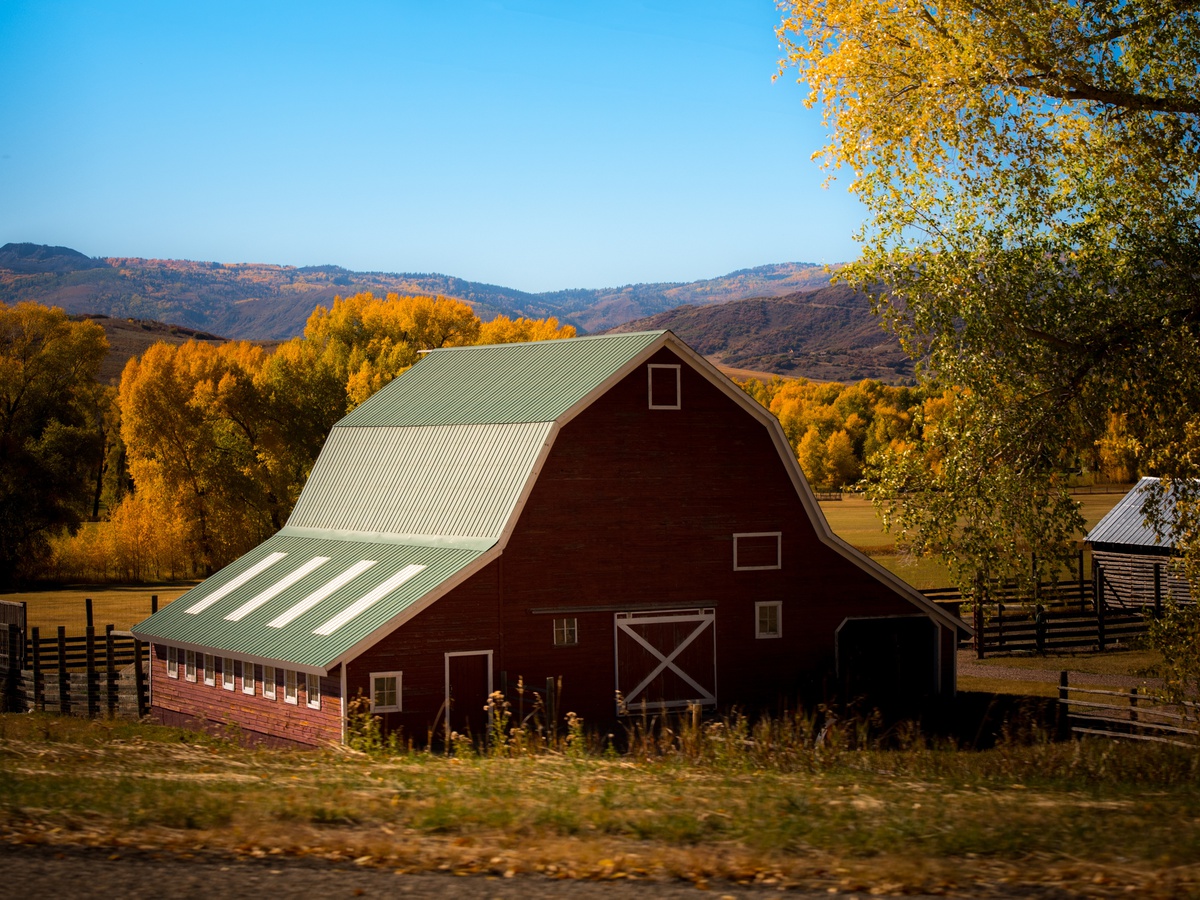}\\ 

    \includegraphics[width=\linewidth]{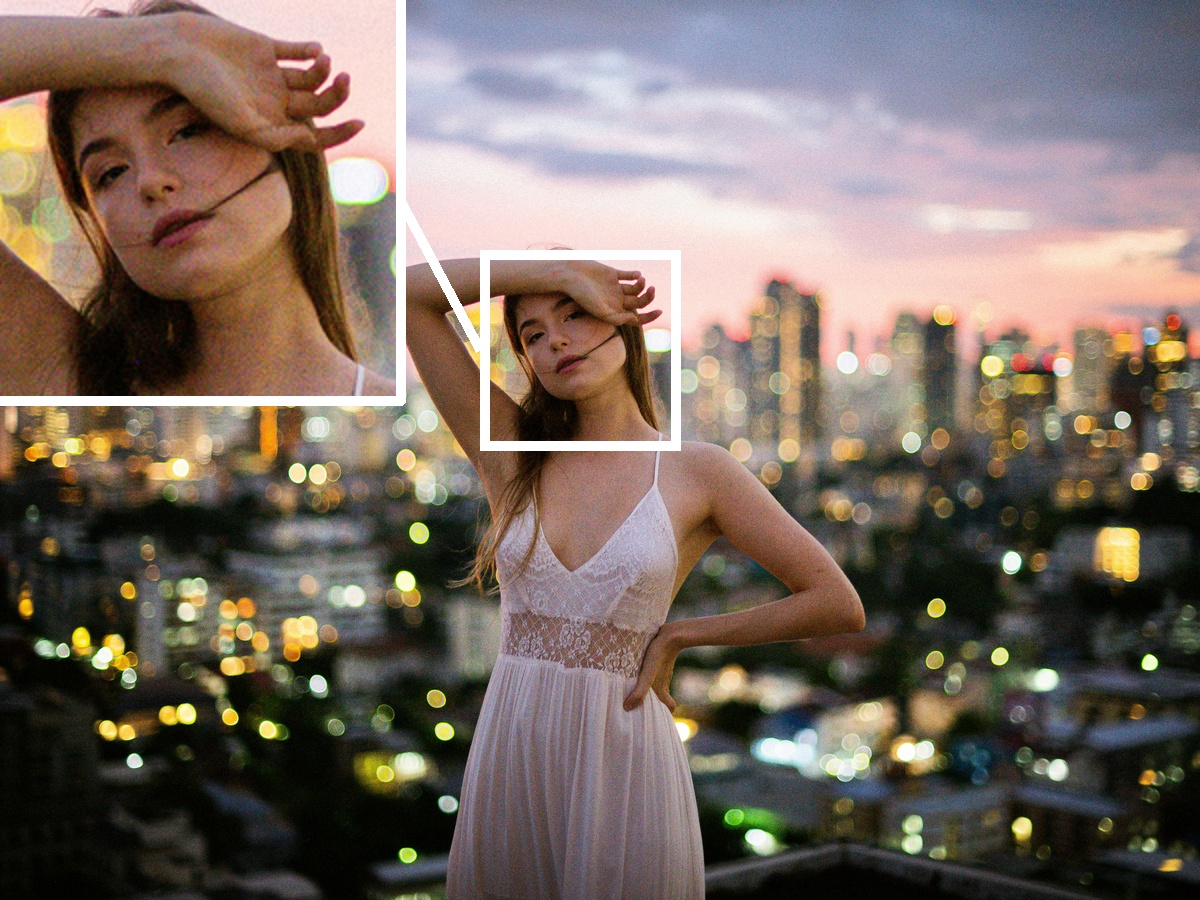} &
    \includegraphics[width=\linewidth]{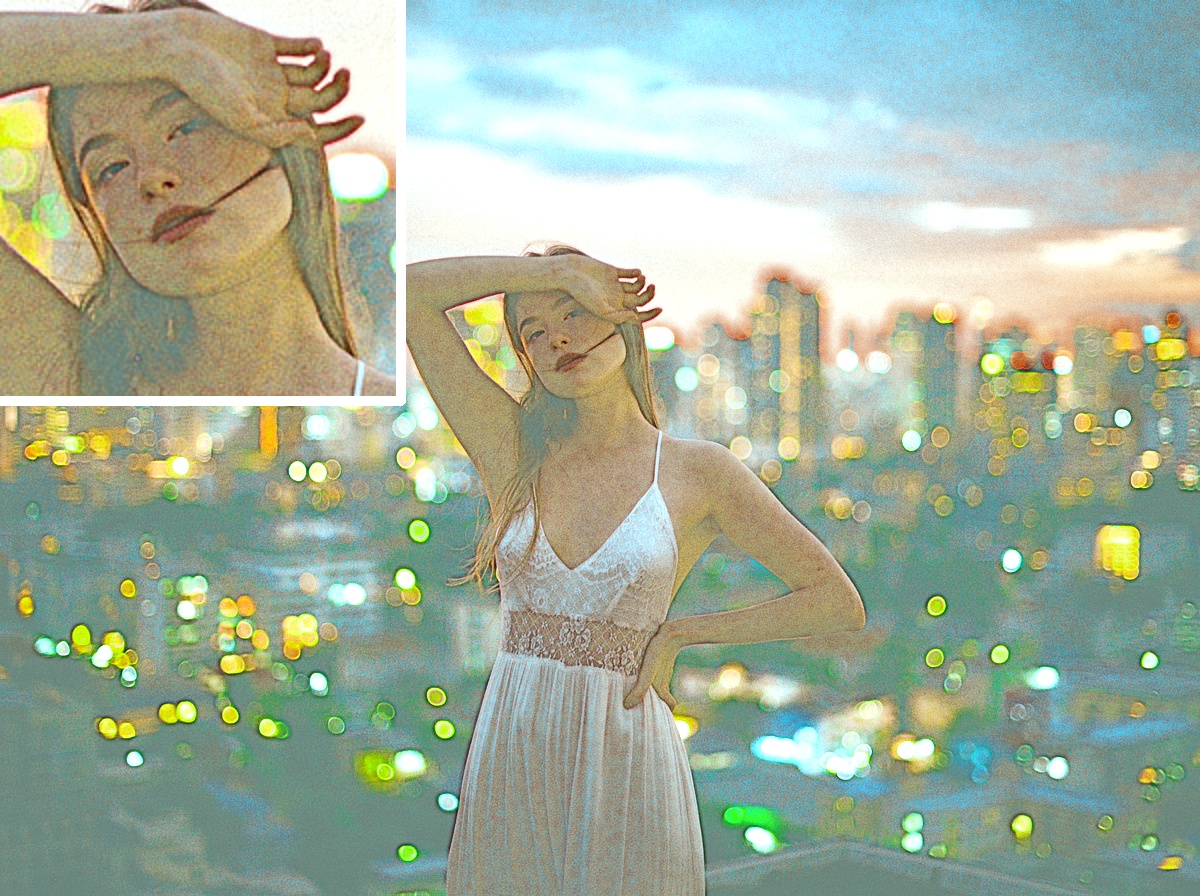} &
    \includegraphics[width=\linewidth]{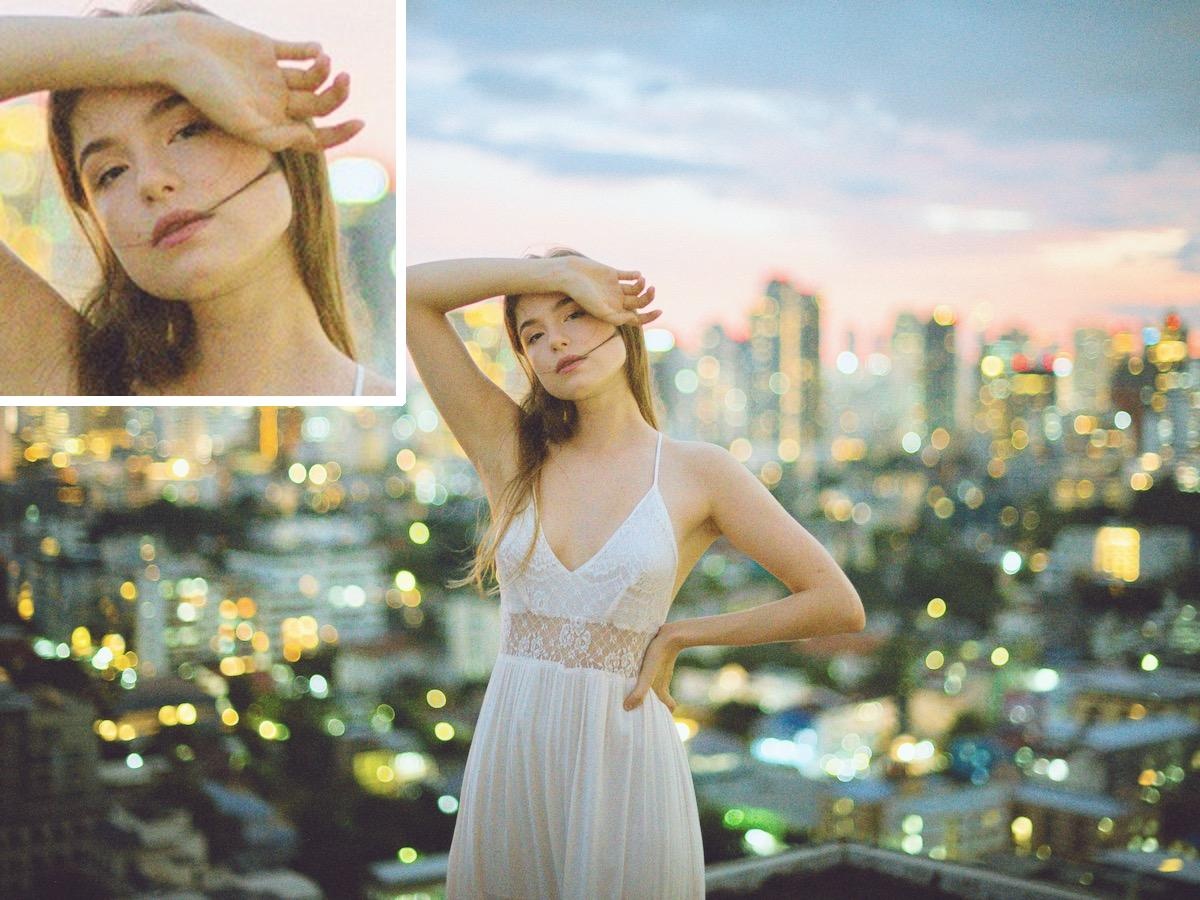} &
    \includegraphics[width=\linewidth]{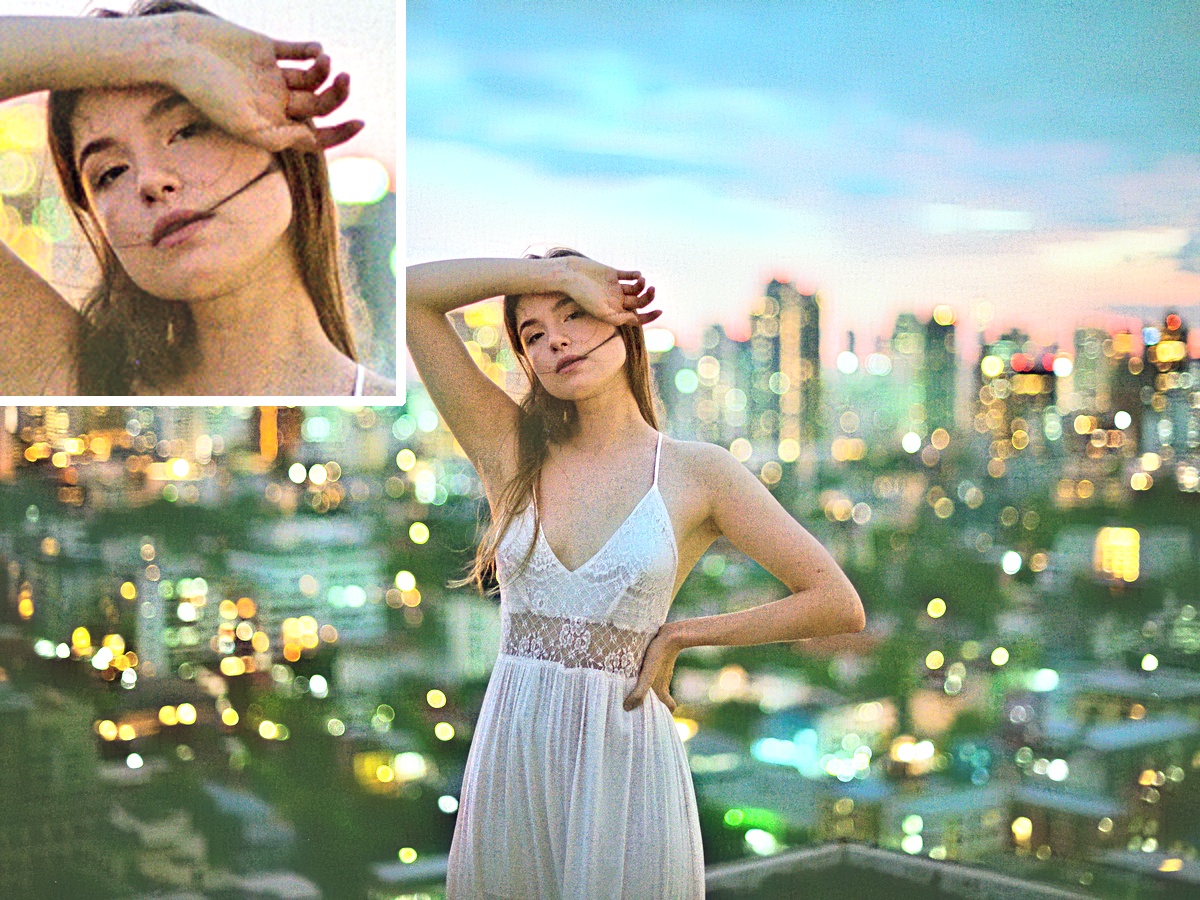} &
    \includegraphics[width=\linewidth]{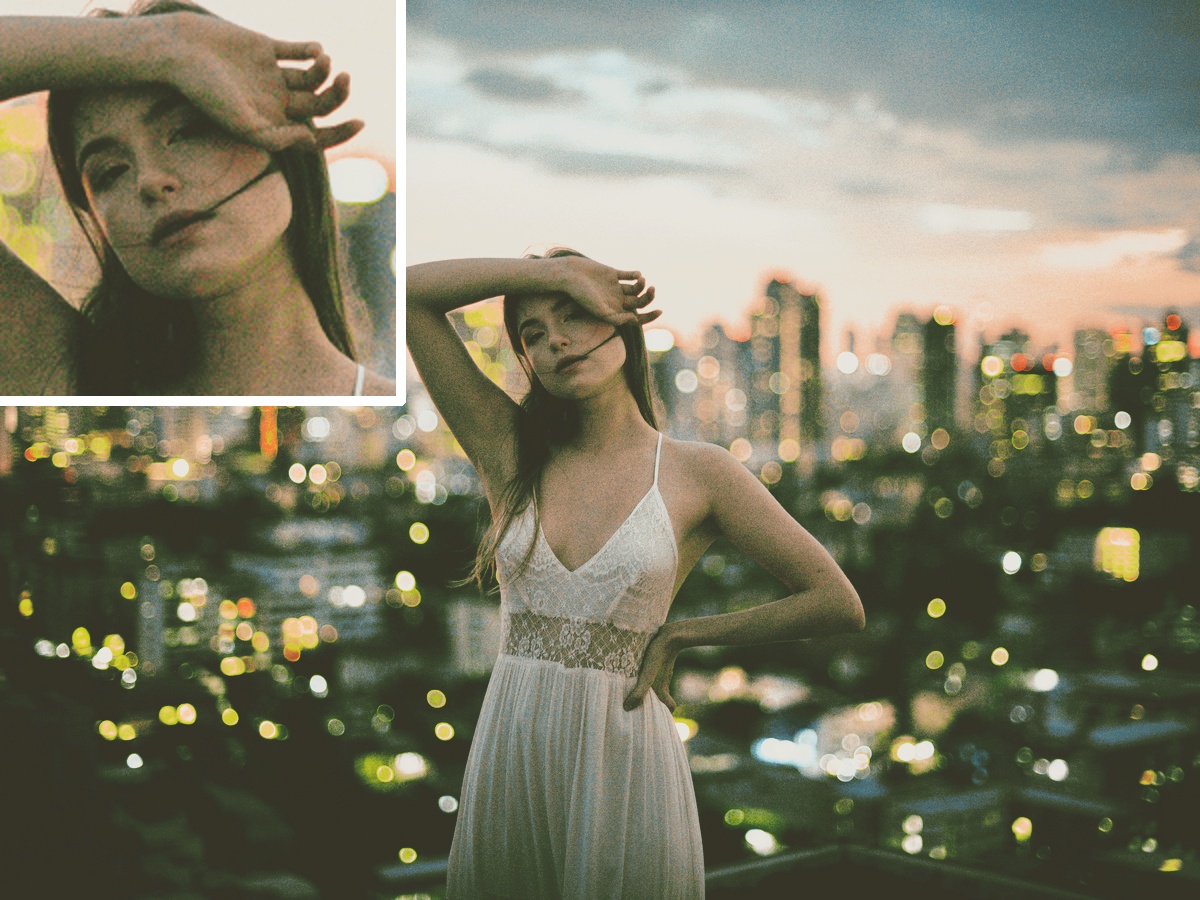} &
    \includegraphics[width=\linewidth]{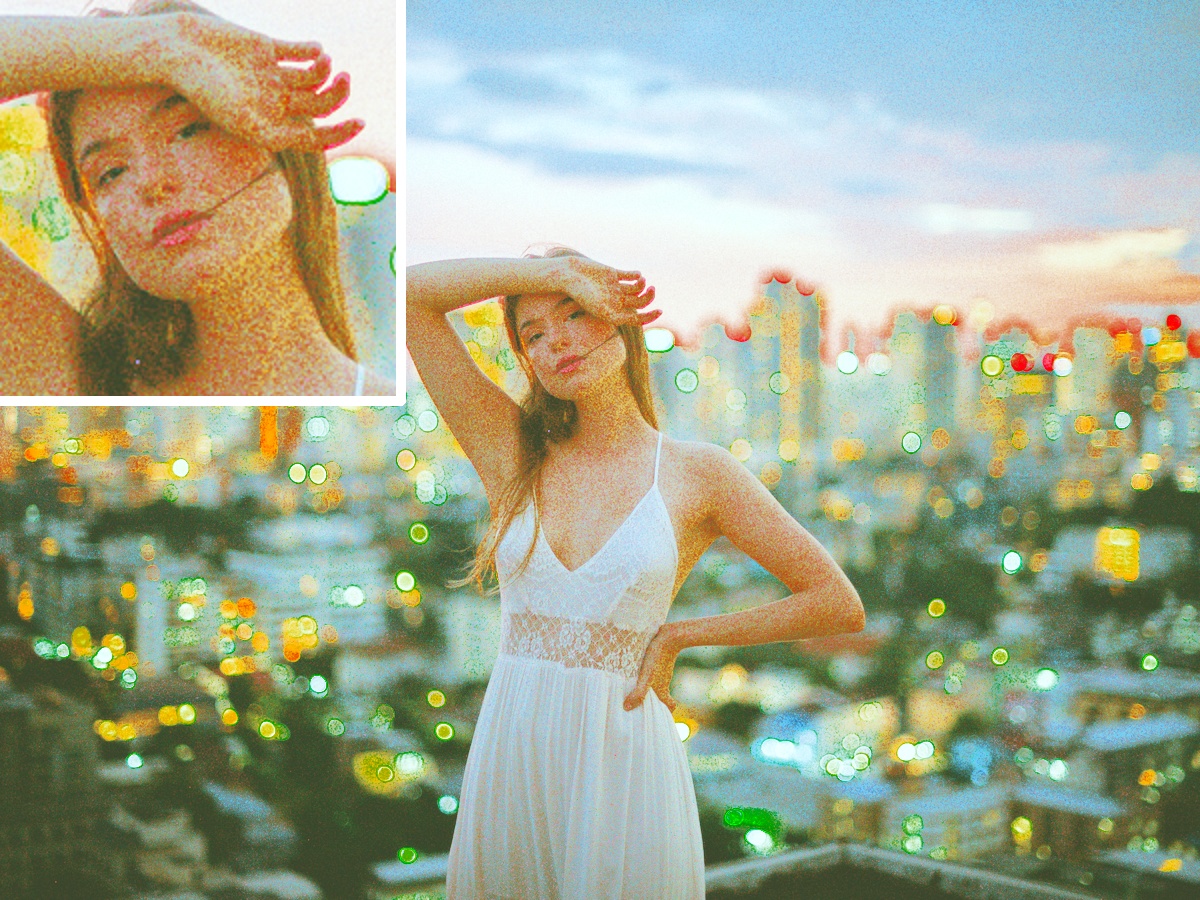} &
    \includegraphics[width=\linewidth]{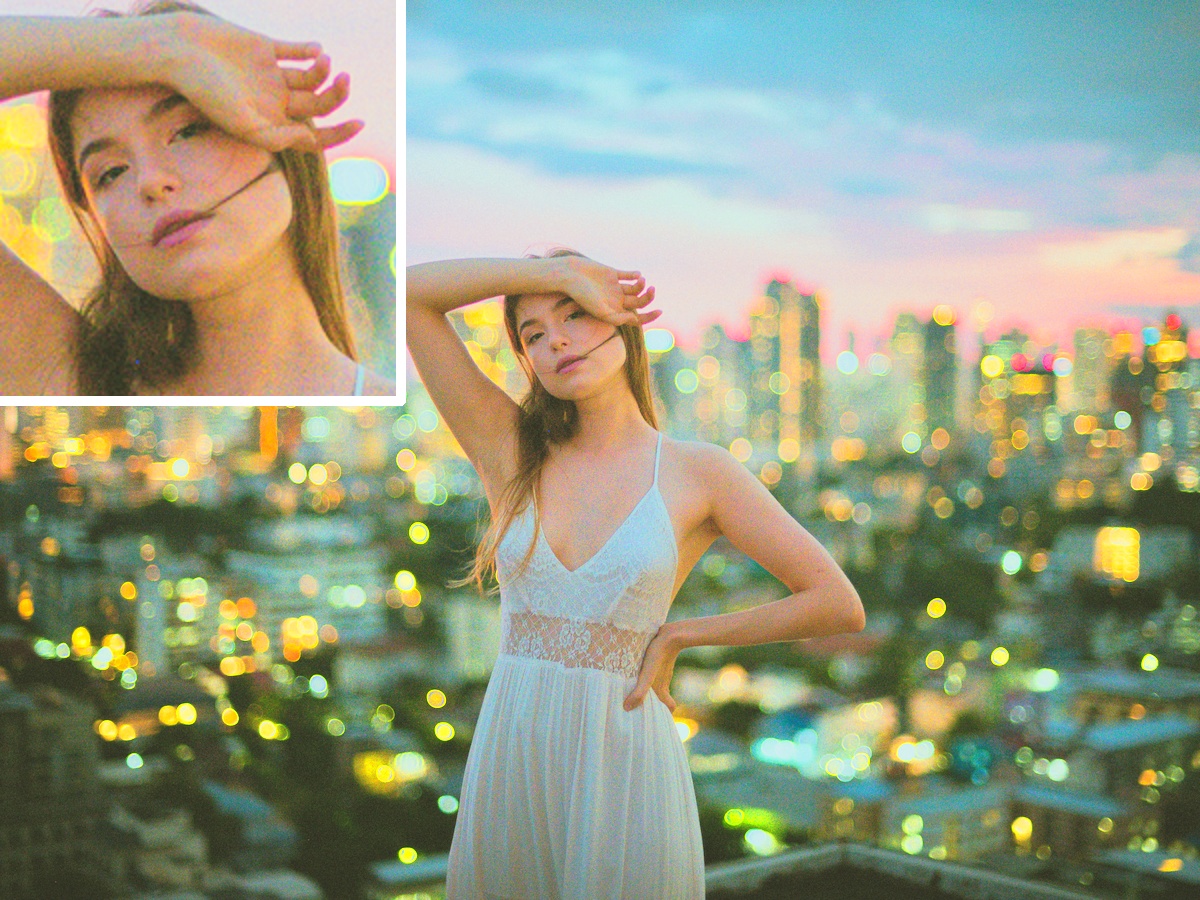}&
    \includegraphics[width=\linewidth]{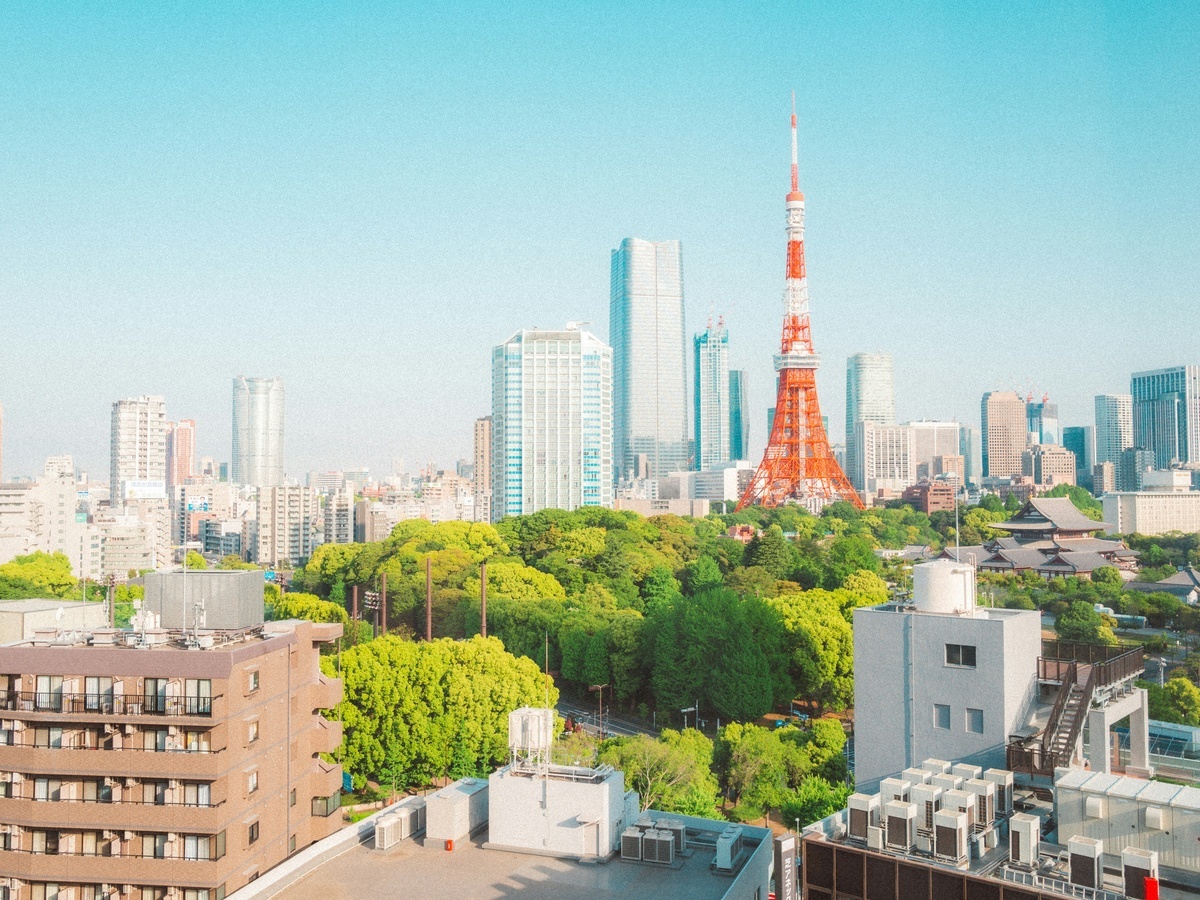}\\
    
    \includegraphics[width=\linewidth]{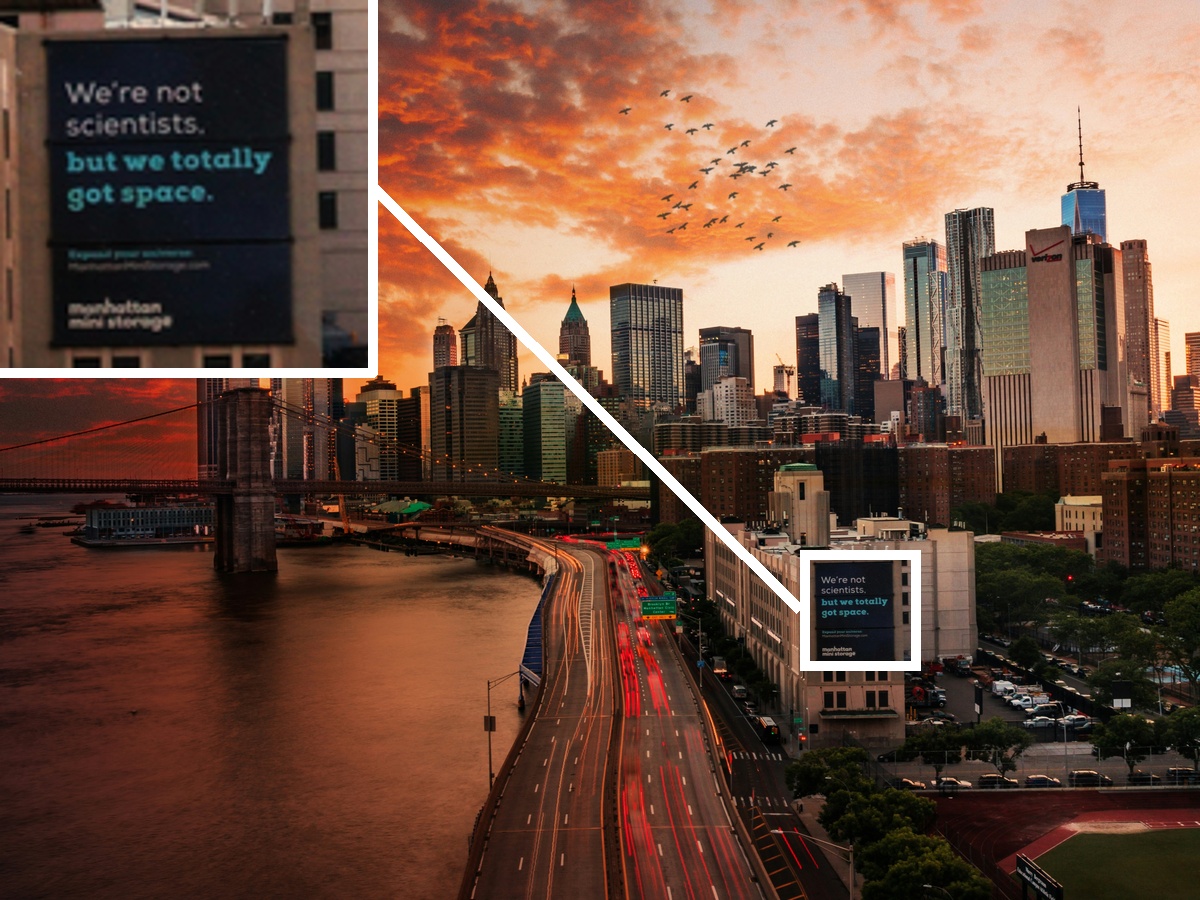} &
    \includegraphics[width=\linewidth]{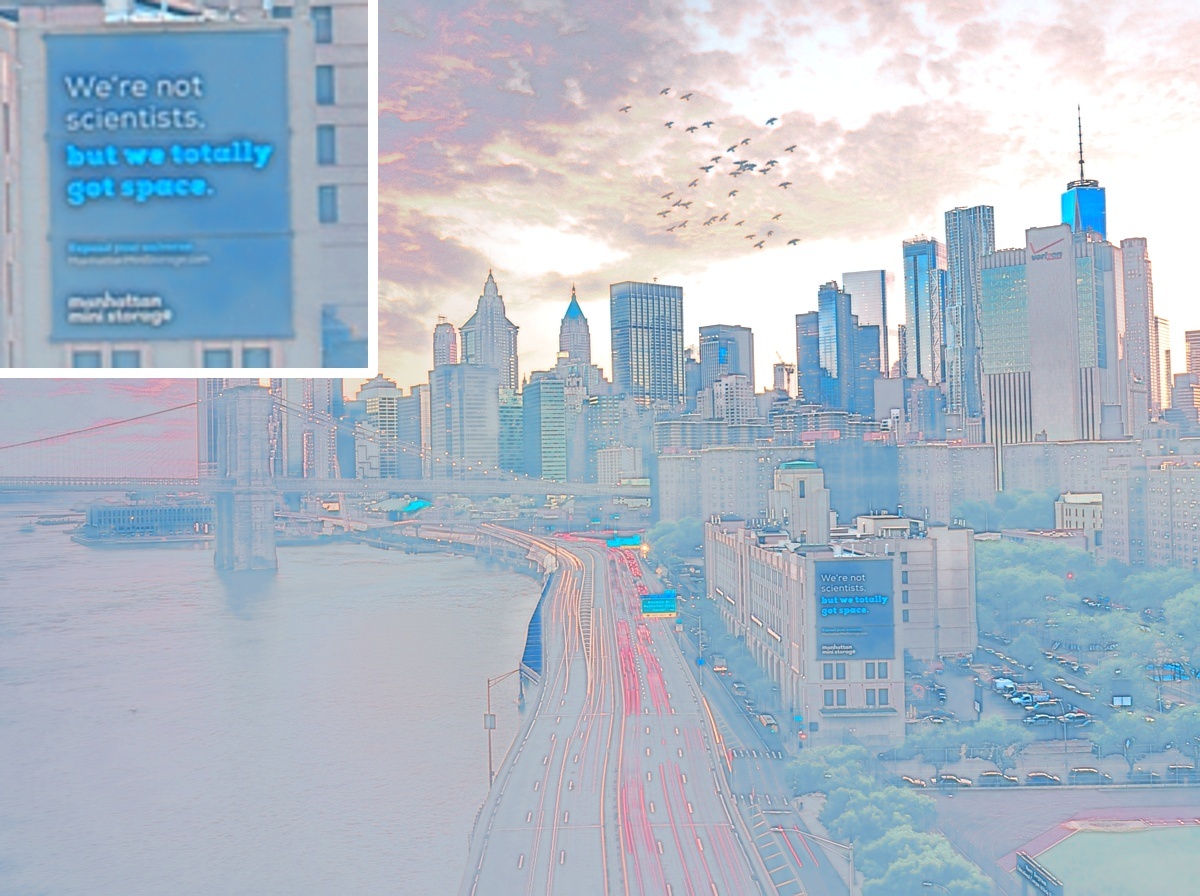} &
    \includegraphics[width=\linewidth]{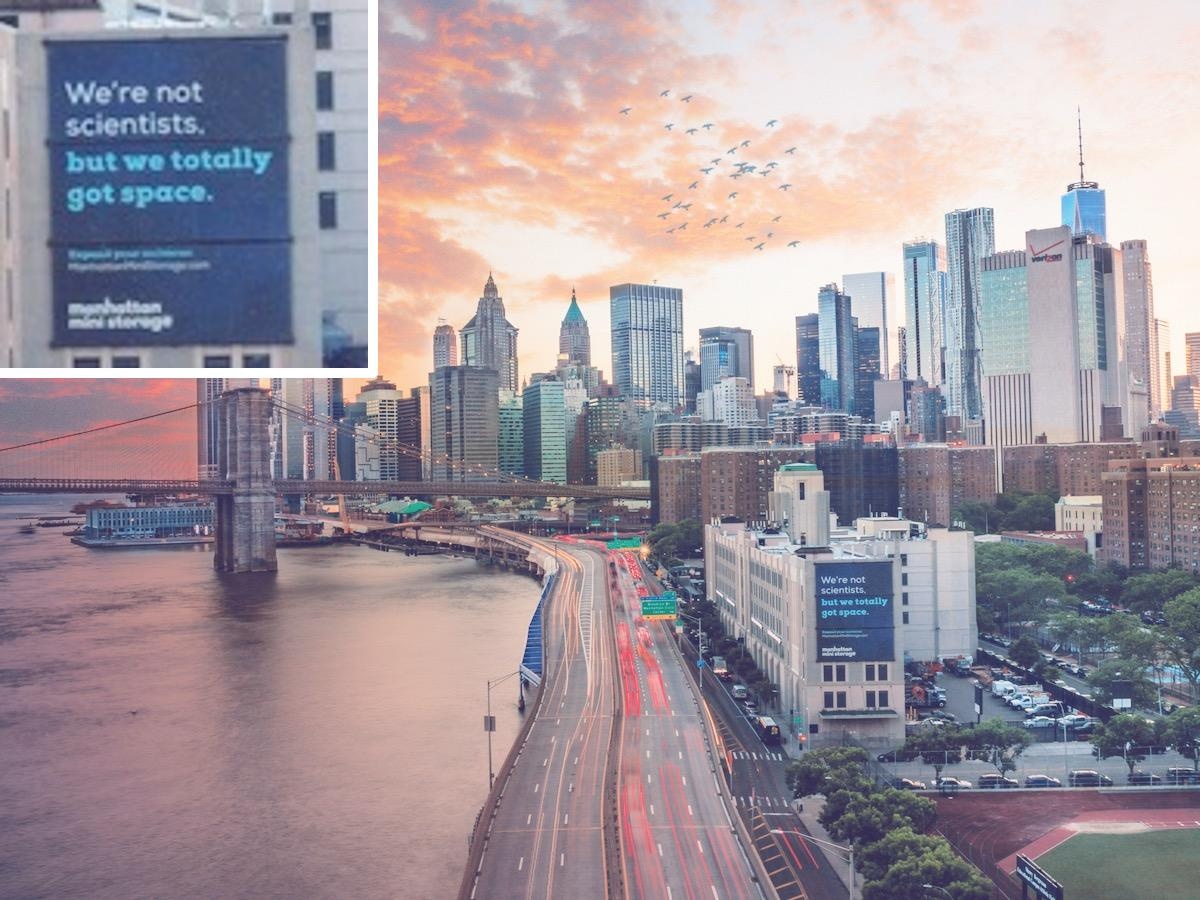} &
    \includegraphics[width=\linewidth]{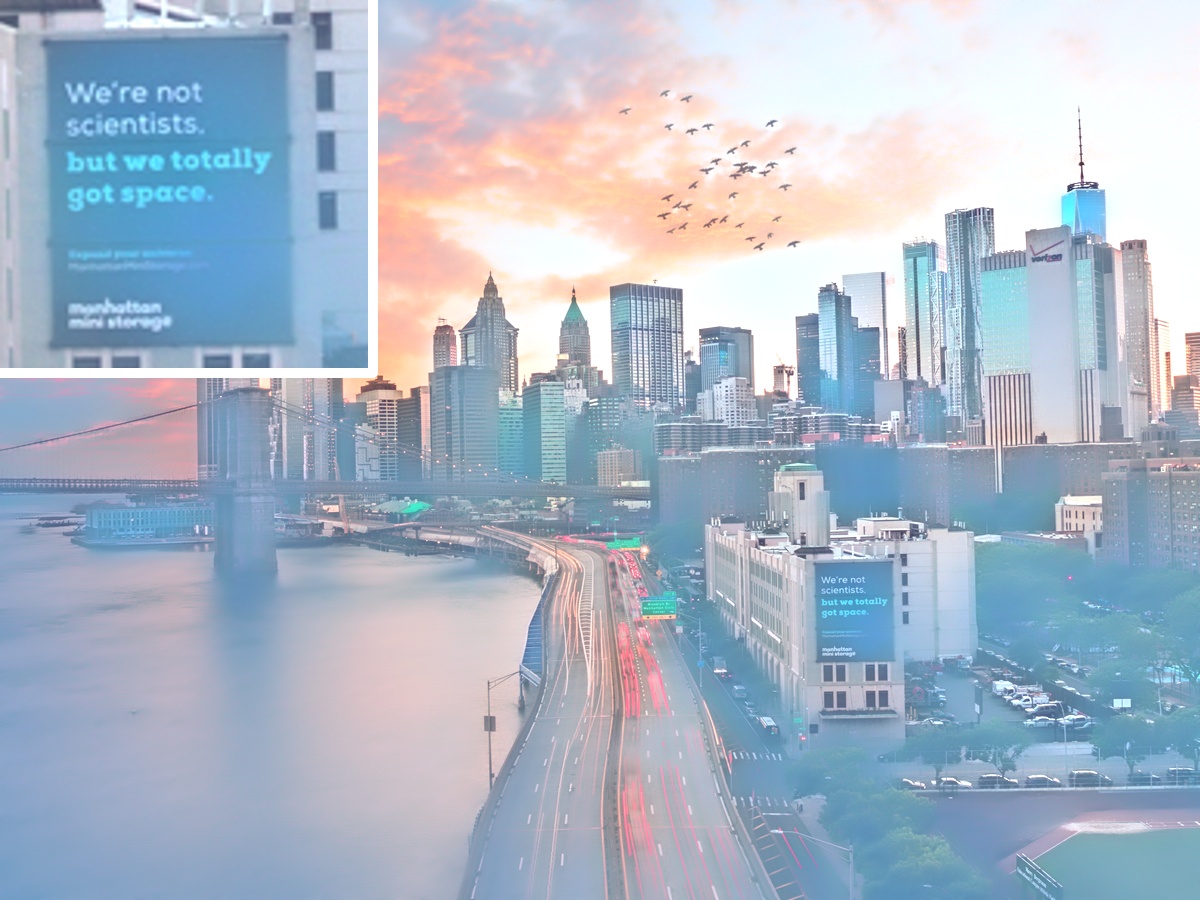} &
    \includegraphics[width=\linewidth]{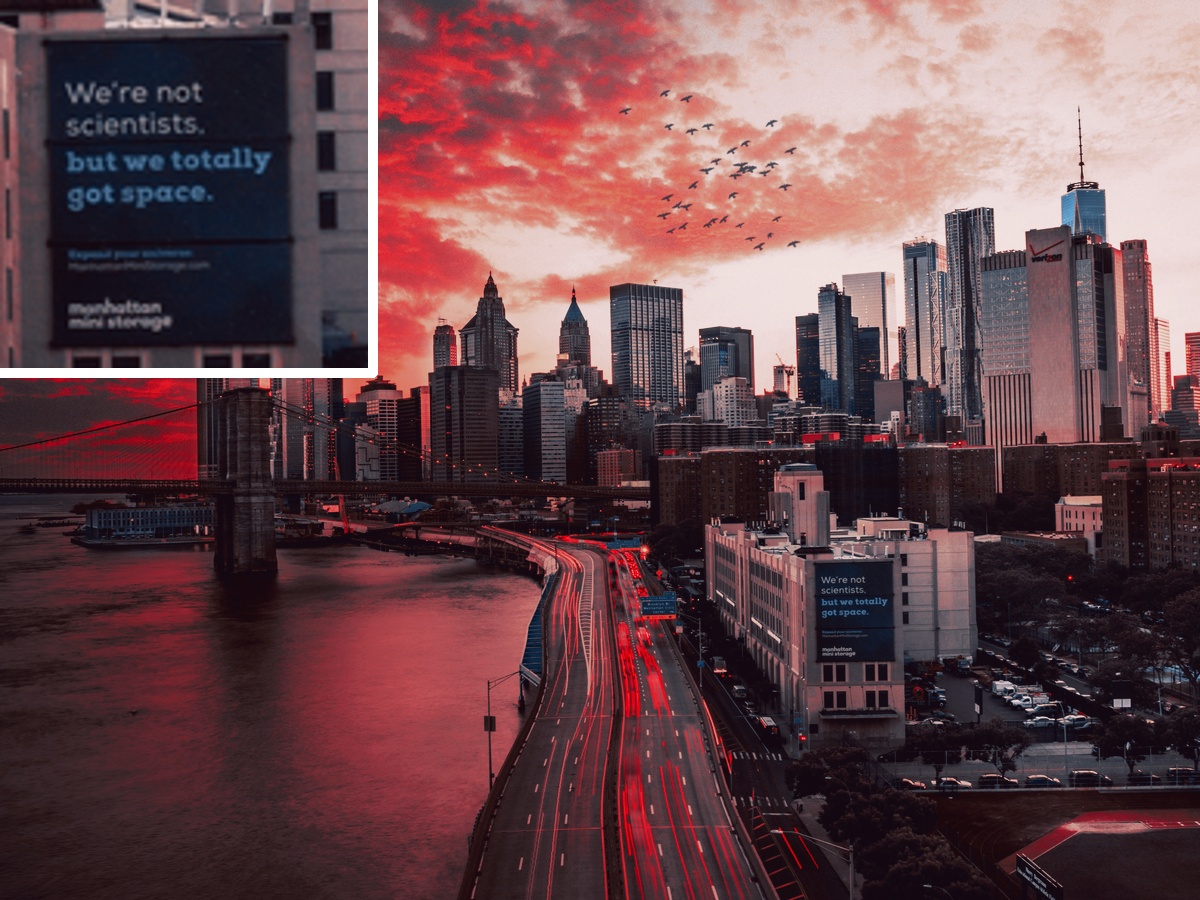} &
    \includegraphics[width=\linewidth]{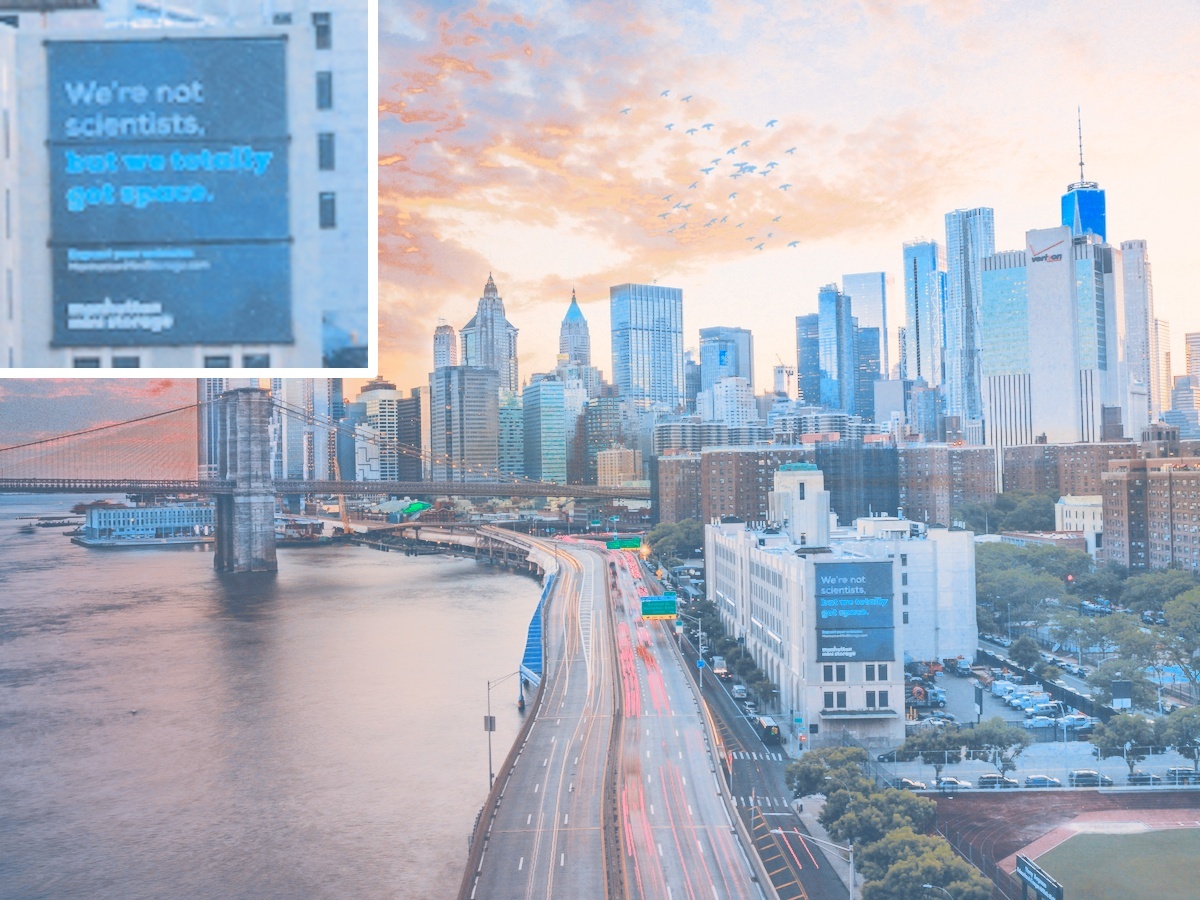} &
    \includegraphics[width=\linewidth]{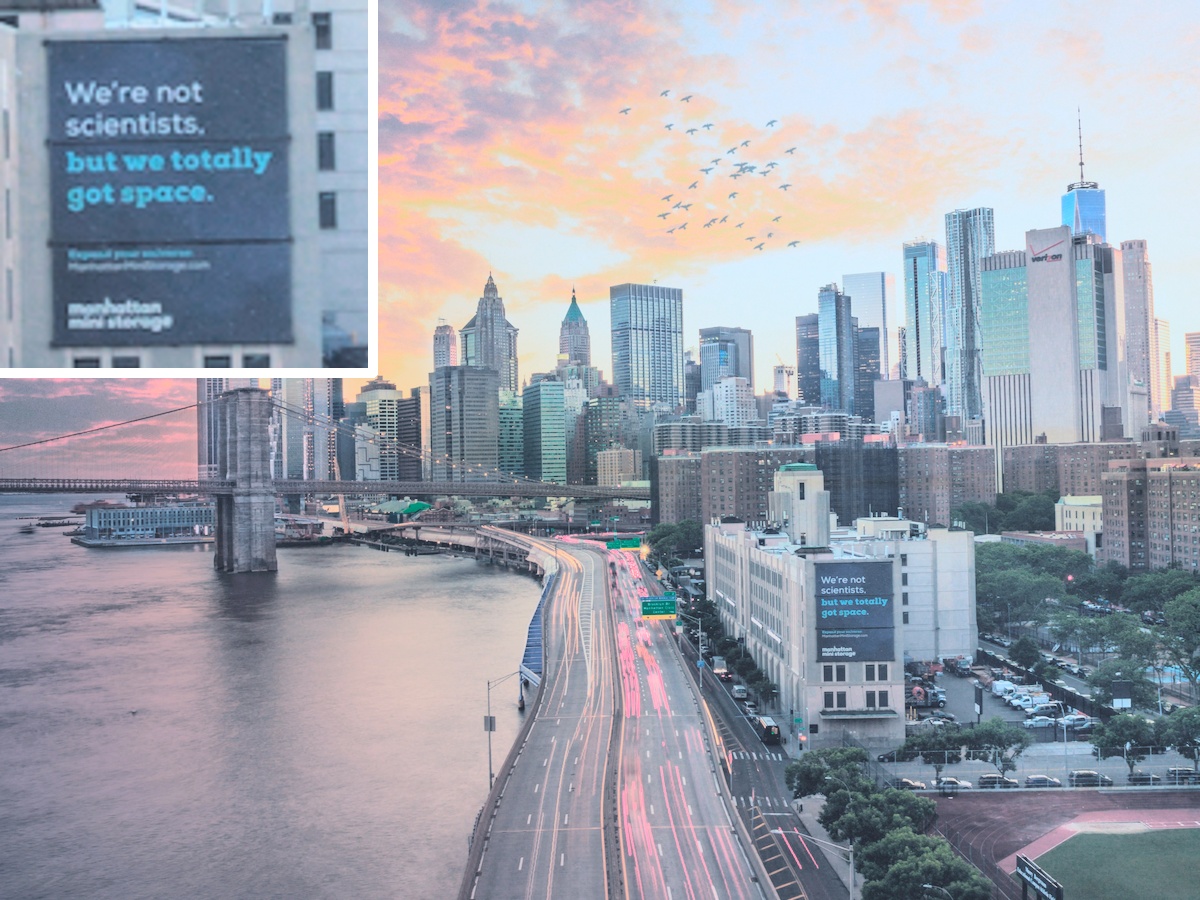}&
    \includegraphics[width=\linewidth]{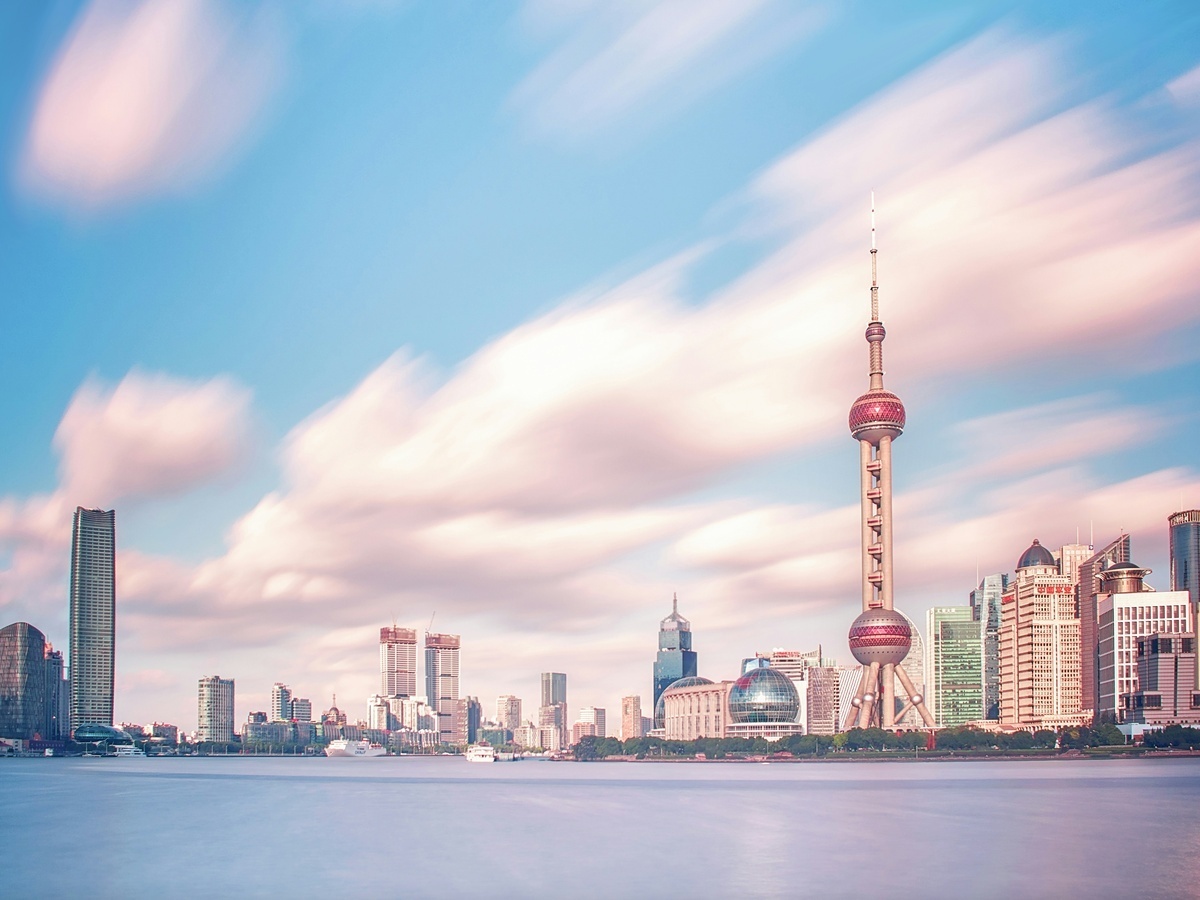}\\

    \fontsize{5pt}{6pt}\selectfont Content & 
    \fontsize{5pt}{6pt}\selectfont WCT$^2$ & 
    \fontsize{5pt}{6pt}\selectfont Neural Preset & 
    \fontsize{5pt}{6pt}\selectfont CAP-VST & 
    \fontsize{5pt}{6pt}\selectfont D-LUT & 
    \fontsize{5pt}{6pt}\selectfont ModFlows & 
    \fontsize{5pt}{6pt}\selectfont ColorFM-L &
    \fontsize{5pt}{6pt}\selectfont Style
    
  \end{tabularx}
  \caption{\textbf{Qualitative Comparison.} Our method exhibits distinct advantages in maintaining global color consistency, ensuring semantic-aware color alignment, and preserving image structure without artifacts.  }
  \label{fig:comparison}
\end{figure*}
\subsection{Comparisons}
We compare both ColorFM-O and ColorFM-L against state-of-the-art methods: WCT$^2$~\cite{yoo2019photorealistic}, PhotoWCT$^2$~\cite{chiu2022photowct2}, Deep Preset~\cite{ho2021deep}, NLUT~\cite{chen2023nlut}, CAP-VST~\cite{wen2023cap}, SA-LUT~\cite{gong2025sa}, D-LUT~\cite{li2025d}, Neural Preset~\cite{ke2023neural}, and ModFlows~\cite{larchenko2025color}. 
As Neural Preset is only available via a mobile application, which prevents large-scale automated evaluation, we restrict its comparison to qualitative visual results.

\begin{table}[t]
\centering

\caption{\textbf{Quantitative Comparison.} 
All images are resized to $512 \times 512$ for fair evaluation. 
Gray backgrounds indicate optimization-based models, and ``Dist. to Ideal'' quantifies the trade-off between style and content metrics. 
\textbf{Bold} and \underline{underlined} indicate the best and second-best results, respectively.
ColorFM-L achieves the best trade-off among all methods.}
\label{tab:quantitative}
\setlength{\tabcolsep}{5pt} 
\begin{tabular}{lccccc}
\toprule
\multirow{2}{*}{Methods} & \multicolumn{3}{c}{Similarity Metrics} &\multirow{2}{*}{\shortstack{Lipschitz\\Constant $\downarrow$}}  & 
\multirow{2}{*}{Time (s) $\downarrow$}\\
\cmidrule(lr){2-4} 
 & Style $\uparrow$ & Content $\uparrow$ & Dist. to Ideal $\downarrow$ &  \\
 \midrule
         \rowcolor{gray!15} NLUT~\cite{chen2023nlut} & 0.731 & 0.650 & 0.441 & 5.15 & 18.937\\
        \rowcolor{gray!15} D-LUT~\cite{li2025d} & 0.513 & 0.724  & 0.560 & 6.66 & 82.845\\
        % CT~\cite{reinhard2002color} & 0.567  & 0.647  & 0.559 & 11.43  \\
        WCT$^2$~\cite{yoo2019photorealistic} & 0.752  & 0.648  & 0.431 &  35.27 & 0.081 \\
        % 0.0804
        PhotoWCT$^2$~\cite{chiu2022photowct2}   & 0.796 & 0.599 & 0.450 & 89.73 & 0.643\\
        Deep Preset~\cite{ho2021deep} & 0.441 & 0.899 & 0.568 & 12.08 & \textbf{0.009} \\
        CAP-VST~\cite{wen2023cap} & 0.844 &  0.627 & 0.404 & 50.30  & 0.039 \\
        SA-LUT~\cite{gong2025sa}& 0.381 & 0.718 & 0.680 & 15.73 & 0.185\\
        ModFlows~\cite{larchenko2025color}  & 0.846 & 0.610 & 0.419 & 5.29 & 0.149\\
        
        \midrule
        \rowcolor{gray!15} ColorFM-O & 0.811 &  0.720 & \underline{0.338} & \underline{2.82} & 19.314\\
        ColorFM-L & 0.825  &  0.732  & \textbf{0.320} & \textbf{2.67} & \underline{0.016}\\
        \bottomrule
\end{tabular}
\end{table}

\noindent\textbf{Quantitative Metrics.}
Following the evaluation protocols established in previous works~\cite{yoo2019photorealistic, ke2023neural, li2025d, larchenko2025color}, we evaluate performance along two primary dimensions: similarity and regularity. 
For similarity, we employ Style Similarity (using the pre-trained discriminator from Neural Preset~\cite{ke2023neural, li2025d}, with range $[0,1]$) and Content Similarity (computed as SSIM~\cite{wang2004image} on LDC-extracted edge maps~\cite{soria2022ldc}). To quantify the balance, we report the Distance to Ideal~\cite{yoo2019photorealistic, ke2023neural, li2025d, larchenko2025color}, defined as the Euclidean distance to the optimal point $(1, 1)$, where a shorter distance indicates a better trade-off between content preservation and style alignment. 
For regularity, we report the Lipschitz Constant as used in ModFlows~\cite{larchenko2025color}, where lower values correspond to fewer artifacts and less color banding. Detailed metric formulations are provided in Supplementary Appendix C.

\noindent\textbf{Quantitative Results.}
\cref{tab:quantitative} presents a comprehensive evaluation of ColorFM-O and ColorFM-L against prior methods. Optimization-based models are marked with gray backgrounds. In terms of similarity metrics, ColorFM-L and ColorFM-O attain the top two results in the ``Distance to Ideal'' metric, as shown in \cref{fig:left_image}, indicating the optimal balance between content preservation and style alignment. While ModFlows and CAP-VST exhibit slightly higher style similarity, they suffer from structural degradation. Conversely, Deep Preset and D-LUT  maintain high content fidelity but fail to effectively transfer complex color styles. Regarding regularity, ColorFM-L achieves the lowest Lipschitz constant among all methods. This result indicates that our model learns a smooth and stable color mapping, effectively suppressing visual artifacts and color banding prevalent in competing approaches. In terms of efficiency, ColorFM-L achieves real-time inference (0.016 s) and maintains exceptional speed, processing 4K resolution images  (3840 $\times$ 2160) in 0.043 s on average. A comprehensive runtime analysis across different resolutions is provided in the Supp.~Tab.~1.

Notably, ColorFM-L outperforms ColorFM-O. This phenomenon can be attributed to two primary factors. On the one hand, ColorFM-O employs a fixed optimization setting (\eg, fixed iteration steps) across all test pairs. This may lead to underfitting or overfitting for samples with varying complexity~\cite{ulyanov2018deep}. On the other hand, ColorFM-L leverages neural network inductive biases~\cite{rahaman2019spectral} and large-scale training to capture global priors. This enables it to smooth out the irregularities of instance-specific optimization, resulting in more robust transfer.

\noindent\textbf{Qualitative Results.}
\cref{fig:comparison} provides visual comparisons of ColorFM-L with competing methods across diverse content-style pairs. 
As shown, WCT$^2$ and CAP-VST often introduce spatial distortions or visual artifacts, while D-LUT and Neural Preset tend to produce under-stylized results with insufficient color transfer. Meanwhile, ModFlows tends toward global color averaging; by neglecting semantic correspondence, it leads to visible color bleeding and artifacts. 
In contrast, ColorFM-L not only performs effective semantically aligned color transfer but also preserves the original image structure. For instance, in Rows 1 and 2, our method accurately distinguishes and aligns the colors of the sky and ground. In Row 3, ColorFM-L successfully transfers the style without degrading the facial details (see the zoomed-in patches). Furthermore, in Row 4, our method retains fine-grained textures, ensuring that the text on the building remains clearly visible.  For a detailed analysis of  ColorFM-O and ColorFM-L, please refer to 
\cref{compare}.

\subsection{Ablation Studies}

\begin{figure*}[t]
  \centering
    \begin{minipage}[b]{0.38\textwidth}
    \centering
    \includegraphics[width=\linewidth, keepaspectratio]{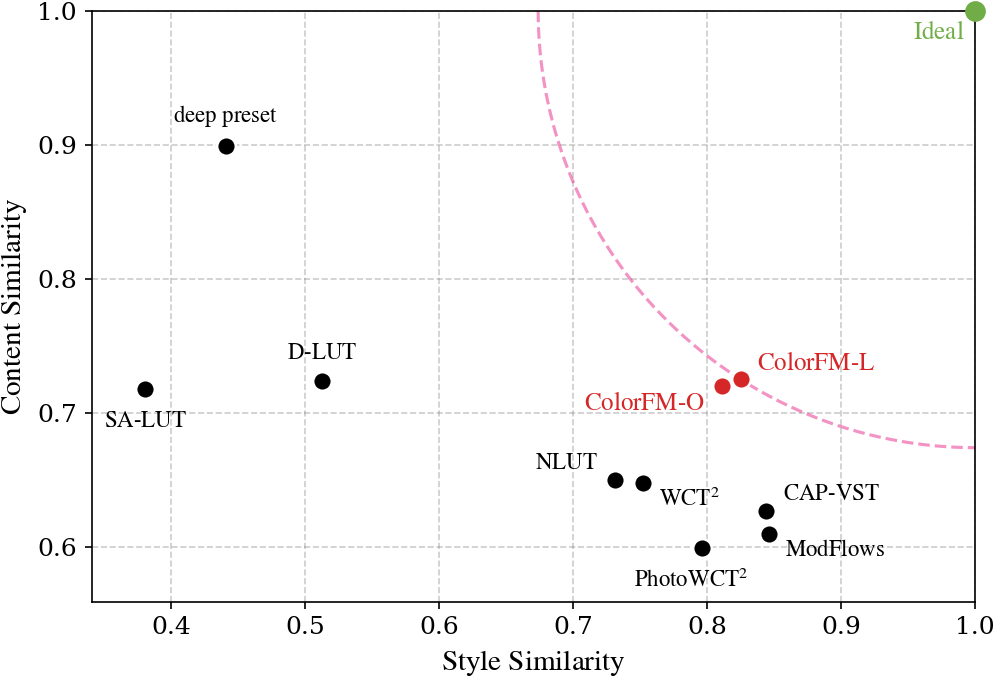}
    \caption{\textbf{Visual Results of Distance to Ideal.} ColorFM achieves the optimal trade-off between content preservation and style alignment.}
    \label{fig:left_image}
  \end{minipage}
    \hfill 
  \begin{minipage}[b]{0.60\textwidth}
    \centering
    \setlength{\tabcolsep}{0.5pt} 
    \renewcommand{\arraystretch}{0.5} 
    \begin{tabularx}{\linewidth}{Y Y Y Y} 
      \includegraphics[width=\linewidth]{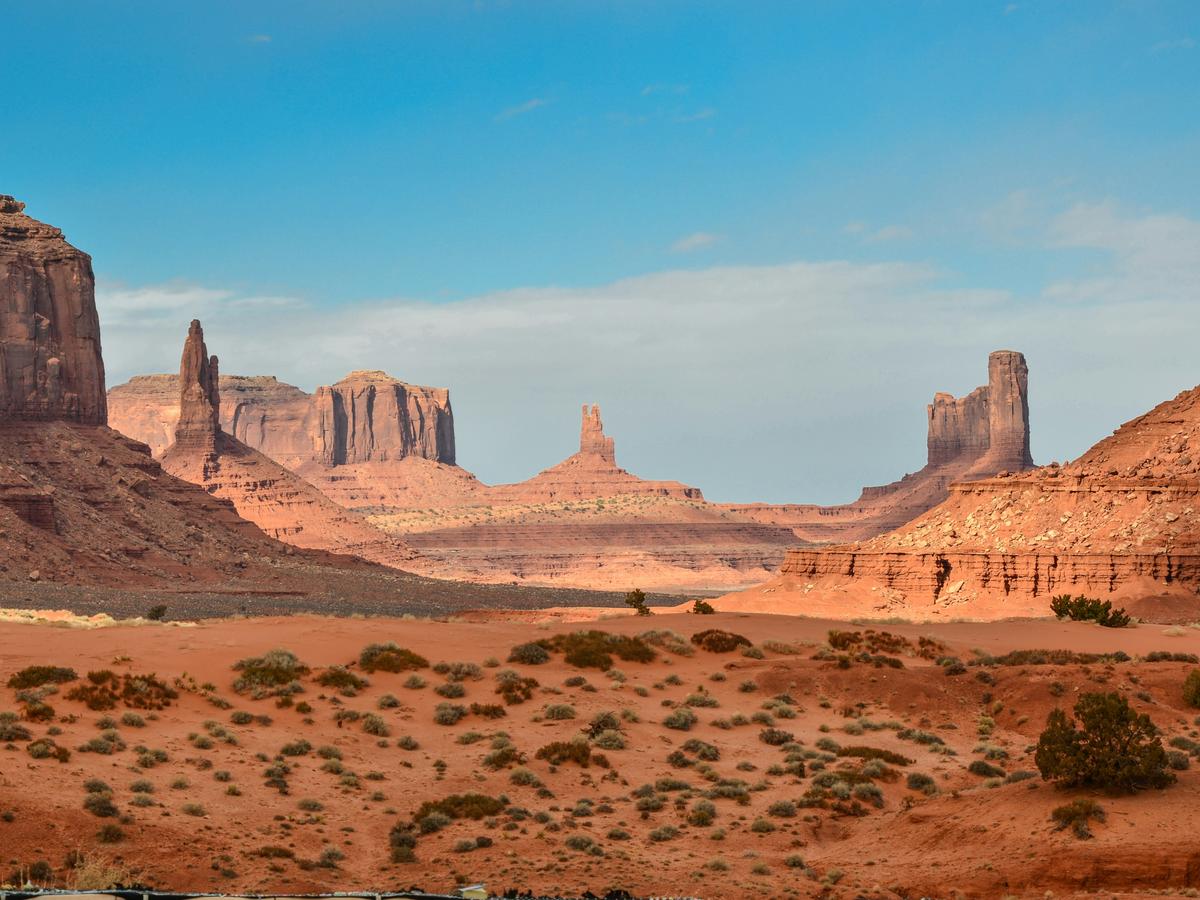} &
      \includegraphics[width=\linewidth]{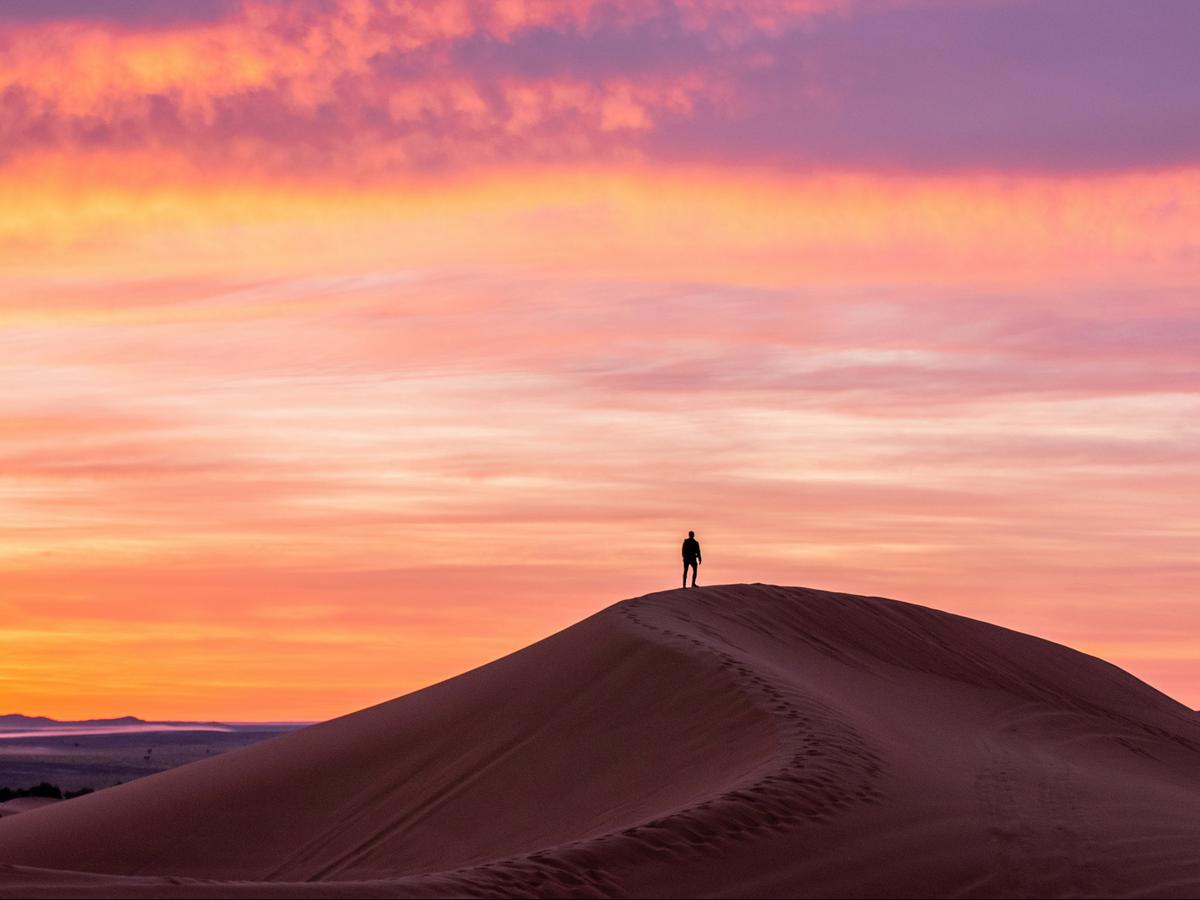} &
      \includegraphics[width=\linewidth]{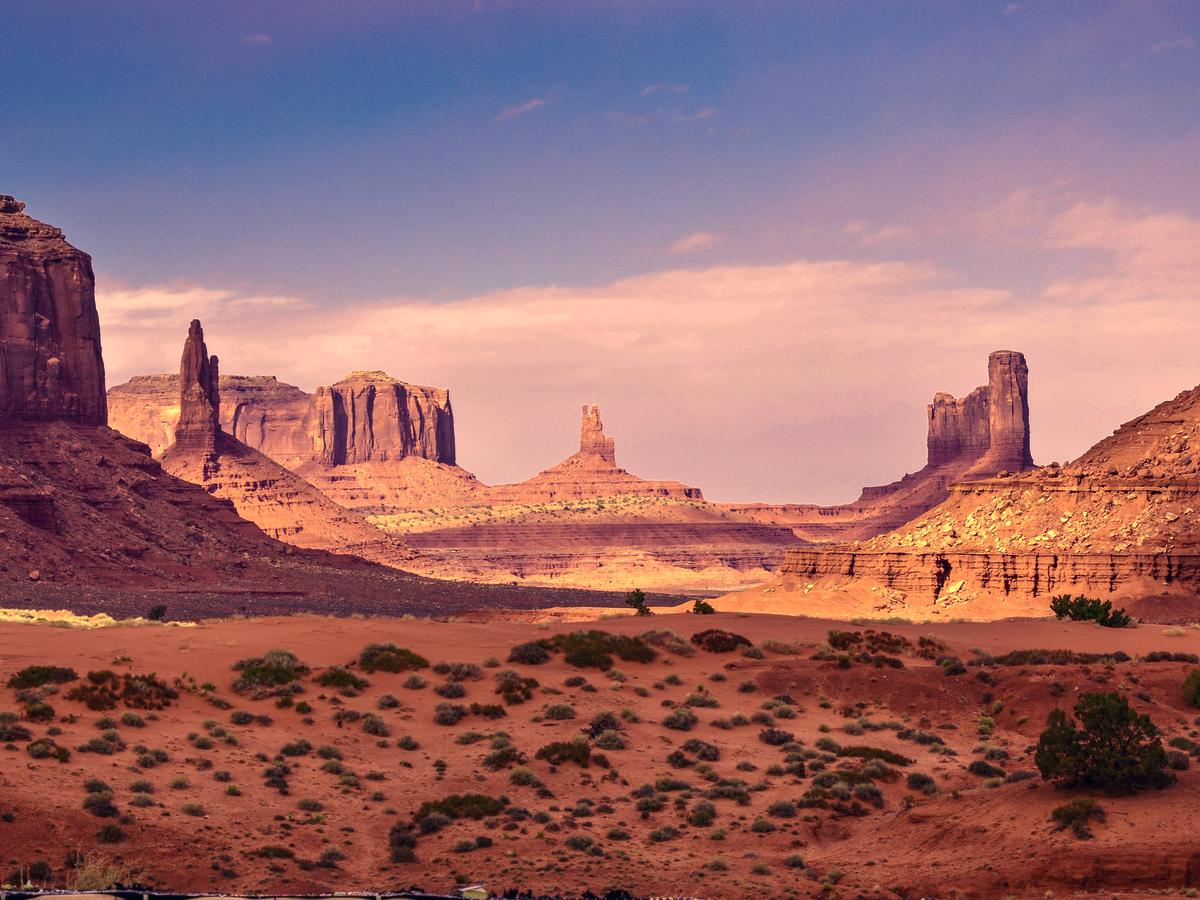} &
      \includegraphics[width=\linewidth]{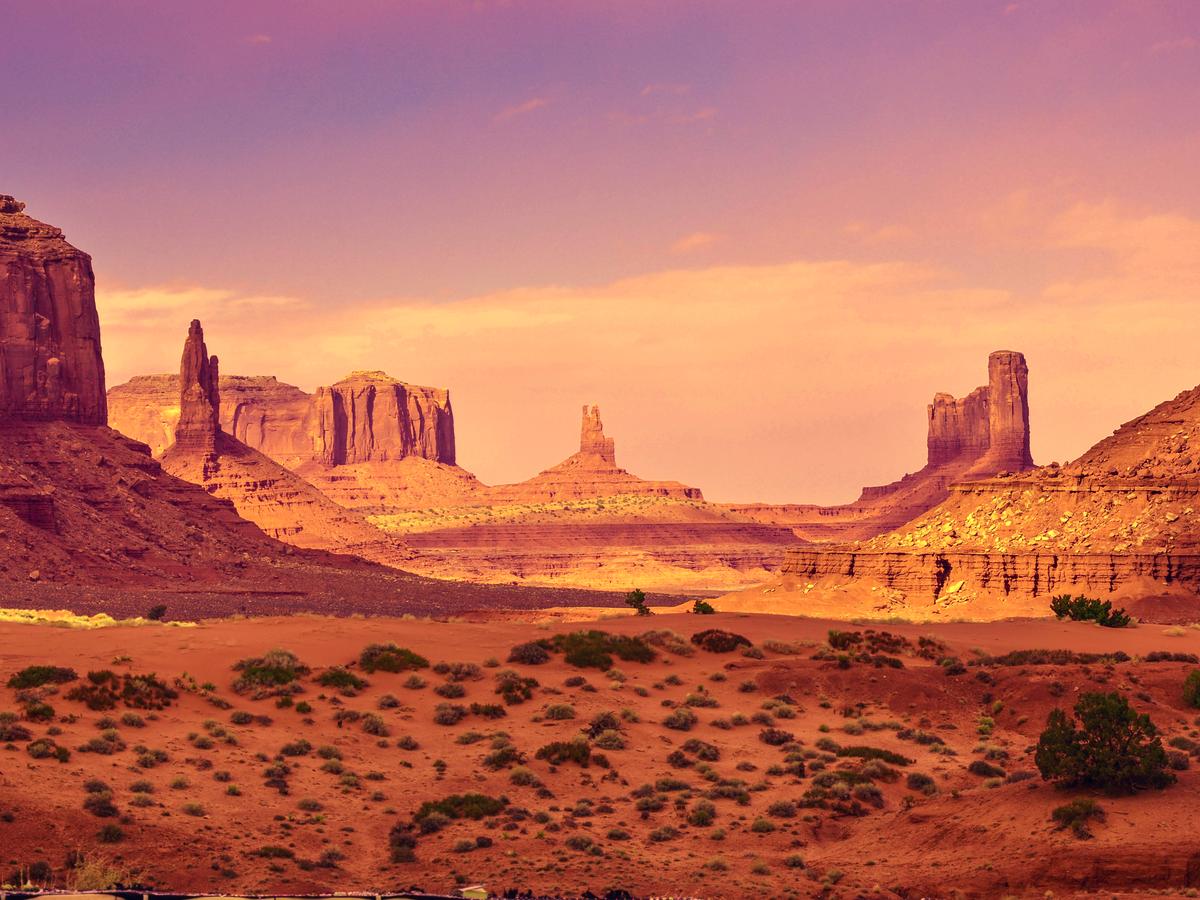}\\
      
      \includegraphics[width=\linewidth]{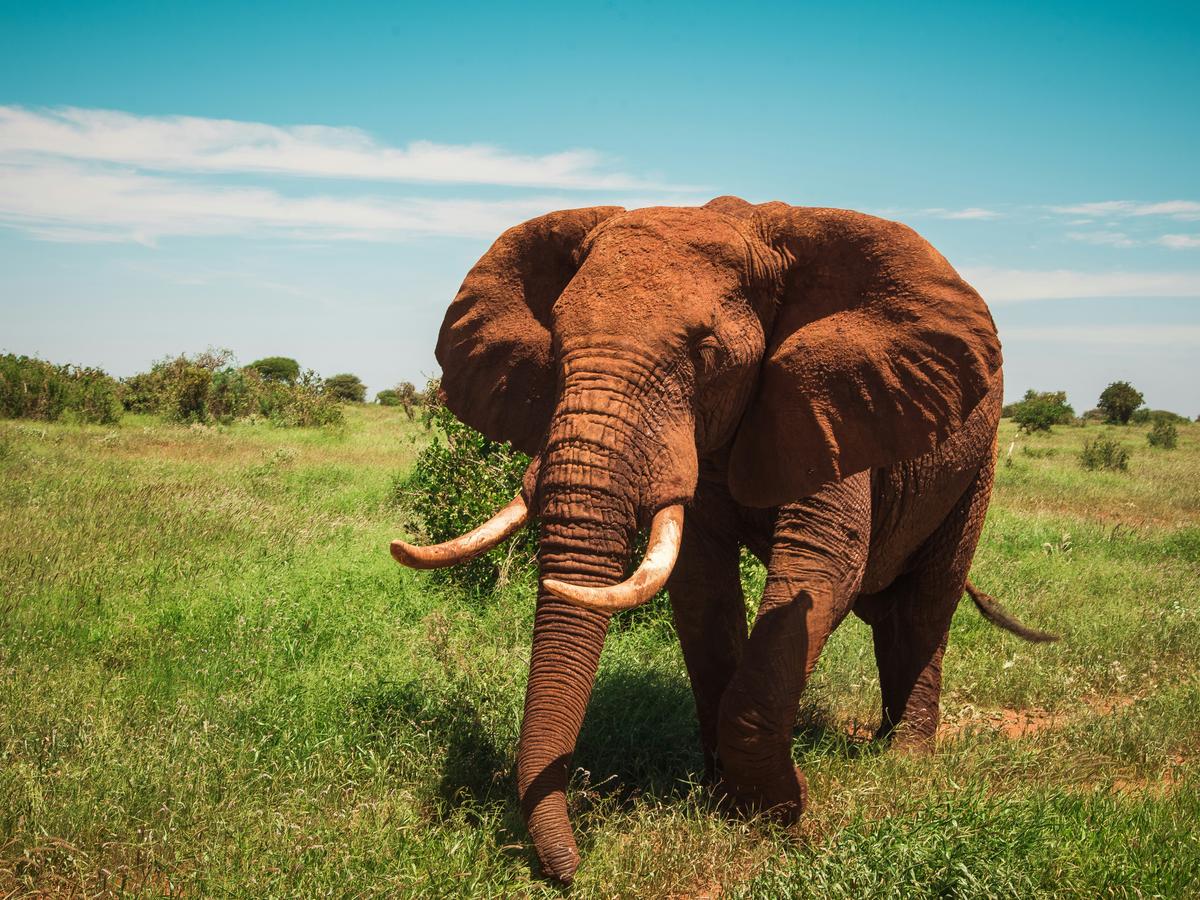} &
      \includegraphics[width=\linewidth]{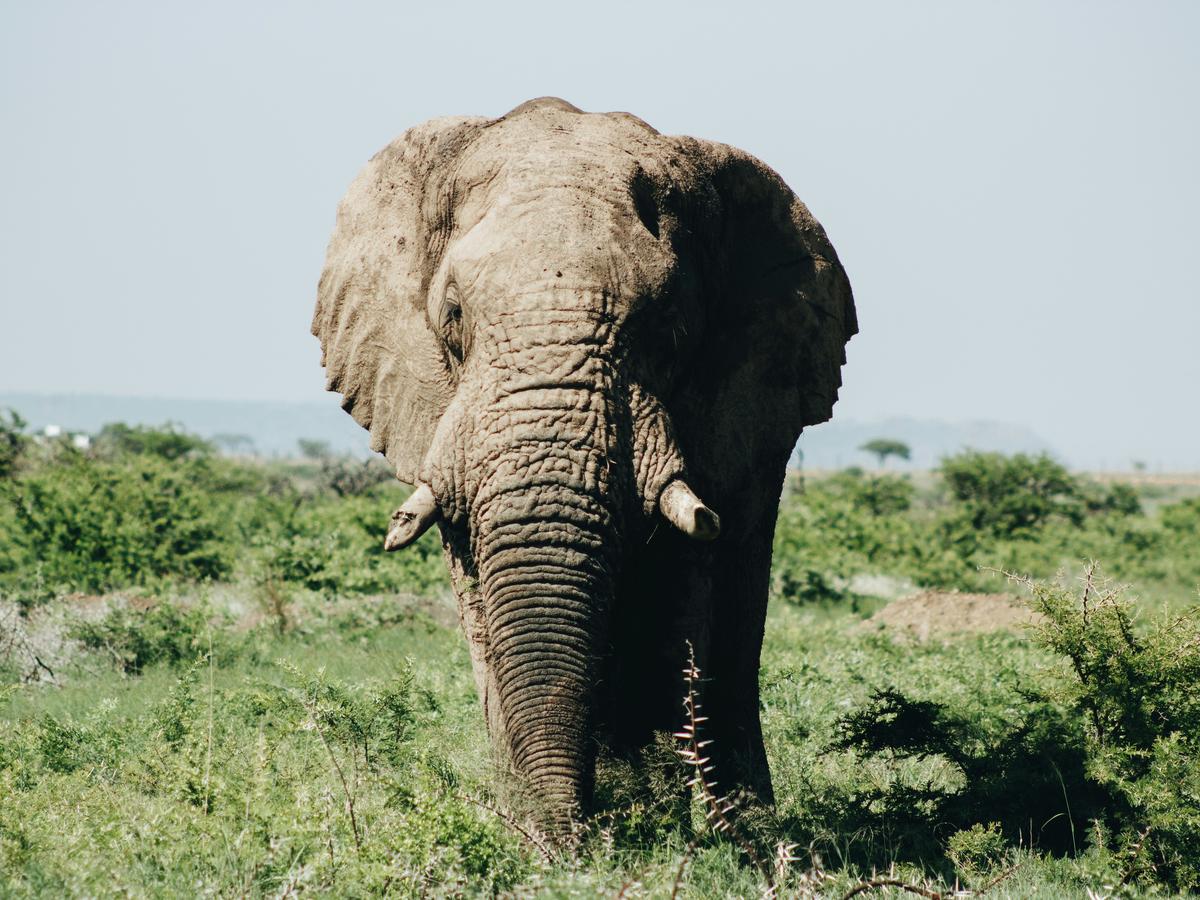} &
      \includegraphics[width=\linewidth]{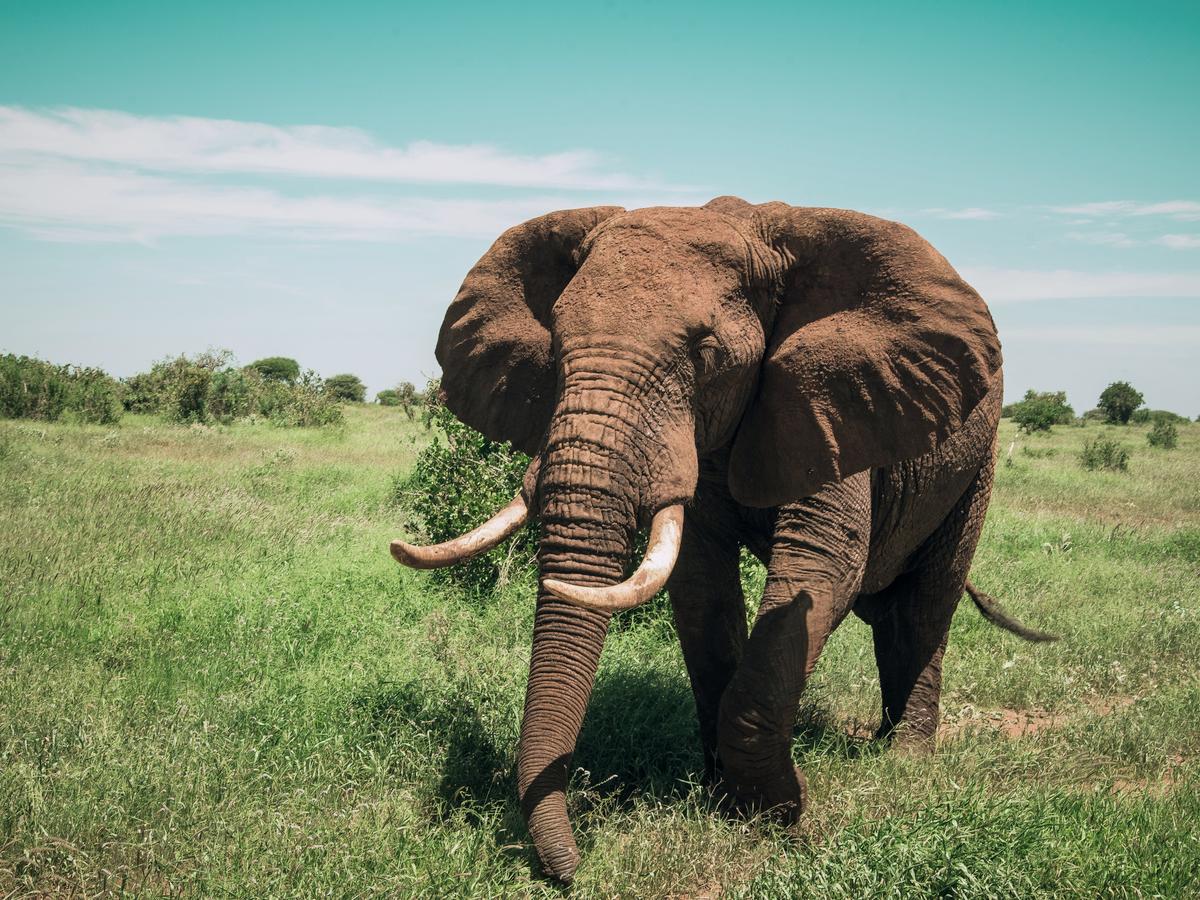} &
      \includegraphics[width=\linewidth]{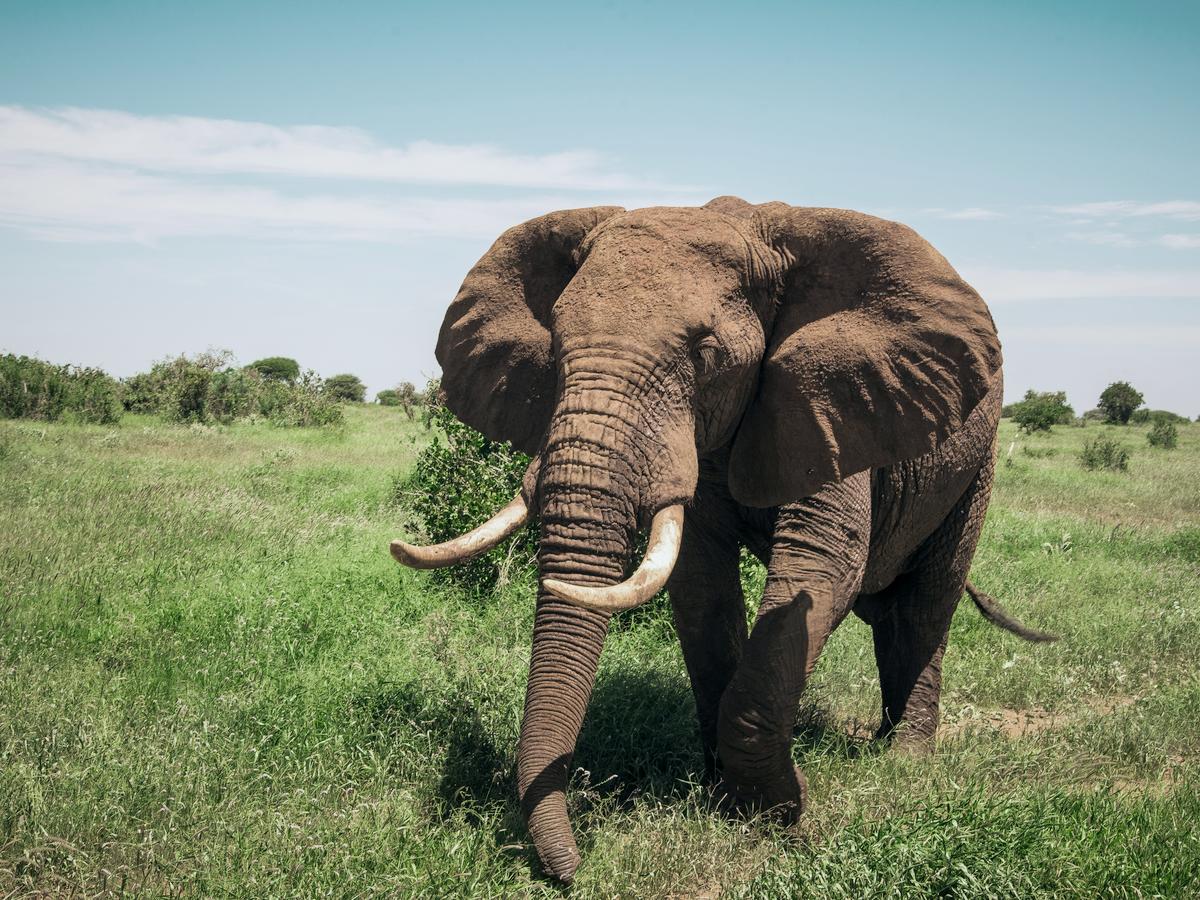}\\

      \scriptsize Content & 
      \scriptsize Style & 
      \scriptsize w/o semantic & 
      \scriptsize w/ semantic
    \end{tabularx}
    \caption{\textbf{Ablation on Semantic Priors.} We compare the transferred images without (w/o) and with (w/) semantic priors. The results demonstrate that semantic guidance significantly improves color consistency.}
    \label{fig:ab_semantic}
  \end{minipage}
\end{figure*}

\noindent\textbf{Effectiveness of Explicit Semantic Alignment.}
\cref{fig:ab_semantic} visualizes the impact of semantic alignment in ColorFM-O. 
Without semantic guidance, the model fails to establish correct semantic correspondence, resulting in imprecise color mapping (\eg, the sky color is inconsistently transferred).
In contrast, enabling explicit alignment enforces region-aware transfer, ensuring that colors are accurately mapped to semantically corresponding regions.

\begin{table}[t]
    \centering
    \caption{\textbf{Effect of Hierarchical Color Coupling.} 
    We compare our strategy with Rectified Flow (the prefix `2-' indicates one reflow step) and OT baselines, and analyze the impact of different values of $D_{max}$. Time is measured per image pair.}
    \label{tab:HCC_ablation}
\setlength{\tabcolsep}{5pt} 
    \resizebox{0.75\linewidth}{!}{%
    \begin{tabular}{c|c|cccc}
        \toprule
         Strategy & $D_{max}$  & Style $\uparrow$ & Content $\uparrow$ & Lipschitz $\downarrow$ & Time (s) $\downarrow$\\
        \midrule
        1-Rectified Flow & - & 0.754 & 0.650 & 5.693 & 12.08\\
        2-Rectified Flow & - & 0.738 & 0.717 & 3.568 & 28.50\\
        Mini-Batch OT & - & 0.574 &  0.710 & 2.624 & 51.48\\        
        \midrule     
        \multirow{4}{*}{\makecell{Hierarchical  Color \\  Coupling}}
          & 0 &  0.750 &  0.644 & 5.811 & 18.89\\ 
          &  1 &  0.774 & 0.727 & 3.351 & 19.04\\ 
          &  3 &  0.779 & 0.739 & 2.937 & 19.26\\ 
          &  5 &  0.773 & 0.741 & 2.905 & 24.37\\       
        \bottomrule
    \end{tabular}%
    }
\end{table}

\begin{figure}[t] 
    \centering
    \begin{minipage}[b]{0.41\linewidth}
        \centering
        \includegraphics[width=\linewidth, keepaspectratio]{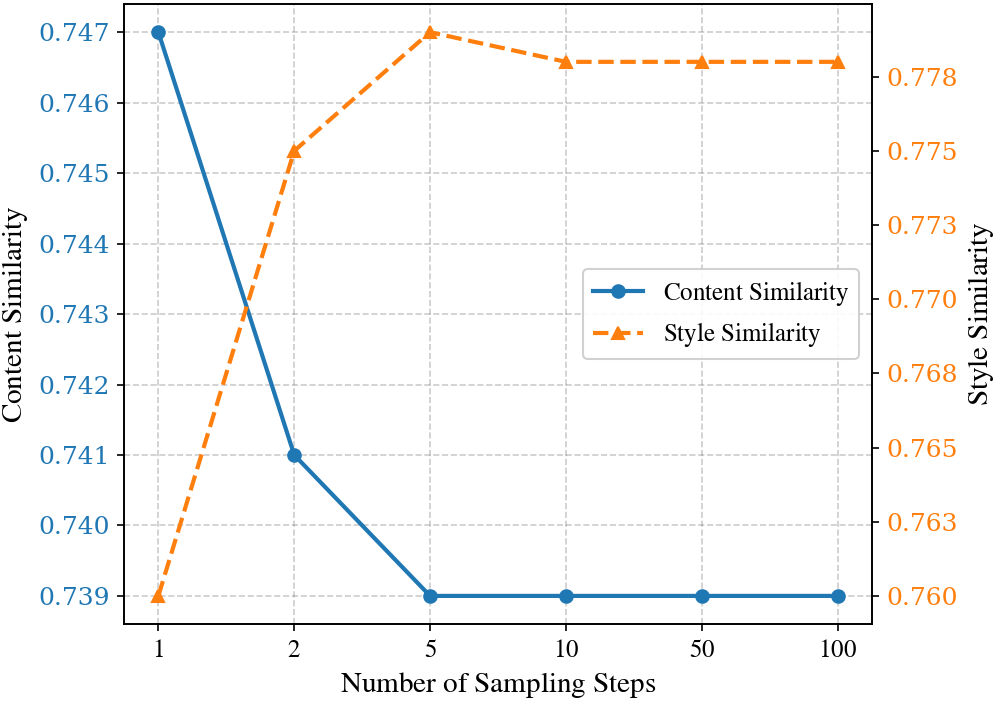}
        \caption{\textbf{Impact of Inference Steps.} Metrics saturate at $T \geq 5$, suggesting quasi-linear trajectories. }
        \label{fig:right_image}
    \end{minipage}
    \hfill 
  \begin{minipage}[b]{0.55\linewidth}
    \centering
    \setlength{\tabcolsep}{0.5pt} 
    \renewcommand{\arraystretch}{0.5} 
    \begin{tabularx}{\linewidth}{Y Y Y} 

      \includegraphics[width=\linewidth]{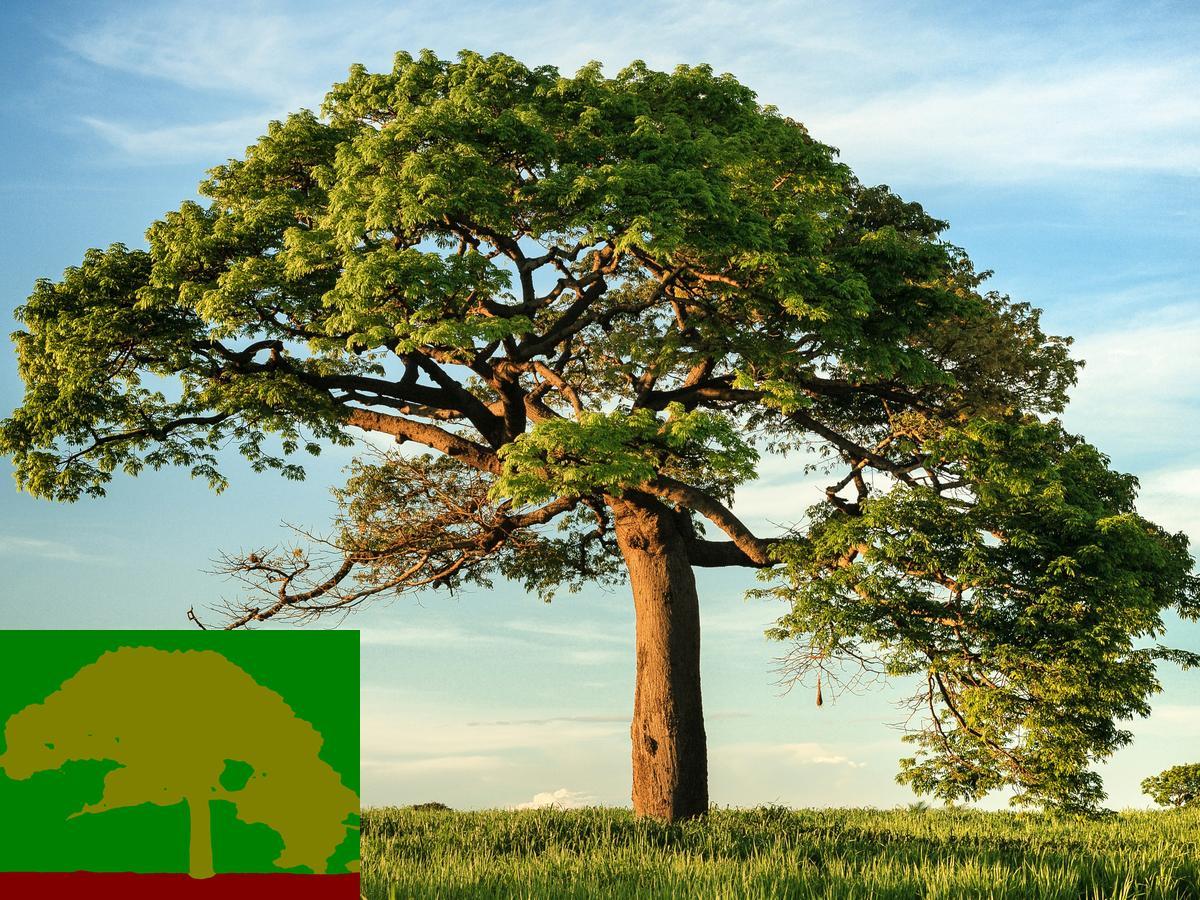} &
      \includegraphics[width=\linewidth]{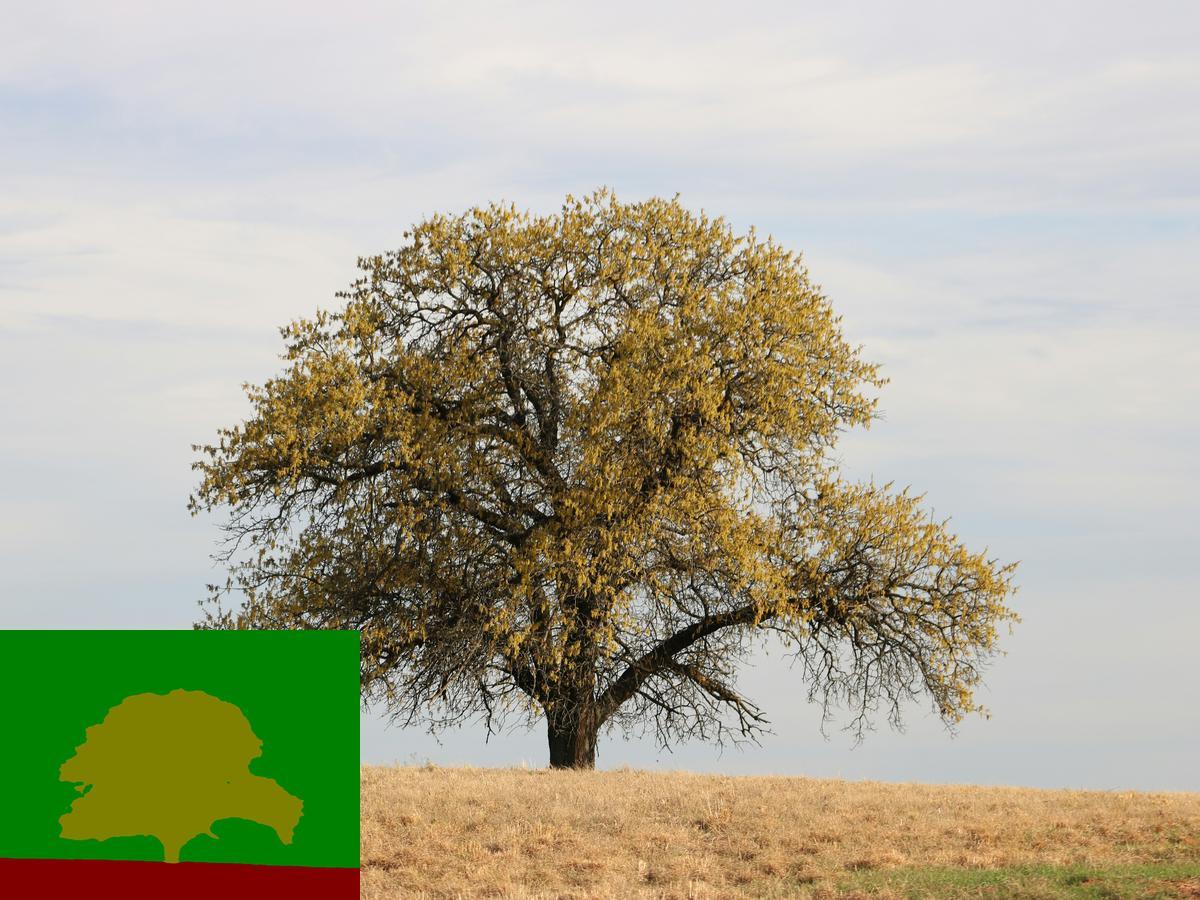} &
      \includegraphics[width=\linewidth]{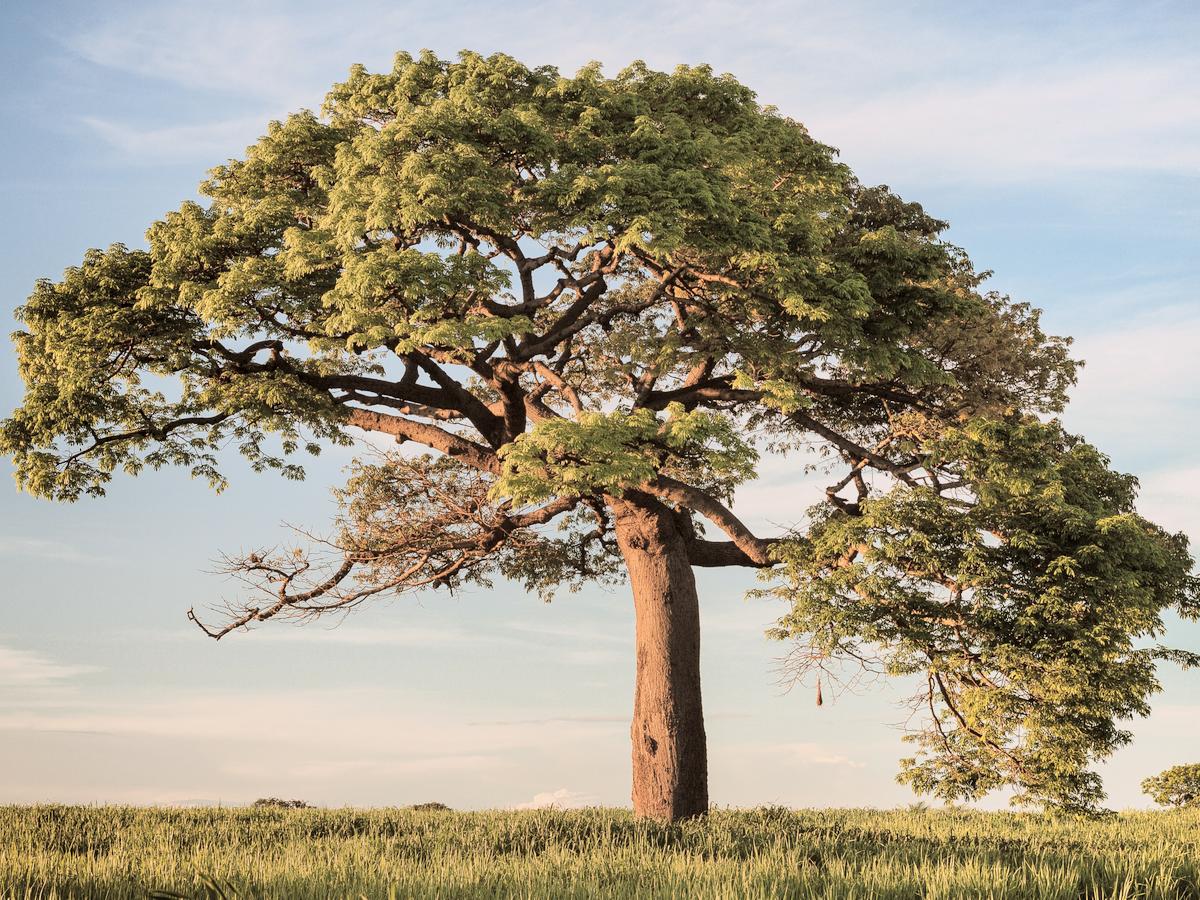}\\

      \includegraphics[width=\linewidth]{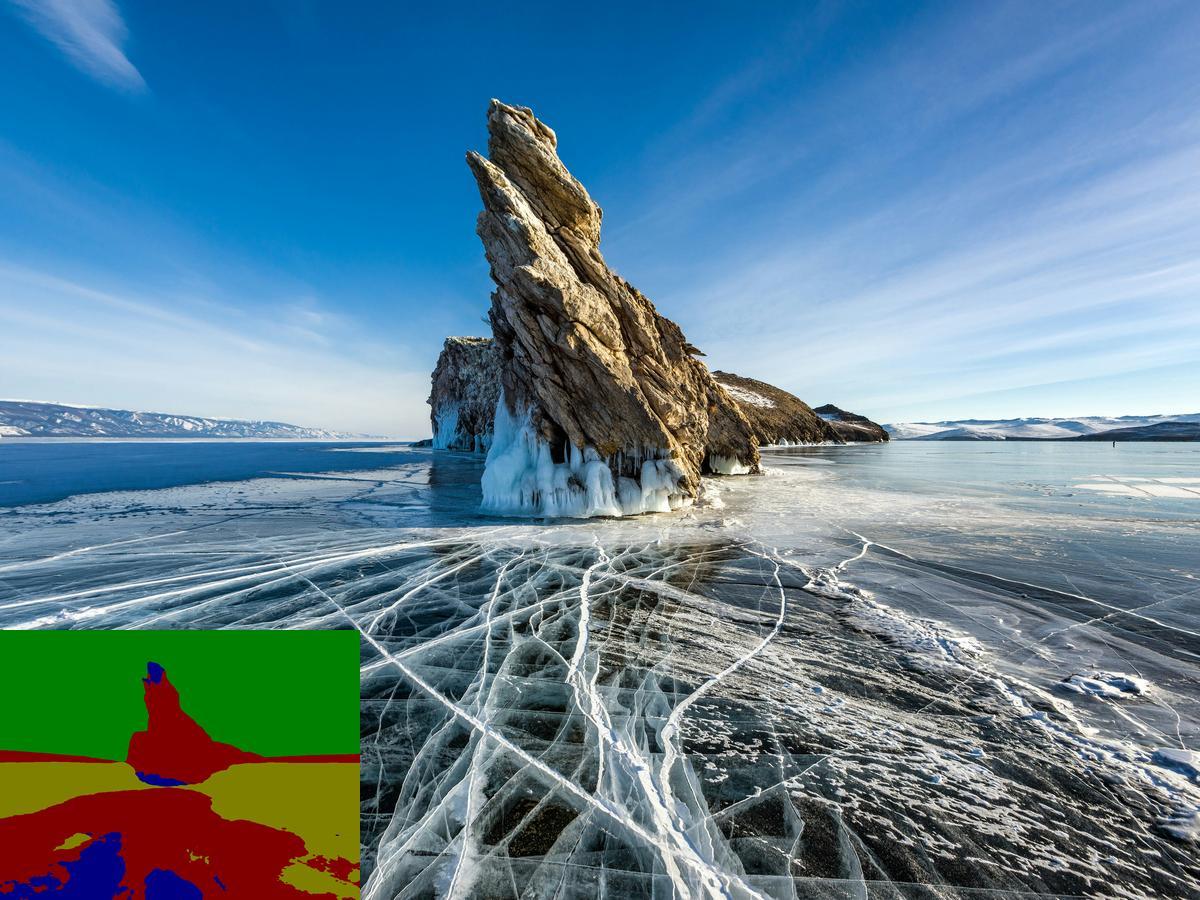} &
      \includegraphics[width=\linewidth]{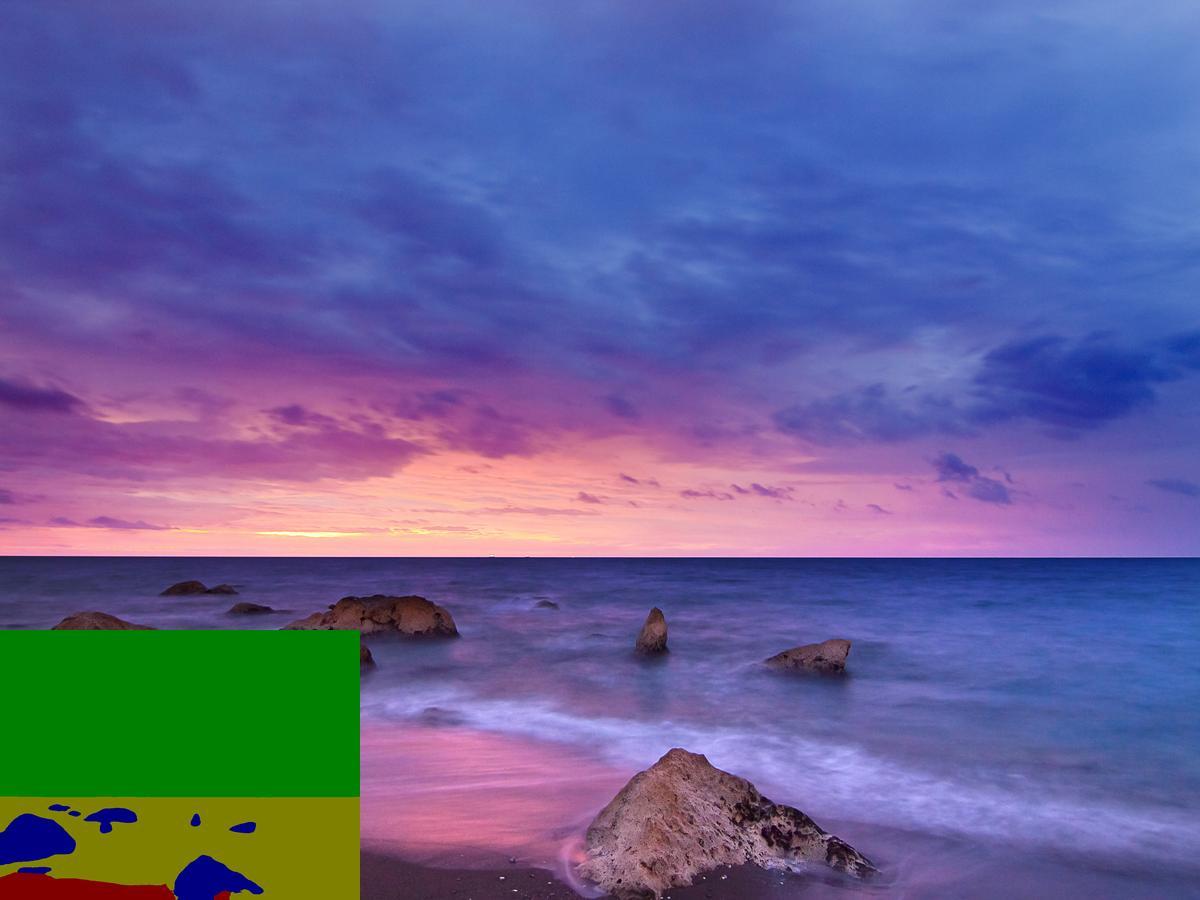} &
      \includegraphics[width=\linewidth]{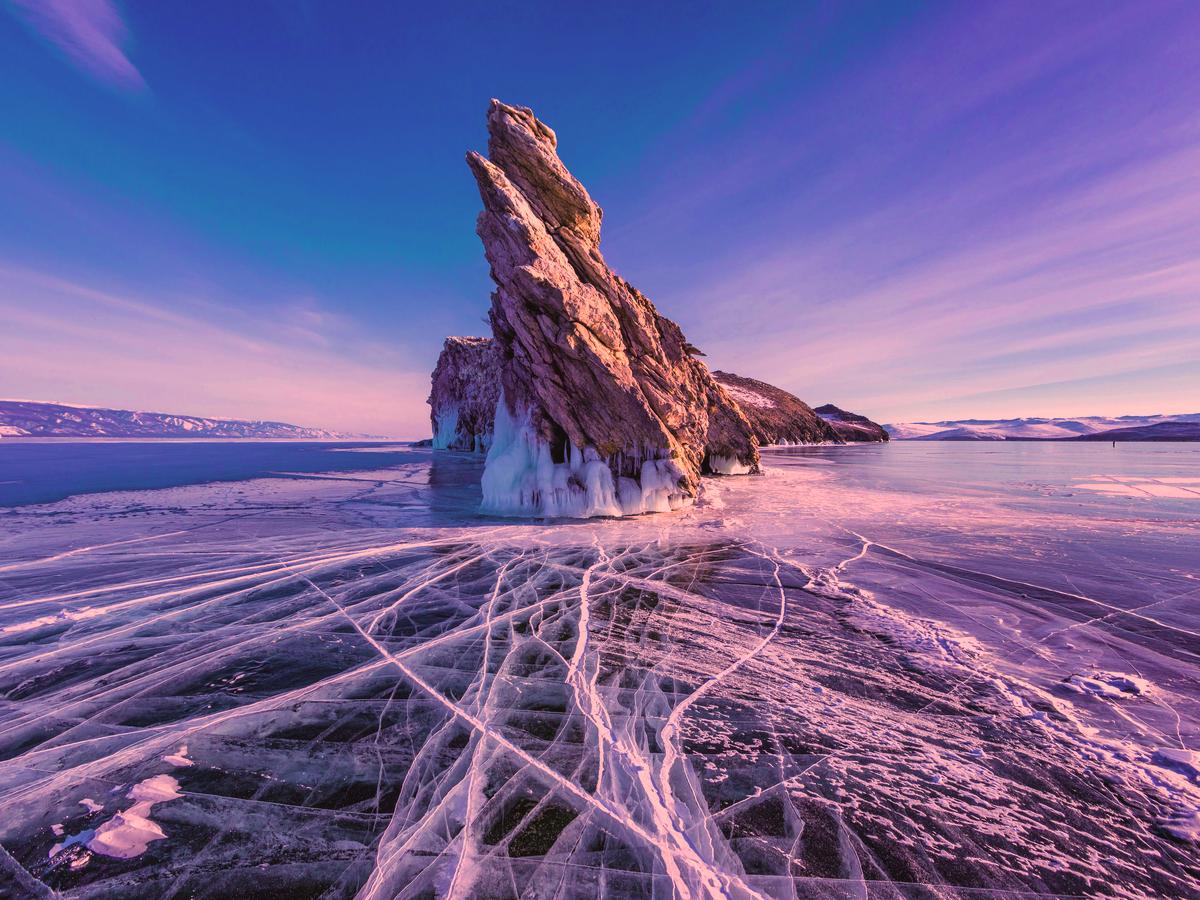}\\

      \scriptsize Content & 
      \scriptsize Style & 
      \scriptsize ColorFM-O \\
      
    \end{tabularx}
    \caption{\textbf{Robustness to Imperfect Masks.} ColorFM-O yields seamless, artifact-free transfer results even with coarse or inaccurate masks.}
    \label{fig:comparison_masks}
  \end{minipage}
\end{figure}

\noindent\textbf{Effect of Hierarchical Color Coupling.} \cref{tab:HCC_ablation} investigates the impact of the HCC strategy. We randomly select a subset of 20 images (380 pairs) from the test set for evaluation.
While 2-Rectified Flow straightens trajectories, it inherits suboptimal style mapping from the initial random coupling in 1-Rectified Flow, resulting in low style similarity.
Similarly, Mini-batch OT incurs high computational cost and suffers from severe statistical distortion. 
In contrast, our HCC strategy achieves a better balance. Increasing the subdivision depth $D_{max}$ from 0 to 3 reduces the Lipschitz value from \textbf{5.811} to \textbf{2.937}, indicating a smoother learned mapping. Further increasing $D_{max}$ to 5 raises computational cost and leads to degraded style similarity. Thus, we adopt $D_{max}=3$ as the default setting.

\noindent\textbf{Number of Sampling Steps.}
Following the experimental setup in \cref{tab:HCC_ablation}, we evaluate the impact of the number of ODE integration steps.
As shown in \cref{fig:right_image}, the performance of ColorFM-O saturates at $T \geq 5$. We therefore adopt a 5-step sampling strategy for ODE integration.
 This rapid convergence suggests the quasi-linearity of the learned trajectories. To verify this quantitatively, we compute the path length ratio over 100 sampling steps, yielding an average value of \textbf{1.009}. This inherent linearity minimizes rectification complexity, serving as a key prerequisite for the efficient one-step inference in ColorFM-L.

\noindent\textbf{Impact of ColorFM-L Components.}
\cref{tab:ablation_rColorFM} analyzes the architectural components of ColorFM-L. Estimating the transport parameters independently (Row b) outperforms direct mapping (Row a), highlighting the structural advantage of the bidirectional formulation. We further add a fixed-state baseline following~\cite{ke2023neural} to isolate the effects of static and implicit state modeling. The implicit variants further boost performance: specifically, Cross-Attention (Row e) demonstrates stronger modeling capability than simple concatenation (Row d). We therefore adopt Row e as the default design for ColorFM-L.

\begin{table}[t]
\centering
\caption{\textbf{Ablation Study on ColorFM-L Components.}
We investigate the impact of the transport scheme and implicit state modeling.}
\label{tab:ablation_rColorFM}

\resizebox{0.95\linewidth}{!}{% 
\begin{tabular}{c|ccc|ccc}
    \toprule
    ID & State Type & Fusion Strategy & Transport Scheme 
    & Style $\uparrow$ & Content $\uparrow$ & Dist. to Ideal $\downarrow$  \\
    \midrule
    a & None    & -          & Direct   & 0.785 & 0.738 & 0.339 \\
    b & None    & -          & Bidirect & 0.809 & 0.736 & 0.326 \\
    c & Fixed   & -          & Bidirect & 0.812 & 0.724 & 0.334    \\
    d & Implicit & Concat     & Bidirect & 0.818 & 0.723 & 0.331 \\
    e & Implicit & Cross-Attn & Bidirect & 0.825 & 0.732 & 0.320 \\
    \bottomrule
\end{tabular}% 
}

\end{table}

\section{Discussion}
\begin{figure*}[t]
  \centering
  \begin{minipage}[b]{0.56\linewidth}
    \centering
    \setlength{\tabcolsep}{0.5pt} 
    \renewcommand{\arraystretch}{0.5} 
    \begin{tabularx}{\linewidth}{Y Y Y Y} 

      \includegraphics[width=\linewidth]{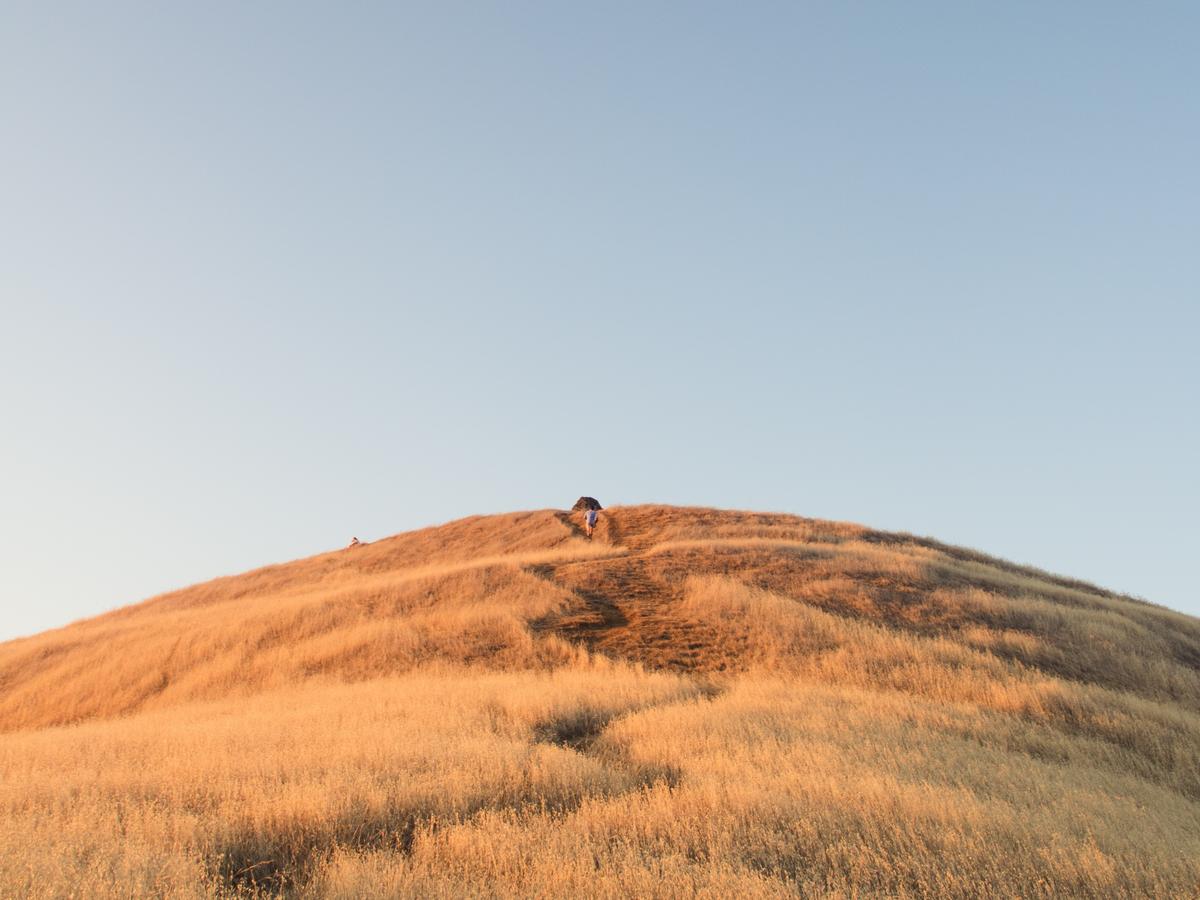} &
      \includegraphics[width=\linewidth]{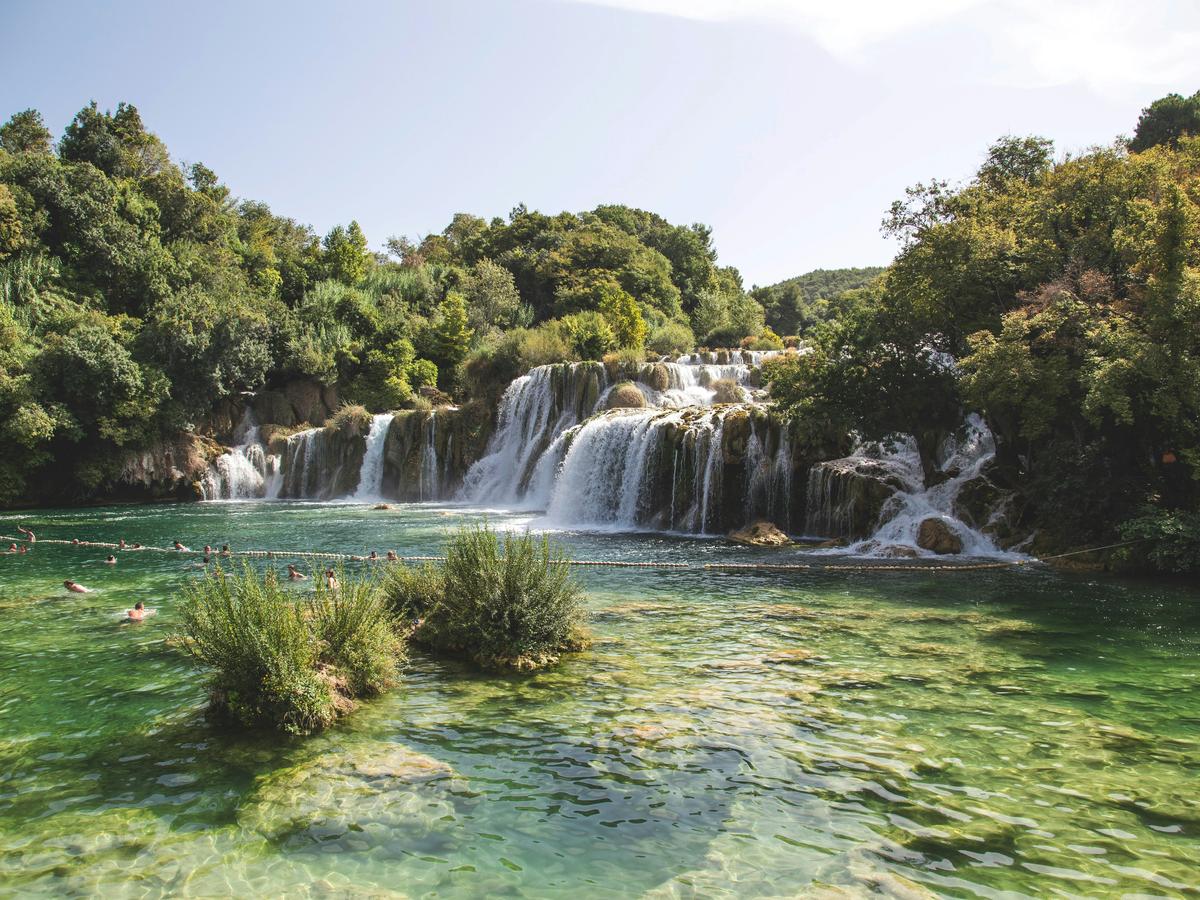} &
      \includegraphics[width=\linewidth]{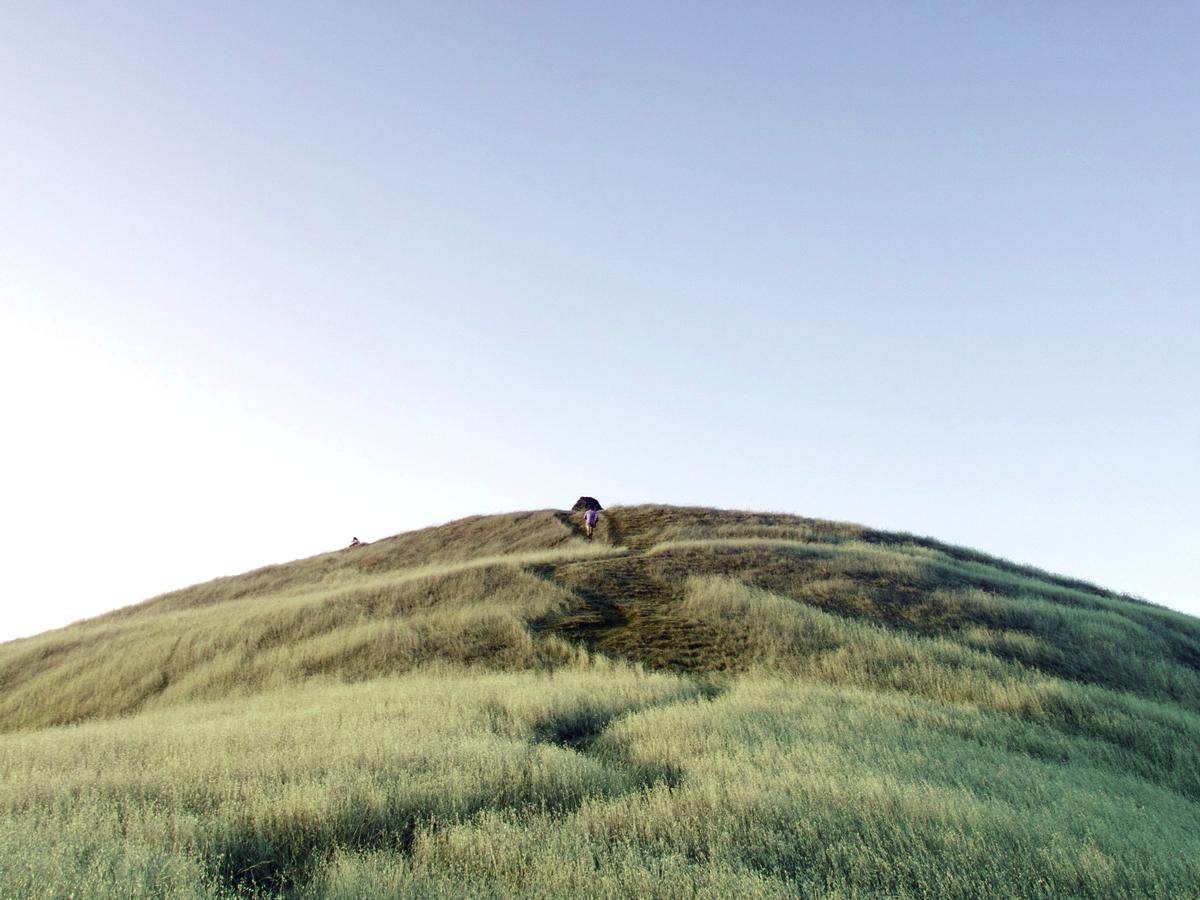} &
      \includegraphics[width=\linewidth]{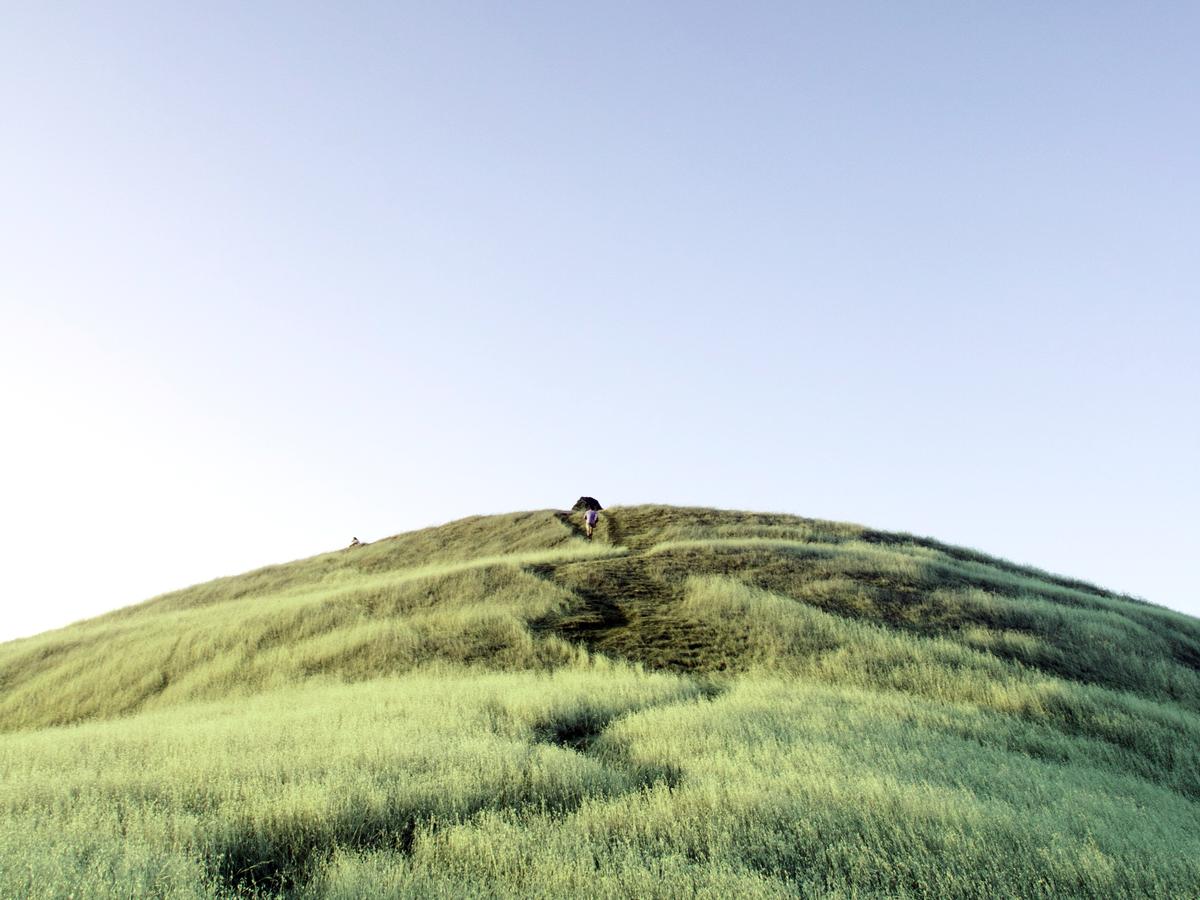}\\

      \includegraphics[width=\linewidth]{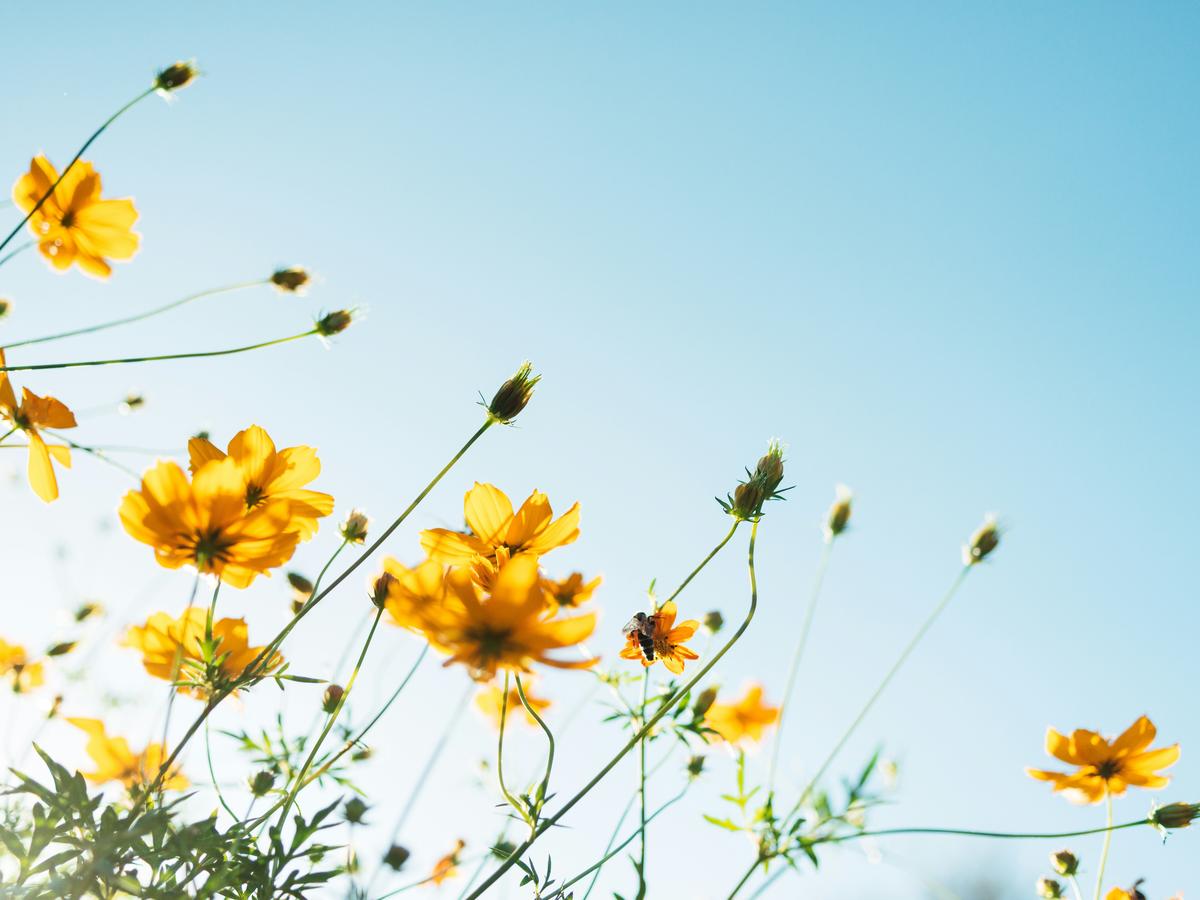} &
      \includegraphics[width=\linewidth]{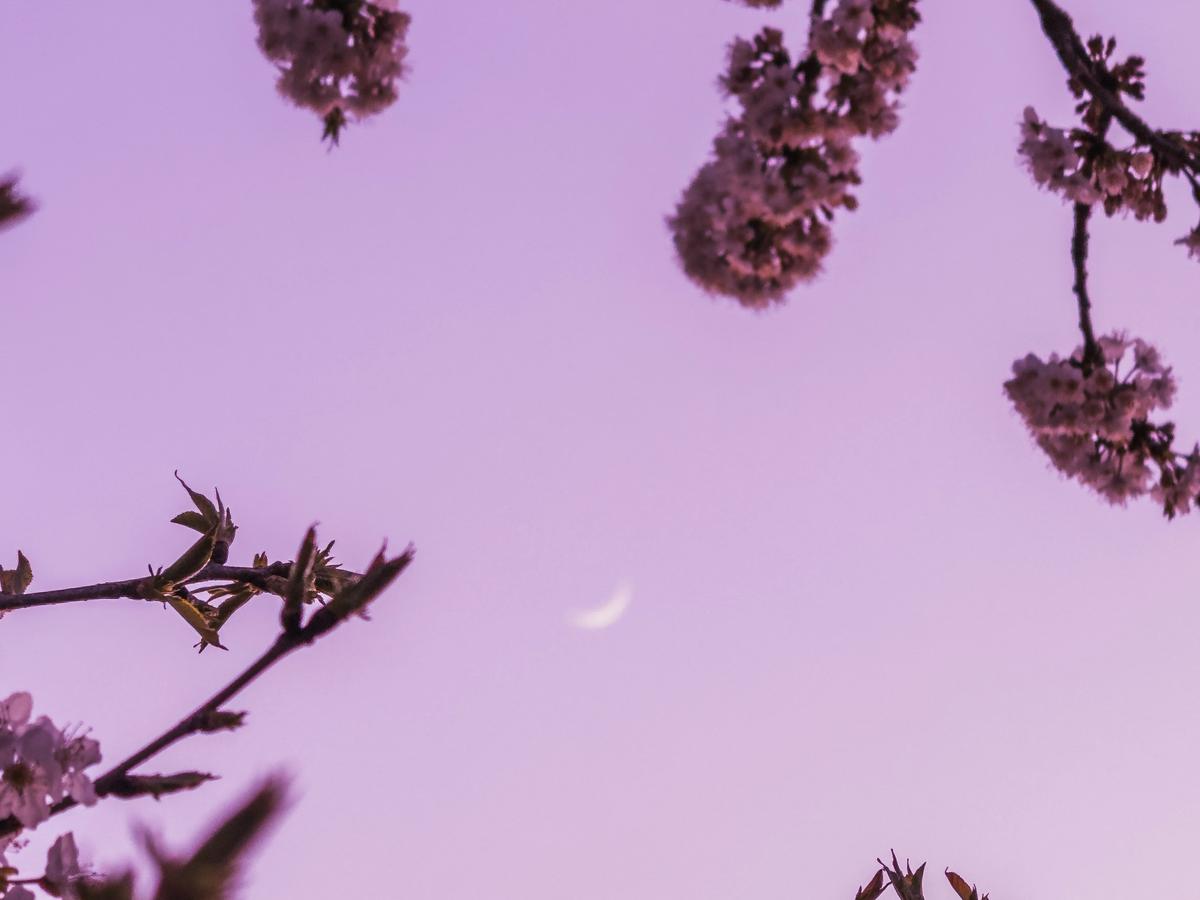} &
      \includegraphics[width=\linewidth]{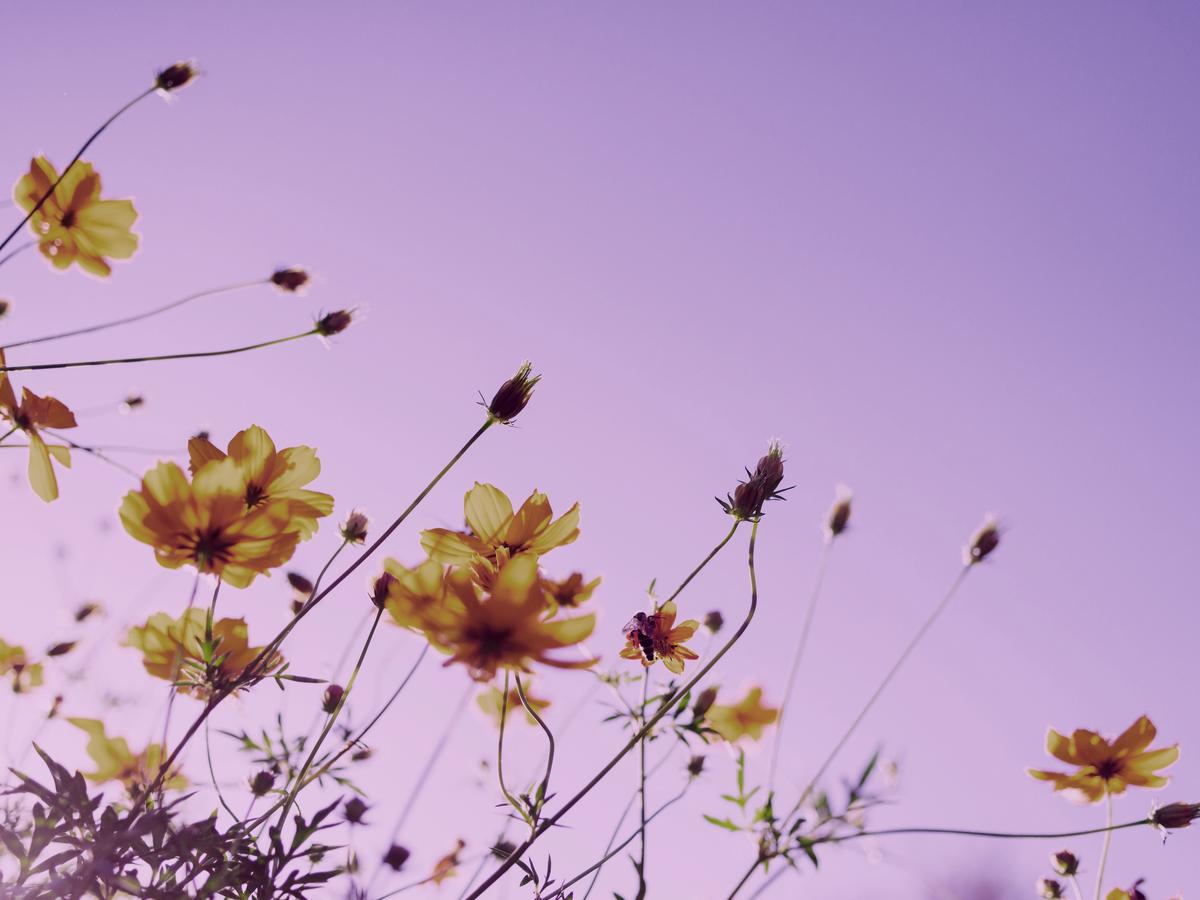} &
      \includegraphics[width=\linewidth]{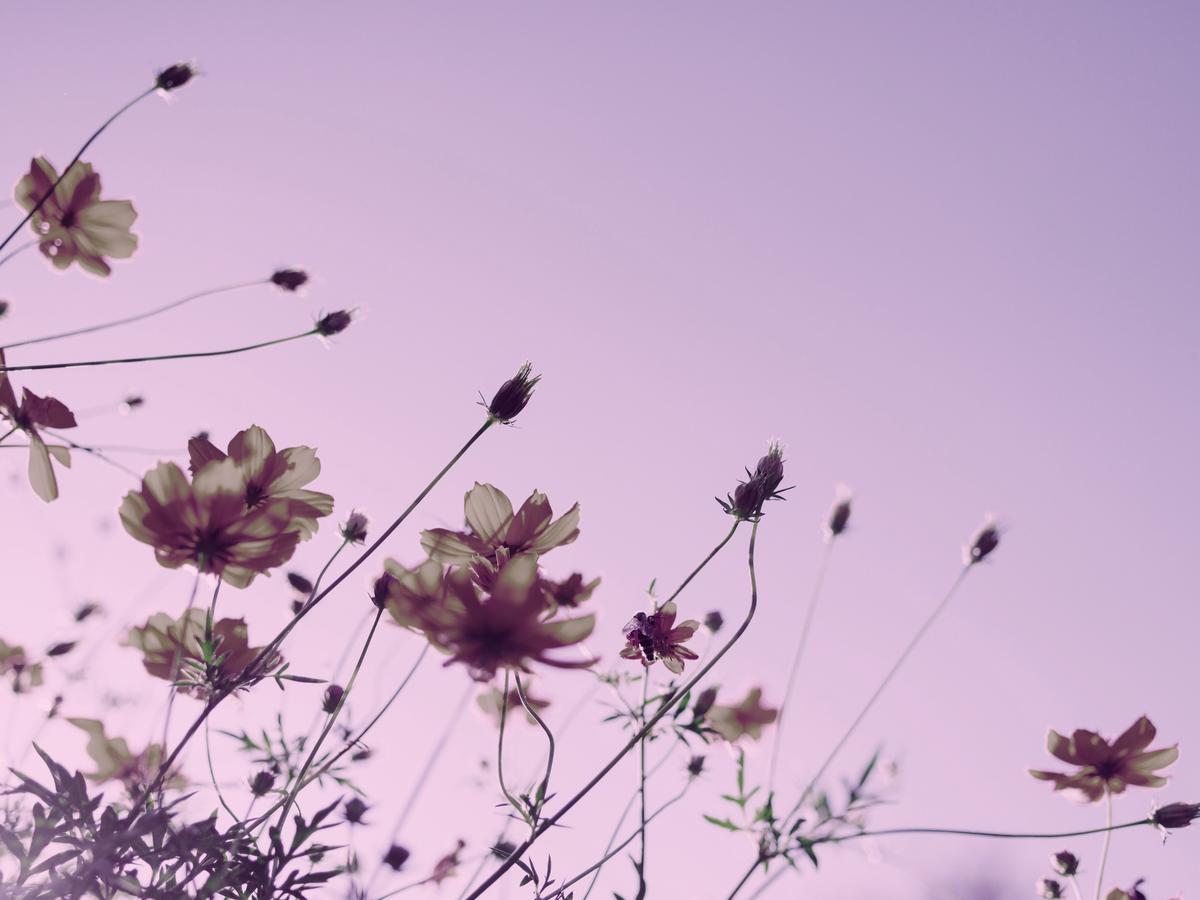}\\

      \scriptsize Content & 
      \scriptsize Style & 
      \scriptsize ColorFM-O & 
      \scriptsize ColorFM-L
      
    \end{tabularx}
    \caption{\textbf{Comparison Between ColorFM-O and ColorFM-L.} While comparable in simple cases (top), ColorFM-L outperforms ColorFM-O in complex scenes (bottom).}
    \label{fig:comparison_ir}
  \end{minipage}
  \hfill
  \begin{minipage}[b]{0.40\linewidth}
    \centering
    \setlength{\tabcolsep}{0.5pt} 
    \renewcommand{\arraystretch}{0.5} 
    \begin{tabularx}{\linewidth}{Y Y Y} 

      \includegraphics[width=\linewidth]{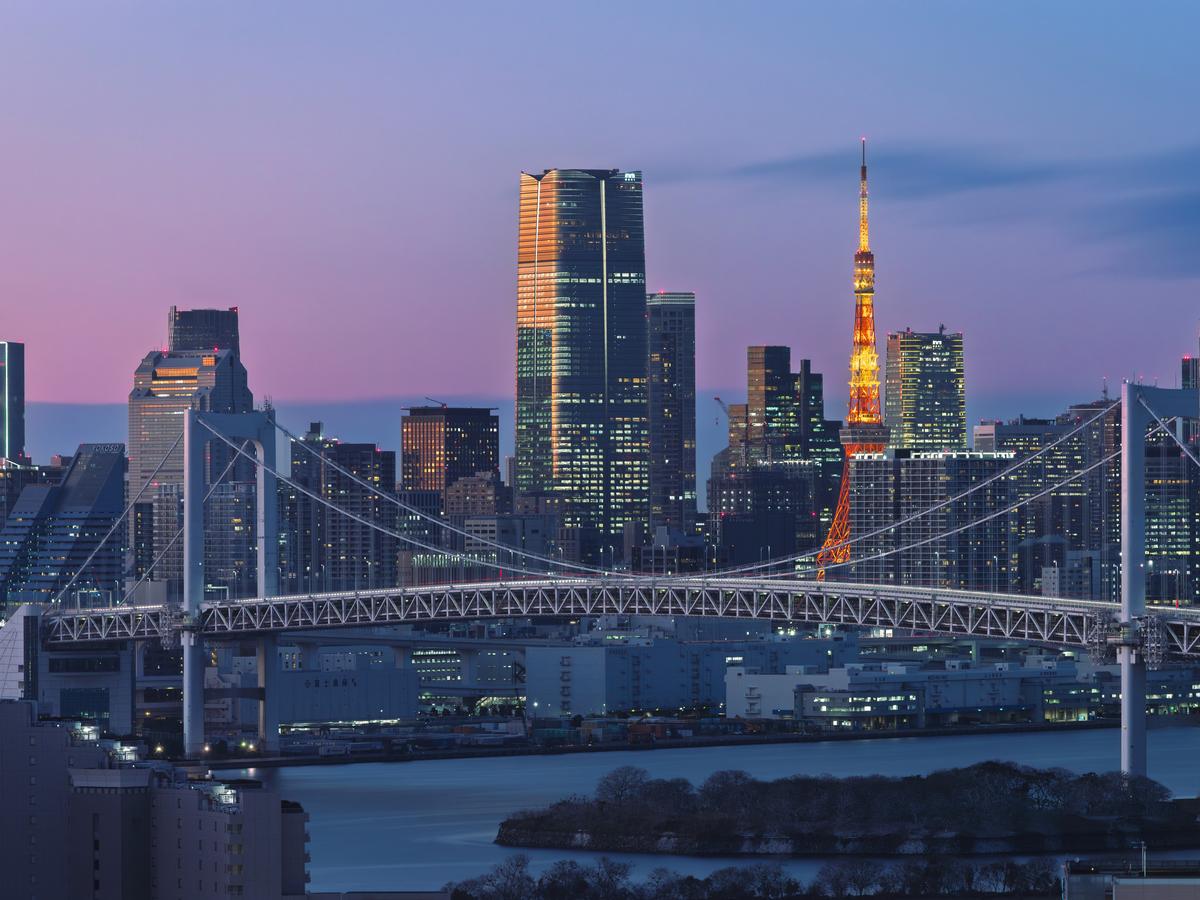} &
      \includegraphics[width=\linewidth]{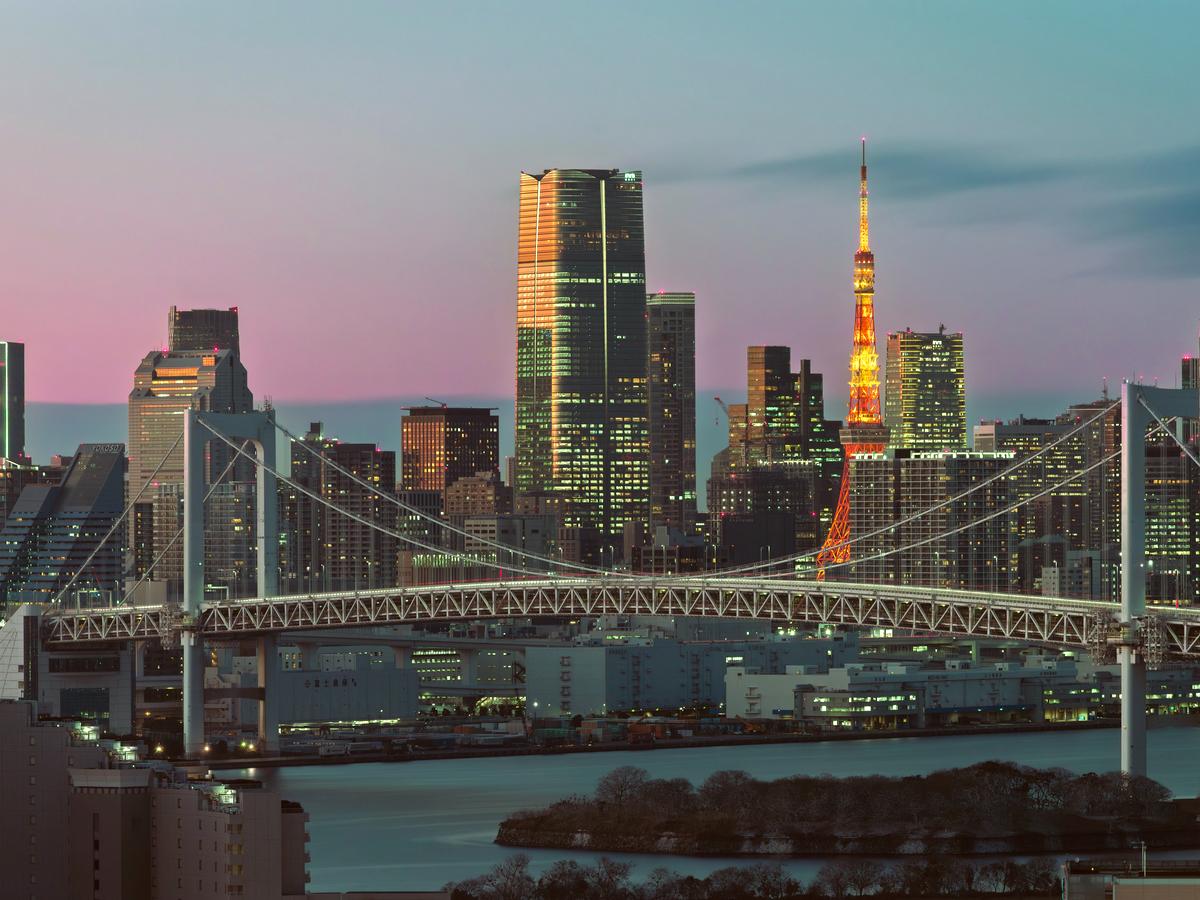} &
      \includegraphics[width=\linewidth]{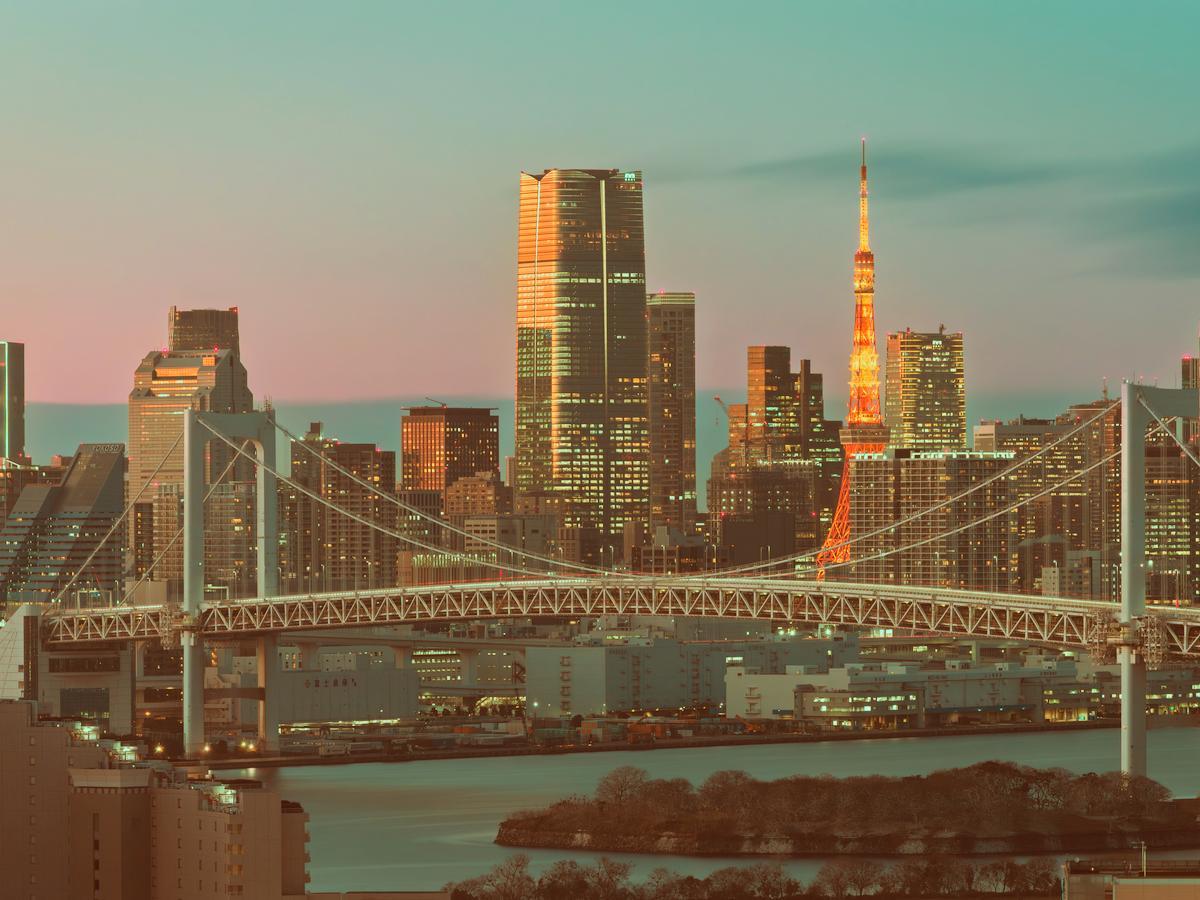}\\
      
      \scriptsize $I_A$ & 
      \scriptsize $Z_A$ & 
      \scriptsize $I_{A \to B}$ \\ 
      
      \includegraphics[width=\linewidth]{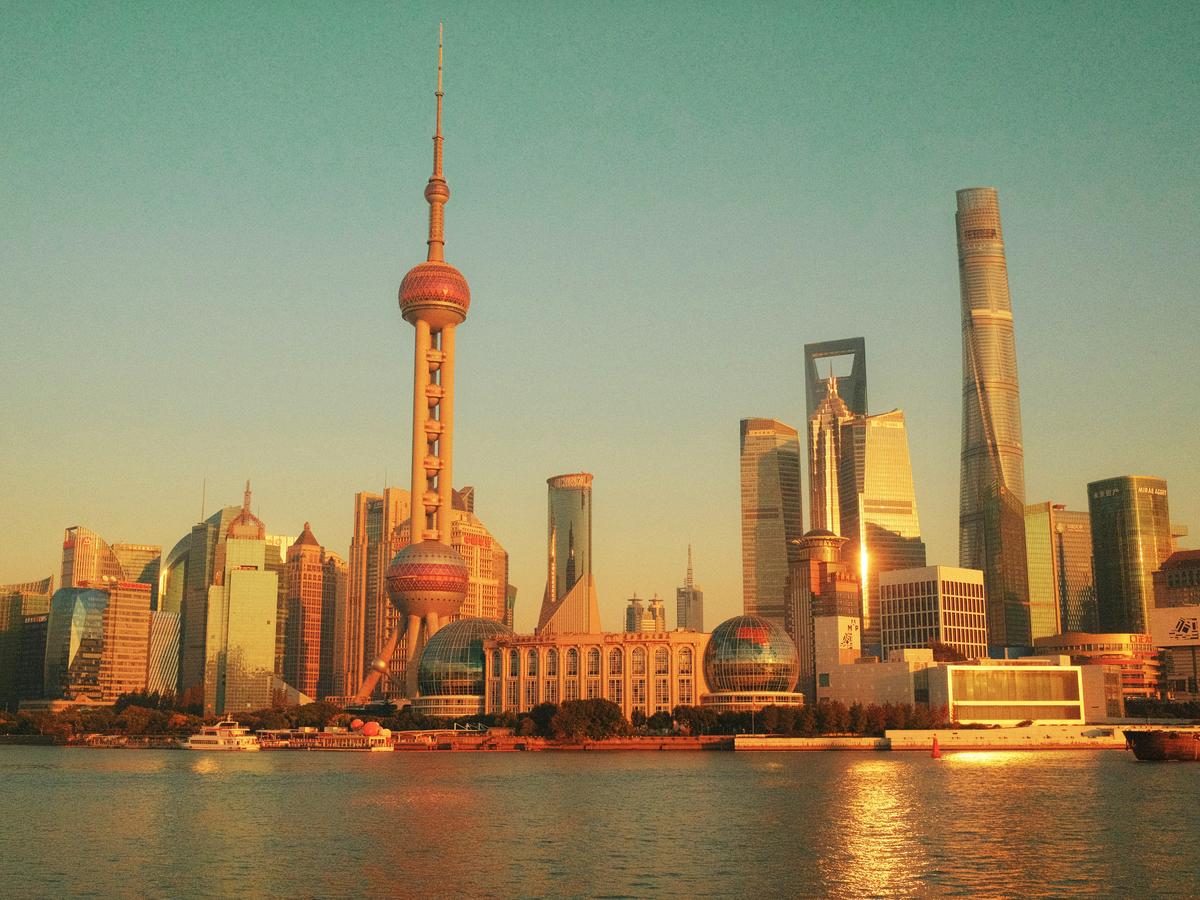} &
      \includegraphics[width=\linewidth]{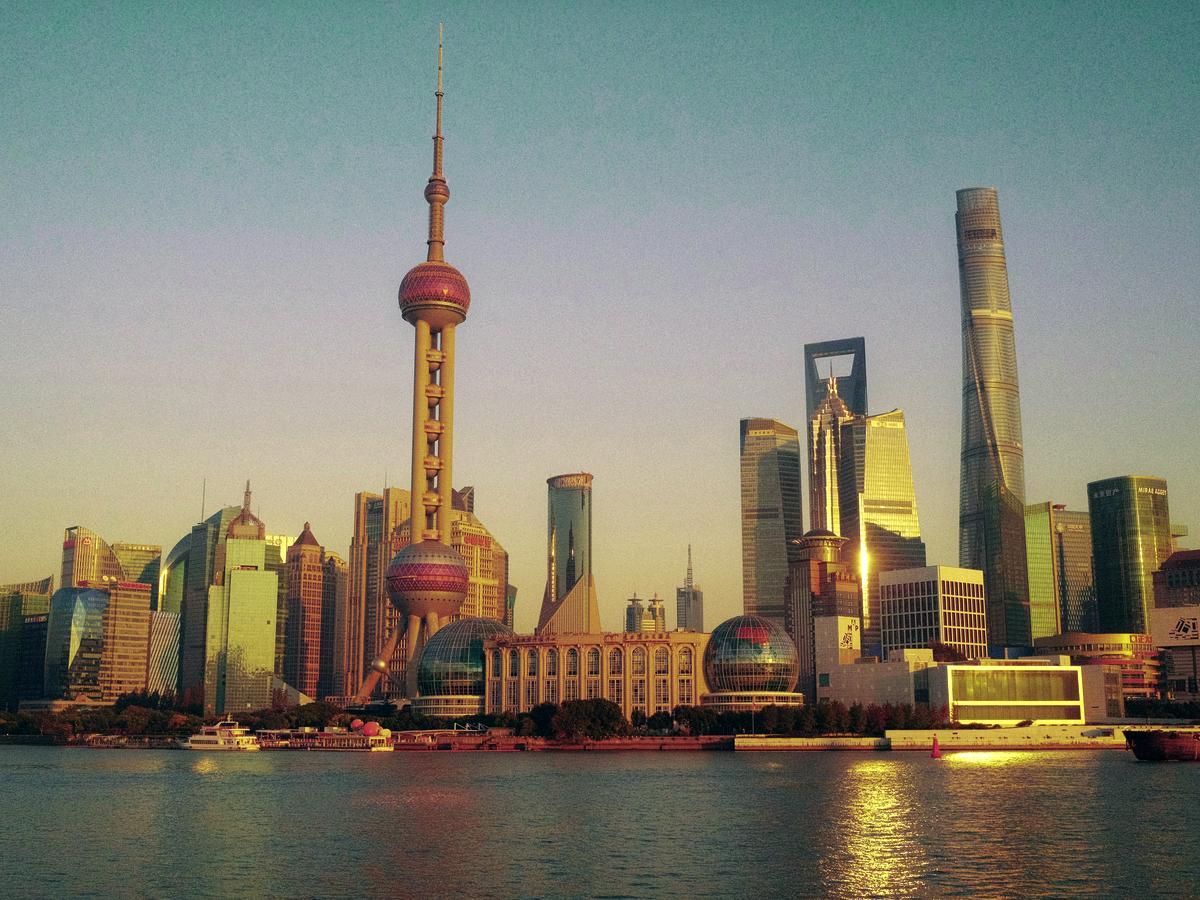} &
      \includegraphics[width=\linewidth]{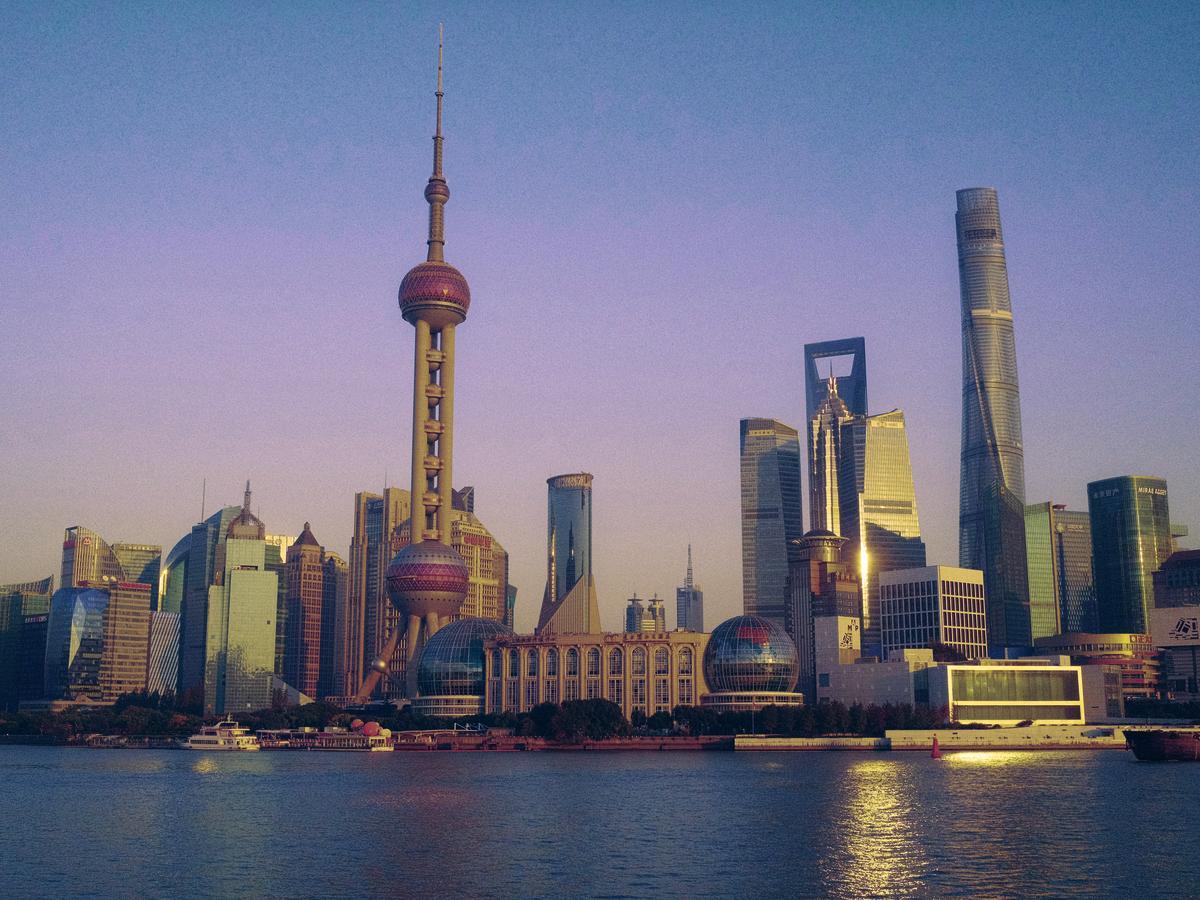}\\

      \scriptsize $I_B$ & 
      \scriptsize $Z_B$ & 
      \scriptsize $I_{B \to A}$ 
      
    \end{tabularx}
    \caption{\textbf{Implicit State Visualization.} $Z_A$ and $Z_B$ share a unified color distribution while maintaining content structures.}
    \label{fig:state}
  \end{minipage}
\end{figure*}

\subsection{Robustness of ColorFM-O to Imperfect Masks}
\cref{fig:comparison_masks} demonstrates ColorFM-O's robustness to coarse and inaccurate masks. 
Since the masks only guide semantic distribution pairing instead of defining hard regional transformations, local mask errors do not directly introduce discontinuous mappings.
The final transfer is then governed by a globally unified velocity field, which is optimized over the entire image rather than applied independently to each region.
% Such unified modeling enforces cross-boundary consistency and implicitly smooths local mask errors, enabling visually seamless transfer despite imperfect masks. 
The quantitative results further show that ColorFM-O remains stable under moderate mask degradation, where the style similarity decreases only slightly from 0.745 to 0.731/0.722 under 20\%/40\% semantic misclassification, while a more noticeable drop to 0.695 appears only under severe degradation of 60\%. Full quantitative results are provided in Supp.~Tab.~3.

\begin{figure*}[t]
  \centering
  \setlength{\tabcolsep}{0pt} 
  \renewcommand{\arraystretch}{0.5} 
  
  \begin{tabularx}{\textwidth}{Y @{\hspace{1mm}} Y Y Y Y Y Y Y} 
  \includegraphics[width=\linewidth]{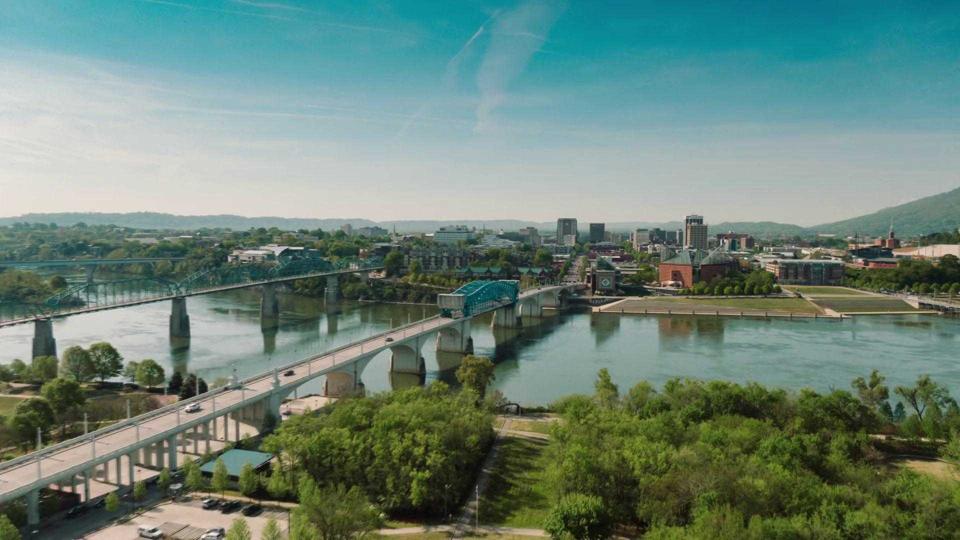} &
    \includegraphics[width=\linewidth]{pics/video_2/0000.jpg} & 
    \includegraphics[width=\linewidth]{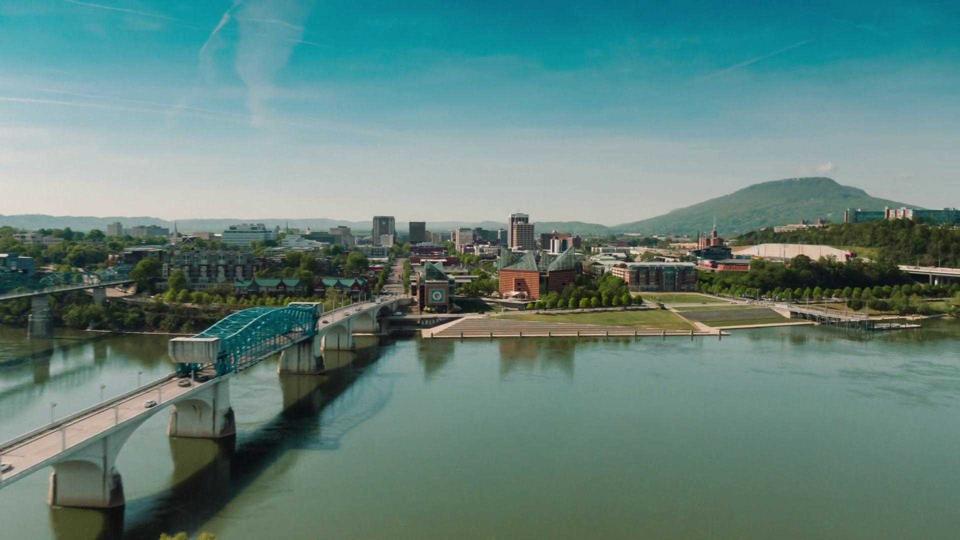} &
    \includegraphics[width=\linewidth]{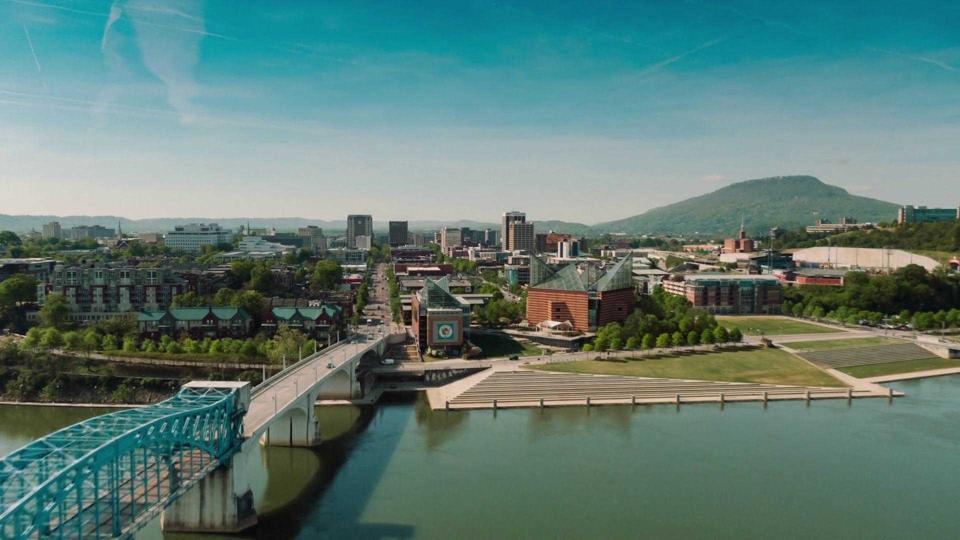} & 
    \includegraphics[width=\linewidth]{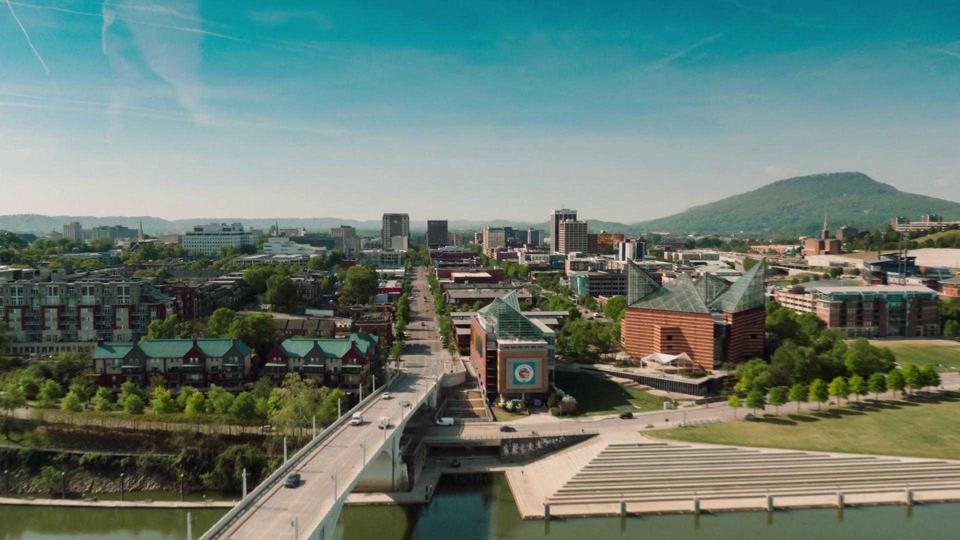} & 
    \includegraphics[width=\linewidth]{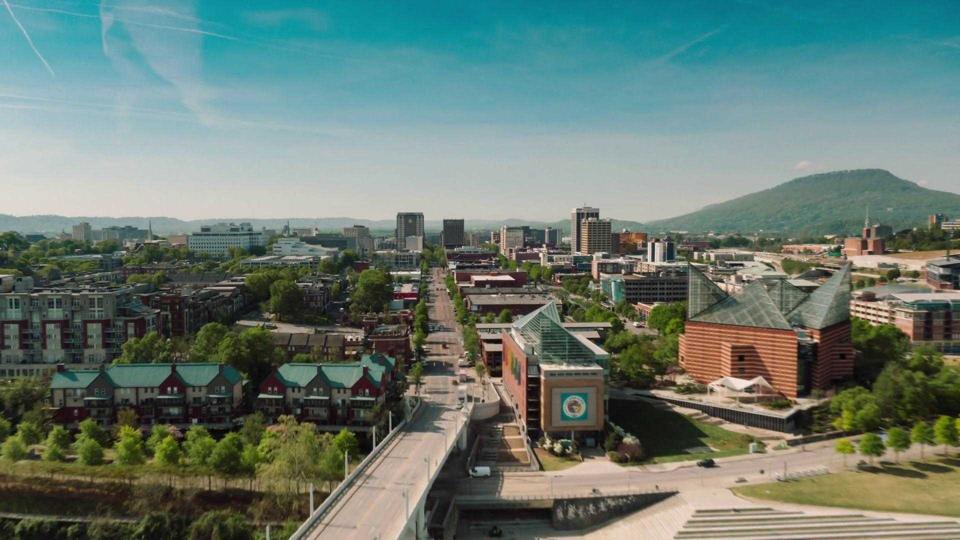} & 
    \includegraphics[width=\linewidth]{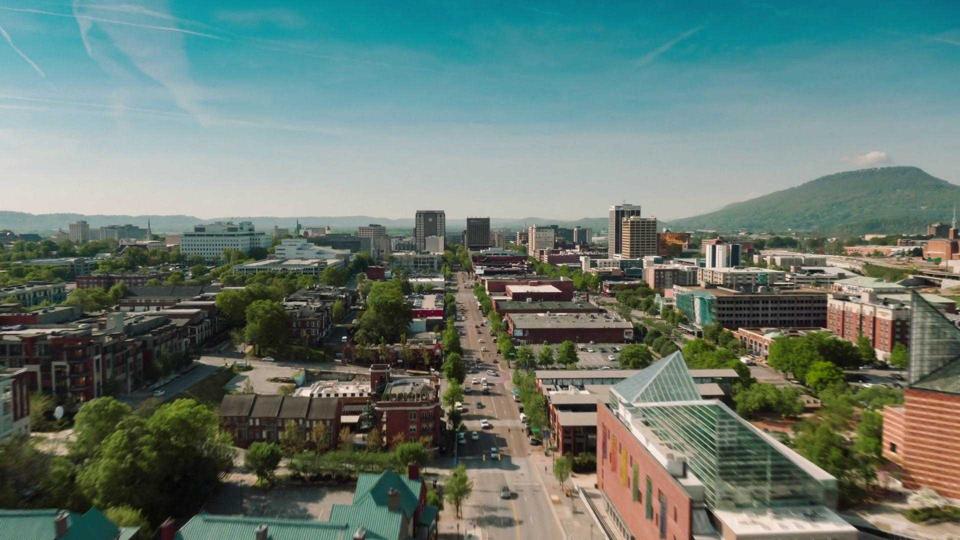} & 
    \includegraphics[width=\linewidth]{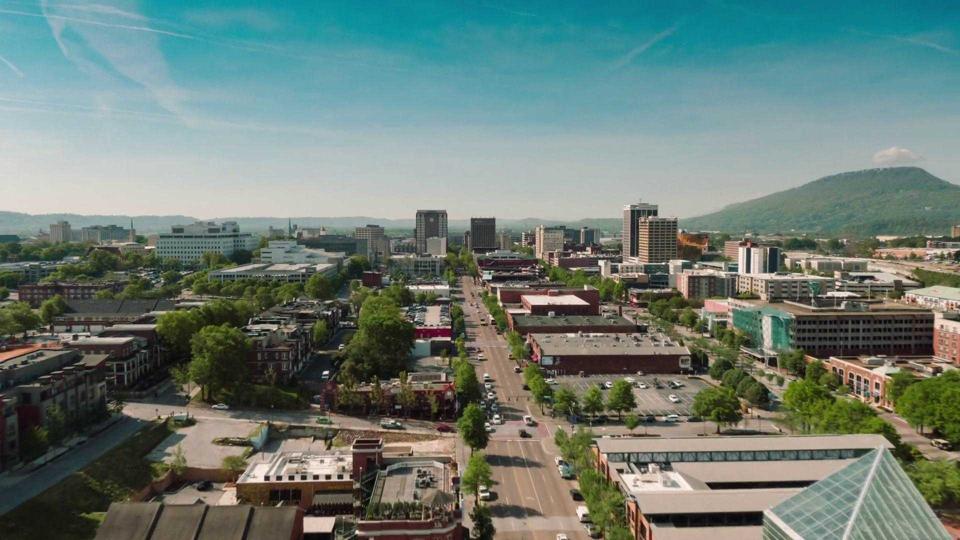} \\

    \centering\scriptsize  Content & \multicolumn{7}{c}{\scriptsize  Video Frames} \\

    \includegraphics[width=\linewidth]{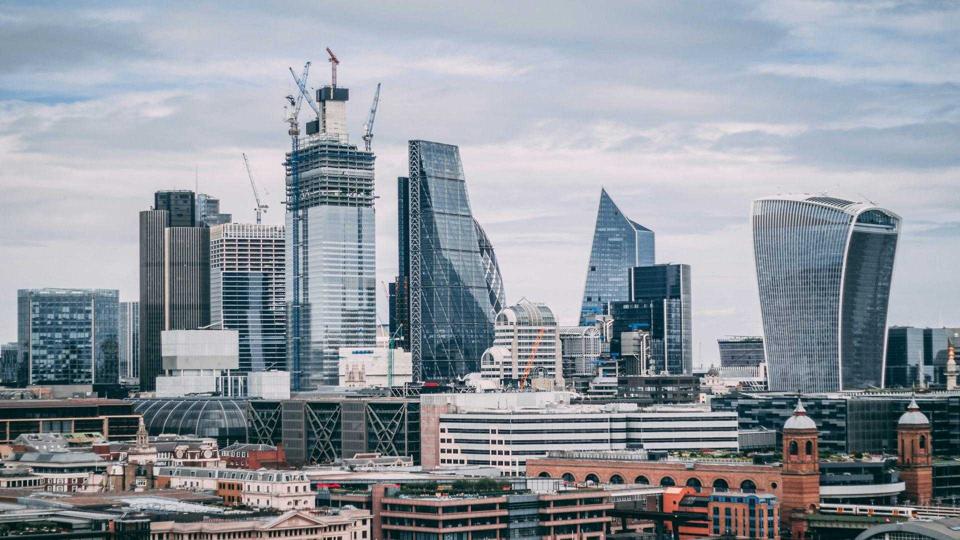} &
    \includegraphics[width=\linewidth]{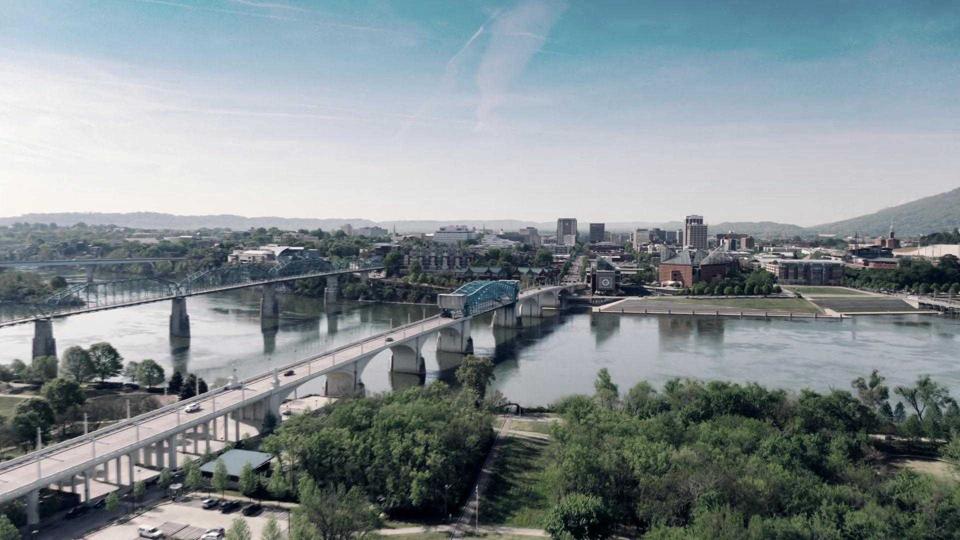} & 
    \includegraphics[width=\linewidth]{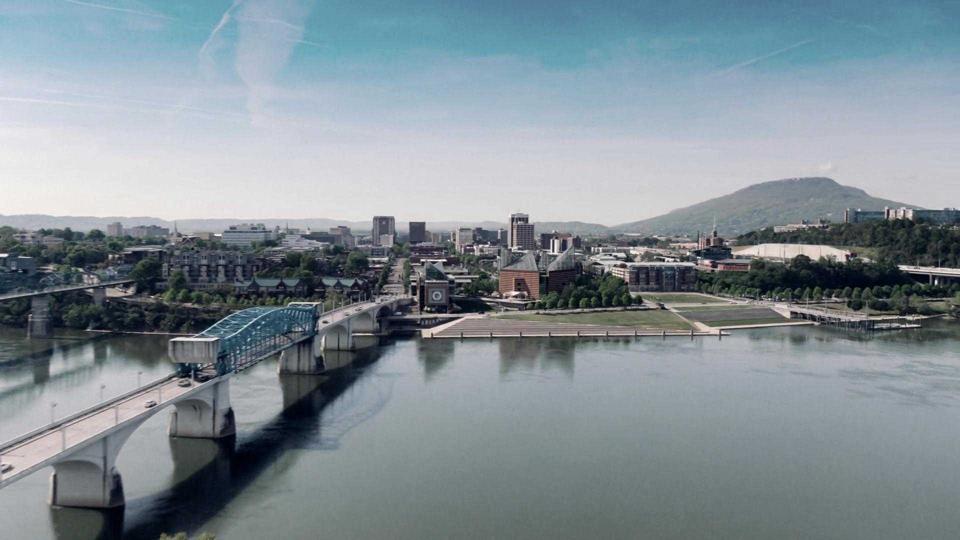} &
    \includegraphics[width=\linewidth]{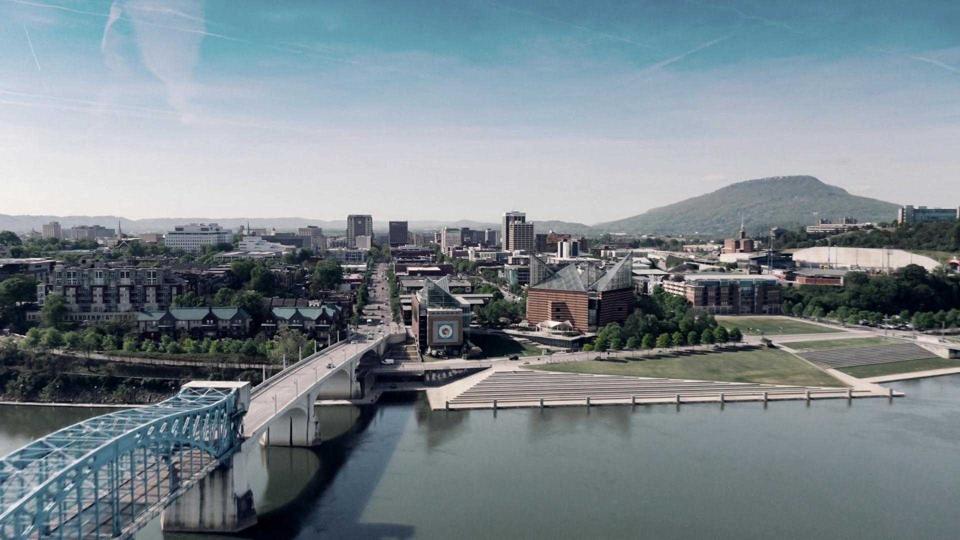} & 
    \includegraphics[width=\linewidth]{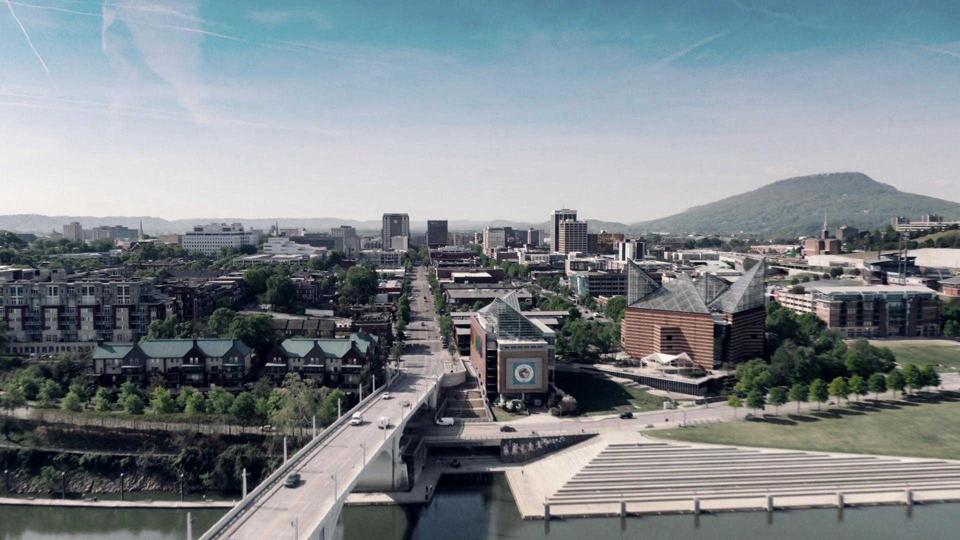} & 
    \includegraphics[width=\linewidth]{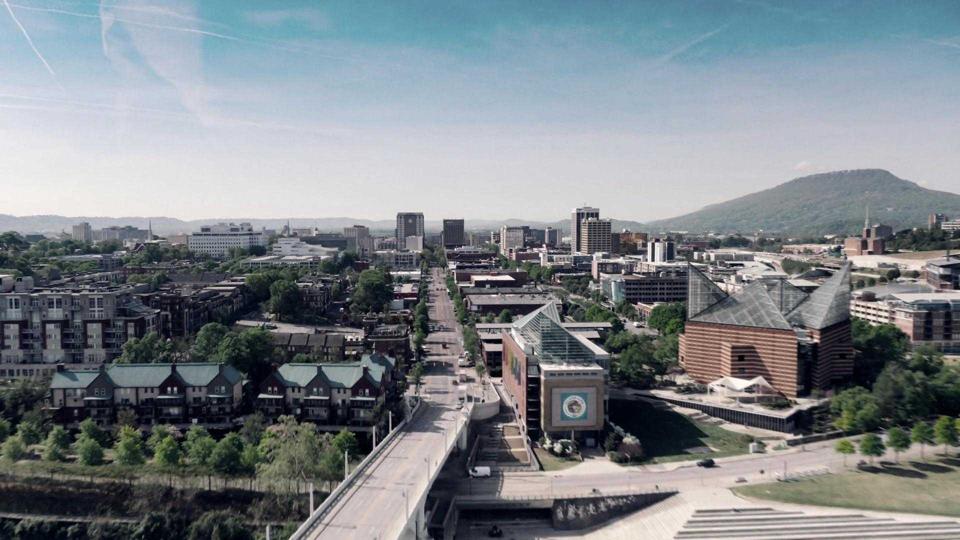} & 
    \includegraphics[width=\linewidth]{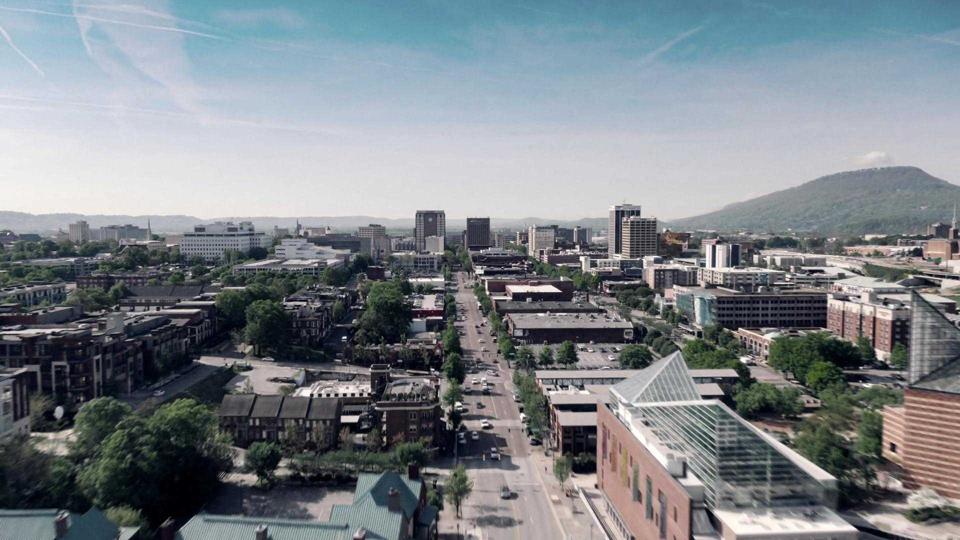} & 
    \includegraphics[width=\linewidth]{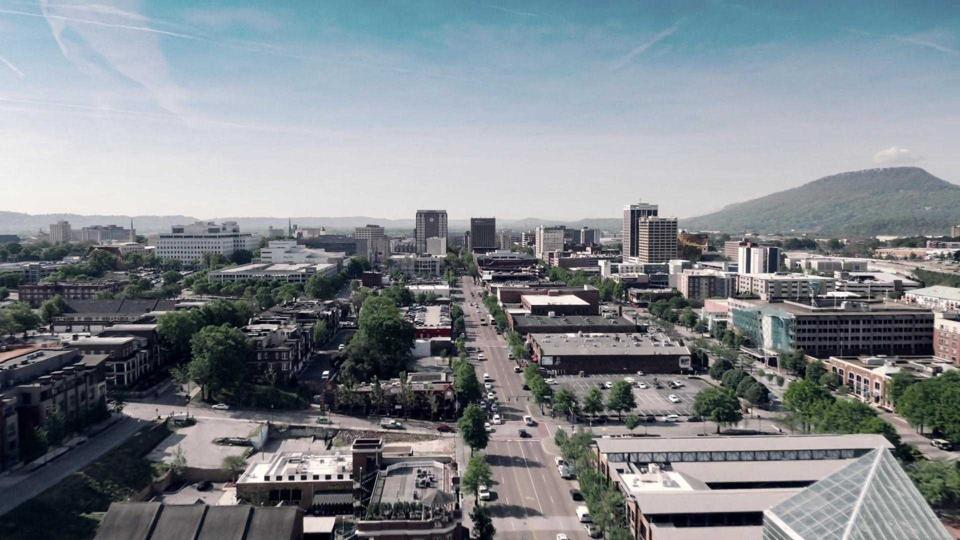} \\

    \centering\scriptsize  Style & 
    \multicolumn{7}{c}{\scriptsize Stylized Video Frames} 
    
  \end{tabularx}
  \caption{\textbf{Results of Video Color Transfer.} ColorFM-L ensures temporally consistent and flicker-free stylization across the sequence.}
  \label{fig:video}
\end{figure*}

\subsection{Comparative Analysis of ColorFM-O and ColorFM-L}
\label{compare}
As shown in \cref{fig:comparison_ir}, while both methods perform comparably in simple scenarios (Row 1), complex scenes reveal their differences.
 Specifically, when distinct semantic regions exhibit varying distribution gaps and transport complexities, the fixed setting of ColorFM-O (\eg, a fixed number of training steps) struggles to establish a precise unified velocity field for all regions simultaneously. In contrast, ColorFM-L achieves robust color transfer by leveraging large-scale data supervision. This suggests that large-scale learning can help ColorFM-L smooth instance-specific optimization irregularities, leading to more stable transfer across diverse scenes.

\subsection{Implicit State Visualization}
\cref{fig:state} visualizes the intermediate states $Z_A$ and $Z_B$, extracted during the $I_A \to I_B$ and $I_B \to I_A$ color transfer processes, respectively. Notably, both states preserve their original structures while converging to a shared color distribution, perceptually appearing as an averaged intermediate style that blends the color characteristics of both $I_A$ and $I_B$. This observation suggests that the implicit state effectively bridges diverse distributions into a unified color manifold. Moreover, ColorFM-L enables efficient bidirectional transfer (\ie, $I_A \leftrightarrow I_B$): by computing the transport parameters once, it reuses them for the reverse path, avoiding redundant network evaluations.

\subsection{Extension to Video Color Transfer}
We extend ColorFM-L to video stylization by sampling a representative frame to compute transfer parameters conditioned on the content/style image. These fixed parameters are then applied to all frames via bidirectional linearized transport, ensuring a consistent style-conditioned mapping across the sequence without using video-specific temporal modules or optical-flow-based constraints. As shown in \cref{fig:video}, this approach avoids the frame-by-frame prediction variance that typically causes flickering artifacts.

\section{Conclusion}
In this paper, we proposed ColorFM, a novel optimization-to-learning framework for color transfer. ColorFM bridges online optimization and offline inference by modeling color transfer as pixel distribution transport via Flow Matching. Specifically, ColorFM-O performs instance-specific optimization through hierarchical color coupling with semantic priors, generating high-quality pseudo-ground truths. ColorFM-L subsequently leverages these data for supervised learning. It predicts flow parameters via implicit state modeling and executes color transfer through bidirectional linearized transport.
Extensive experiments demonstrate that ColorFM-L outperforms existing state-of-the-art methods.

Despite these advantages, ColorFM has several limitations. For example, ColorFM-O inherits the sensitivity of optimization-based approaches and may require per-instance hyperparameter tuning to achieve optimal performance. Moreover, ColorFM-L may face generalization challenges for extreme color styles or out-of-distribution semantic layouts.

% \clearpage  % TODO FINAL: This \clearpage needs to be removed from both review and camera-ready versions.

\section*{Acknowledgements}
This work was supported by the National Natural Science Foundation of China (Grant No.~62572234), the Gusu Innovation and Entrepreneurship Leading Talent Program (Grant No.~ZXL20254324), and the Suzhou Key Technologies Project (Grant No.~SGY2023136).

\bibliographystyle{splncs04}
\bibliography{main}
\end{document}